\documentclass[twoside,11pt]{article}

\usepackage[abbrvbib, preprint]{jmlr2e}

\usepackage{amsmath}
\usepackage{amsfonts}       
\usepackage{mathrsfs}       
\usepackage{comment}
\usepackage{mathtools}
\usepackage{bbm}
\usepackage[usenames]{color} 
\usepackage{tikz}
\usetikzlibrary{shapes,snakes,overlay-beamer-styles,fadings,shadows,positioning,automata,arrows,fit,backgrounds,calc,bayesnet,mindmap,spy}
\usepackage{caption}
\usepackage{subcaption}
\usepackage{url}
\graphicspath{ {./figures/} }
\usepackage[ruled,linesnumbered]{algorithm2e}
\usepackage[utf8]{inputenc} 
\usepackage[T1]{fontenc}    
\usepackage{ulem}
\usepackage{url}            
\usepackage{booktabs}       
\usepackage{nicefrac}       
\usepackage{microtype}      
\usepackage[section]{placeins}

\definecolor{darkblue}{rgb}{0,0,.50} 
\hypersetup{colorlinks=true, allcolors=darkblue, pdfborder={0 0 0}}

\newcommand{\bi}{\begin{itemize}}
\newcommand{\ei}{\end{itemize}}

\newcommand{\be}{\begin{enumerate}}
\newcommand{\ee}{\end{enumerate}}

\newcommand{\ie}{\textit{i.e.,}}
\newcommand{\eg}{\textit{e.g.,}}

\renewcommand{\eqref}[1]{Eq.~\ref{eq:#1}}

\newcommand{\textsim}{\raise.17ex\hbox{$\scriptstyle\sim$}}

\newcommand{\betabinomialpmf}{\mathrm{BeBi}} 
\newcommand{\binomialpmf}{\mathrm{Bin}}  
\newcommand{\categoricalpmf}{\mathrm{Cat}} 
\newcommand{\dirichletmultinomialpmf}{\mathrm{DirMult}} 
\newcommand{\gammapoissonpmf}{\mathrm{GaPo}} 
\newcommand{\multinomialpmf}{\mathrm{Mult}} 
\newcommand{\poissonpmf}{\mathrm{Po}}  

\newcommand{\betapdf}{\mathrm{Beta}} 
\newcommand{\catpmf}{\mathrm{Cat}} 
\newcommand{\dirpdf}{\mathrm{Dir}} 
\newcommand{\dirichletpdf}{\mathrm{Dir}} 
\newcommand{\gammapdf}{\mathrm{Ga}} 
\newcommand{\poissonpdf}{\mathrm{Poisson}} 

\newcommand{\KL}{\mathrm{KL}}  

\newcommand{\given}{\:|\:}      
\newcommand{\dif}{\,d}          

\newcommand{\iidsim}{\stackrel{\mathrm{iid}}{\sim}}

\DeclareMathOperator*{\argmax}{arg\,max}  
\DeclareMathOperator{\ind}{\mathbb{I}}    

\newcommand{\calL}{{\cal L}}
\newcommand{\calM}{{\cal M}}
\newcommand{\calN}{{\cal N}}

\newcommand{\calQ}{{\cal Q}}

\newcommand{\bbE}{\mathbb{E}}

\newcommand{\bbH}{\mathbb{H}}

\newcommand{\bbN}{\mathbb{N}}

\newcommand{\bbR}{\mathbb{R}}
\newcommand{\bbS}{\mathbb{S}}

\newcommand{\sfx}{\mathsf{x}}
\newcommand{\sfy}{\mathsf{y}}

\DeclareSymbolFont{sfletters}{OML}{cmbrm}{m}{it}
\DeclareMathSymbol{\sftheta}{\mathord}{sfletters}{"12}

\newcommand{\kilometers}{{\rm km}} 

\definecolor{nodered}{RGB}{255,99,71}     
\definecolor{nodeyellow}{RGB}{255,215,0}  
\definecolor{nodeblue}{RGB}{135,206,250}  
\definecolor{nodegreen}{RGB}{144,238,144} 
\definecolor{nodegray}{RGB}{192,192,192}  

\jmlrheading{1}{}{}{}{}{}{}

\ShortHeadings{Lightweight Data Fusion with Conjugate Mappings}{Dean, Lee, Pacheco, and Fisher}
\firstpageno{1}

\begin{document}

\title{Lightweight Data Fusion with Conjugate Mappings}

\author{\name Christopher L. Dean \email cdean@csail.mit.edu \\
        \addr Computer Science and Artificial Intelligence Laboratory \\
        Massachusetts Institute of Technology\\
        Cambridge, MA 02139, USA
        \AND
        \name Stephen J. Lee \email leesj@csail.mit.edu \\
        \addr Computer Science and Artificial Intelligence Laboratory \\
        Massachusetts Institute of Technology\\
        Cambridge, MA 02139, USA
        \AND
        \name Jason Pacheco \email pachecoj@cs.arizona.edu \\
        \addr Department of Computer Science \\
        University of Arizona\\
        Tucson, AZ 85721, USA
        \AND
        \name John W. Fisher III \email fisher@csail.mit.edu \\
        \addr Computer Science and Artificial Intelligence Laboratory \\
        Massachusetts Institute of Technology\\
        Cambridge, MA 02139, USA
       }


\maketitle

\begin{abstract}
We present an approach to data fusion that combines the interpretability of structured probabilistic graphical models with the flexibility of neural networks.   
The proposed method, lightweight data fusion (LDF), emphasizes posterior analysis over latent variables using two types of information: primary data, which are well-characterized but with limited availability, and auxiliary data, readily available but lacking a well-characterized statistical relationship to the latent quantity of interest.  
The lack of a forward model for the auxiliary data precludes the use of standard data fusion approaches, while the inability to acquire latent variable observations severely limits direct application of most supervised learning methods.
LDF addresses these issues by utilizing neural networks as conjugate mappings of the auxiliary data: nonlinear transformations into sufficient statistics with respect to the latent variables.
This facilitates efficient inference by preserving the conjugacy properties of the primary data and leads to compact representations of the latent variable posterior distributions.
We demonstrate the LDF methodology on two challenging inference problems:  (1)~learning electrification rates in Rwanda from satellite imagery, high-level grid infrastructure,  and other sources; and (2)~inferring county-level homicide rates in the USA by integrating socio-economic data using a mixture model of multiple conjugate mappings. 

\end{abstract}

\section{Introduction}
\label{sec-introduction}

In many data fusion applications one has access to a variety of data sources for making inferences about unobserved latent quantities.
Effective integration of disparate data sources yields more precise inferences than is possible with each data source used in isolation.
For example, multi-sensor fusion methods typically utilize a common joint representation to combine various sensor modalities \citep{fisher02ijhpca, cetin2006,khaleghi2013multisensor}.  
Specializations of data fusion are often applied to combine audio, video, and text data to improve scene understanding \citep{fisher00nips, fisher03icme, fisher04, siracusa03icmi, srivastava2012multimodal, ngiam2011multimodal, cabezas2015iccv}. 
Many data fusion methods assume a known forward model for all data types, precluding their use in many important problems where some of the data lacks a well-characterized forward model, e.g. in computational biology or the social sciences.

\begin{figure*}[t]
\vskip 0.1in
\begin{center}
\centerline{\includegraphics[width=\linewidth, trim={0cm 0cm 0cm 0.0cm},clip]{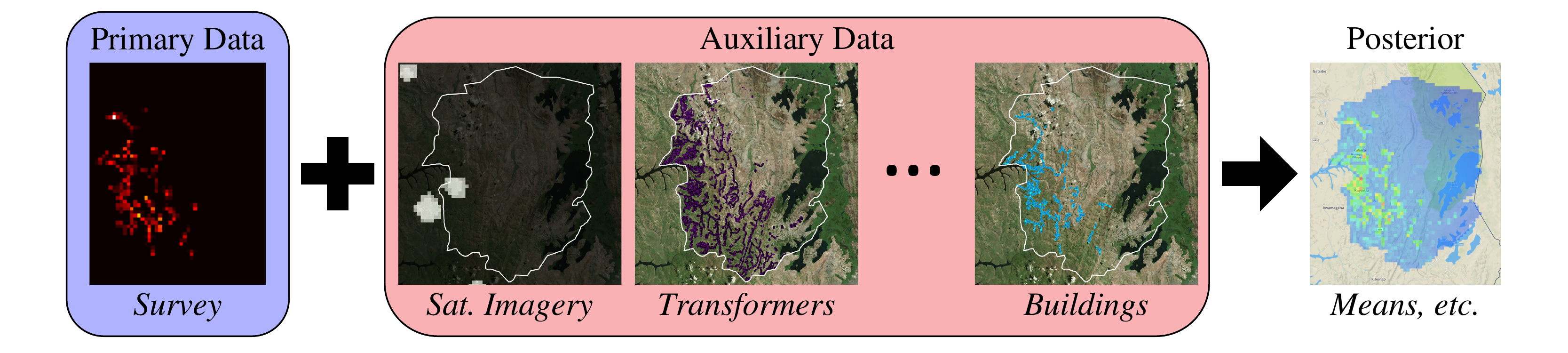}}

\caption{ \small\textbf{Lightweight data fusion} jointly models primary and auxiliary data.  Our motivating example models a primary electrification survey (left) and auxiliary features (center) which include satellite imagery and known infrastructure locations.  Through nonlinear transformations auxiliary data is mapped to yield closed-form posterior updates by leveraging conjugacy properties of the exponential family.  The resulting predictions and uncertainties (right) are available at all locations.}
\label{fig-summary}
\end{center}
\vskip -0.2in
\end{figure*}

We introduce a general methodology for flexible and efficient data fusion, \textit{lightweight data fusion} (LDF), that combines the interpretability of probabilistic graphical models (PGMs) with the flexibility and expressiveness of neural networks (NNs).
LDF considers two distinct classes of data that differ in their respective characterizations. 
\textit{Primary} data has a well-characterized statistical relationship to latent variables of interest.  
\textit{Auxiliary} data conveys information about the latent variables, but the statistical relationship is unknown and/or complex. 
Figure~\ref{fig-summary} depicts an application for modeling electricity access in developing nations.  
Here, the primary data is a survey of electrical infrastructure \citep{eucl2020,Sofreco2013}, characterized as a binomial random variable, but costly to obtain. 
Conversely, auxiliary data comes from a variety of ubiquitous, readily available, and lower-cost sources as compared to the primary data.  
Examples include both daytime and nighttime satellite imagery and other information about existing infrastructure (e.g. high-to-low-voltage transformers), but their precise statistical relationship to the underlying electrification rate is unknown. LDF utilizes primary data to \textit{learn} a compact and interpretable mapping of auxiliary data suitable for computationally-efficient posterior inference over \textit{latent} variables.

PGMs are a natural framework for multi-modal data fusion \citep{Wainwright2007, Koller2009}. 
PGMs have appealing properties including robustness to nonstationarity, well-defined model validation, and interpretability. 
However, data types with complex joint statistical relationships often prohibit efficient learning and inference, even when the model is known.
Conversely, NNs have led to great advances in data fusion for predictive tasks \citep{Goodfellow2016}.
They naturally apply to cases where the data generating process is unknown, as theoretically an NN with a single hidden layer of infinite width can represent any function \citep{Neal1995}.  In practice, however, these models can be challenging to train and the learned latent state representations lack interpretability.
Finally, discriminative deep learning (e.g. regression and classification) typically requires prohibitively large amounts of explicitly-labeled data; as noted above explicit latent variable labels are generally unavailable for many applications.

LDF avoids the need for explicit representation of complex generative models by learning NN transformations of auxiliary data into exponential family sufficient statistics. 
These transformations, which we call \textit{conjugate mappings}, yield compact representations of the auxiliary data while preserving conjugacy properties of the primary data model. 
Our discussion focuses on conjugate mappings that are NNs owing to their flexibility, however, LDF accommodates any number of nonlinear function classes, e.g. Gaussian processes or basis functions.
One might consider learning conjugate mappings via any number of supervised learning methods. Unfortunately, latent variable observations (labeled or otherwise) are generally unavailable precluding most, if not all, supervised methods. 
LDF learns mappings that make the auxiliary data maximally-informative about the primary data.
The learned mapping allows for efficient posterior inference using both data types or the primary or auxiliary data individually.

Despite apparent similarities between LDF and generalized linear models (GLMs)~\citep{Nelder1972, Mccullagh1989} the distinctions are fundamental, as we show in Section~\ref{sec-related}. 
In particular, GLM approaches aim to \textit{predict} the primary data (response variables) from the auxiliary data (covariates), potentially in ways that are unrelated to a latent variable model. 
While LDF makes use of primary data, regression methods are inadequate for the explicit aim of \textit{posterior inference} over \textit{latent variables}, \ie~\textit{conditioning} on primary \textit{and} auxiliary data. 
The LDF learning criterion is derived \textit{from} the PGM structure, whereas to the extent that GLMs provide a latent variable model, it is limited to the coefficients of the linear transformation and other hierarchical parameters. By contrast, LDF readily and flexibly extends to more complex PGM structures.

Combining PGM structures with NN mappings, as proposed here, mitigates issues that would arise when using deep generative models in the same fashion.  For example, generative adversarial networks (GANs) and variational auto-encoders (VAEs) attempt to approximate unknown forward models but lack an interpretable latent structure suitable for subsequent reasoning \citep{Rezende2014, Goodfellow2014, Kingma2014}.
Structured VAEs (SVAEs) \citep{Johnson2016} are closely related to our approach and combine a latent PGM structure, an unknown NN likelihood model, and inference using a conjugate potential with a recognition network as we describe in Sec.~\ref{sec-related}.  
Critically, these models treat all data as auxiliary data, leading to potential degeneracies in the latent variable posteriors. 
Our LDF setting incorporates primary data, which conditions \textit{directly} on the underlying latent variable structure through a known conditional distribution.  This ensures that the underlying latent variable structure captures meaningful variations in the data, rather than this complexity being learned as part of the NN.
Additionally, the LDF training and inference procedures are generally more straightforward to formulate and implement.

\textbf{Contributions.} 
We introduce a novel methodology for fusion of disparate data types that improves on existing methods.  
(1)~Our approach is \textit{lightweight} based on properties of exponential families, as we develop and learn \textit{conjugate mappings} that map auxiliary data into exponential family sufficient statistics w.r.t. the latent variables.  
(2)~The proposed method addresses the lack of direct latent variable observations by exploiting the known primary data models and facilitates efficient inference with respect to latent variables. 
(3)~We demonstrate the application of our approach to two challenging real-world inference problems: predicting electrification status in developing nations using satellite imagery and infrastructure data, and inferring county-level homicide rates from socio-economic data.


\section{Exponential Families, Sufficiency, and Conjugate Mappings}
\label{sec-mappings}

LDF is a novel method for inference in PGMs that allows the use of auxiliary data via conjugate mappings.  
This renders inference using auxiliary data as equivalent to inference using primary data while preserving interpretability, compactness of representation, and computational efficiency.
This section motivates and defines conjugate mappings; subsequently, Section~\ref{sec-learning-and-inference} describes a flexible learning approach suitable for a broad class of mappings, including NNs.

\begin{figure}[t]
  \centering
  \resizebox{0.75\textwidth}{!}{%
%

\begin{tikzpicture}
%
%
\tikzstyle{box}+=[ rounded corners = 5pt,
                       align           = left,
                       font            = \footnotesize,
                       text width      = 2.45cm]

\tikzstyle{every path}+=[thin,>=latex];
\tikzstyle{plate}+=[thin,sharp corners,inner sep=5pt,outer sep=0pt]; 
%
%

%
%
\def \xmncolor {red!70}
\def \thetcolor {cyan!70!black}
\def \zmncolor {yellow!70!black}
\def \betcolor {green!70!black}

%
%
%
\tikzstyle{latent}+=[node distance=1.25 and 1.5, on grid];
\tikzstyle{obs}+=[fill=blue!30,node distance=1.25 and 1.5,on grid];
\tikzstyle{obsaux}=[latent,fill=red!30,node distance=1.25 and 1.5,on grid];
\tikzstyle{obsauximp}=[latent,fill=green!30,node distance=1.25 and 1.5,on grid];
\tikzstyle{const}+=[node distance=1.25 and 1.5,on grid];

%
%

%
%
\node[obsaux] (Ttildx1) {$x_1$};
\node[latent,right=1.5 of Ttildx1] (theta1) {$\theta_1$};
\node[obs,right = 1.5 of theta1] (y1j) {$y_{1j}$};
%
%
\edge {theta1} {Ttildx1};
\edge {theta1} {y1j}; 

\plate {yplate1} {(y1j)} {$N_1$}; 

\node[const,right=4 of Ttildx1] (cdots) {$\cdots$};

%
%
\node[obsaux,right=1 of cdots] (TtildxM) {$x_M$};
\node[latent,right=1.5 of TtildxM] (thetaM) {$\theta_M$};
\node[obs,right = 1.5 of thetaM] (yMj) {$y_{Mj}$};
%
%
\edge {thetaM} {TtildxM};
\edge {thetaM} {yMj}; 

\plate {yplateM} {(yMj)} {$N_M$}; 

%
%
\node[const,above=2 of theta1] (thetaparam) {$\lambda_0$};
\draw[->,>=latex,shorten <=2pt] (thetaparam) to [out= -90,in=90] (theta1);
\draw[->,>=latex,shorten <=2pt] (thetaparam) to [out=0,in=135] (thetaM);


\tikzstyle{latent}+=[densely dashed];
\tikzstyle{plate}+=[color=black];

%
%





\end{tikzpicture}

}
  \caption{\small{Data fusion in a simple model} where $M$ latent variables $\theta_i$ are accompanied by corresponding shaded observation nodes $x_i$ and $y_i = \{y_{i1}, \ldots, y_{Mj}\}$.  The direction of the arrows indicate \textit{known} conditional distributions.}
  \label{fig:LDF-gmgen}
\end{figure}
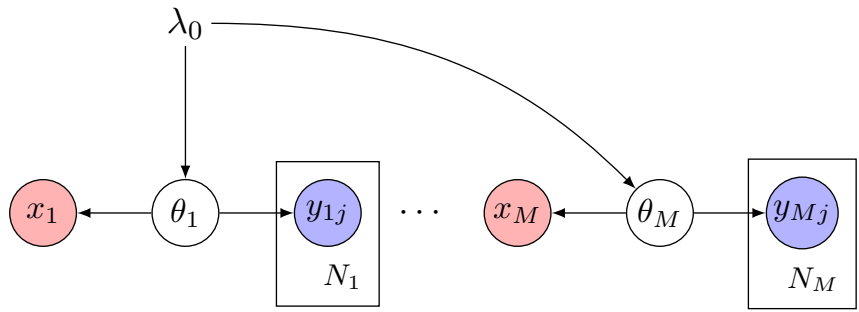

We begin by specifying a general data fusion inference problem.
Consider a model with two data sources, \textit{auxiliary} data $x = (x_1, \ldots, x_M)$ and \textit{primary} data $y = (y_1, \ldots, y_M)$, that are parameterized by \textit{local} latent variables $\theta = (\theta_1, \ldots, \theta_M)$.
\textit{Global} latent variables $\gamma$ have arbitrary (known) internal structure that parameterize the \textit{local} latent variables $\theta_i$.
The goal of inference is the posterior distribution:
\begin{align}
    p(\theta, \gamma \given x, y)
        \propto p(\gamma) \prod_{i=1}^M p(\theta_i \given \gamma) p(x_i \given \theta_i) p(y_i \given \theta_i).
    \label{eq-fusion-general}
\end{align}
In many settings, $\theta$ are the random variables of interest and so $\gamma$ may be treated as nuisance variables.
When all the terms in Equation~\ref{eq-fusion-general} are known, this posterior may be calculated approximately using variational inference \citep{Jordan1999, Wainwright2007, Hoffman2013} or Markov chain Monte Carlo (MCMC) methods \citep{Metropolis1953, Hastings1970, Geman1984, Robert2004}.

We focus on the case where the auxiliary data likelihood $p(x_i \given \theta_i)$ is unknown, precluding direct application of standard Bayesian inference methods.
Here we propose an approach that extracts information from auxiliary data $x$ \textit{without} explicitly specifying $p(x_i \given \theta_i)$ or overly-complicating inference $p(\theta, \gamma \given x, y)$ relative to primary data inference:
\begin{align}
    p(\theta, \gamma \given y) 
    \propto p(\gamma) \prod_{i=1}^M p(\theta_i \given \gamma) p(y_i \given \theta_i).
    \label{eq-fusion-primary}
\end{align}
We refer to our framework as \textit{lightweight} data fusion.
It is lightweight in the sense that posterior inference with primary and auxiliary data types jointly is not much more complicated than with primary data in isolation, Eq.~\ref{eq-fusion-primary}.

While our LDF approach and Eq.~\ref{eq-fusion-general} contains a variety of PGM structures (e.g. more complex latent variable models, multiple primary and auxiliary data types) for clarity of exposition we initially omit any complex global structure: the set of \textit{global} latent variables $\gamma \triangleq \emptyset$.  
Therefore we obtain the model in Figure~\ref{fig:LDF-gmgen} that factorizes as
\begin{align}
    p(\theta \given x, y)
        \propto \prod_{i=1}^M p(\theta_i) p(x_i \given \theta_i) p(y_i \given \theta_i).
    \label{eq-fusion-iid}
\end{align}
While apparently simple, the model of Eq.~\ref{eq-fusion-iid} and Figure~\ref{fig:LDF-gmgen} depicts the central properties: 
(1)~primary data $y$ is of interest solely for purposes of posterior inference, $p(\theta \given x, y)$, and 
(2)~unobserved $\theta_i$'s \textit{differ} for each pair-wise instance of $\left\{x_i, y_i\right\}$. 

\subsection{Exponential Families and Sufficient Statistics}

We consider the case when the \textit{known} forward model $p(y_i \given \theta_i)$ is a distribution in the exponential family  and $p(\theta_i \given \gamma)$ is a conjugate prior \citep{Bernardo2000}.
It is in this sense that primary data $y_i$ has a \textit{well-characterized} probabilistic relationship to $\theta_i$.
The exponential family contains many commonly-used distributions, including the Bernoulli, binomial, multinomial, Poisson, Gaussian, beta, Dirichlet, gamma, exponential, and others.
They are the \textit{only} class of distributions for which conjugate priors exist, reducing the complexity of doing inference to calculating and adding sufficient statistics.  
Finally, they make the fewest \textit{a priori} assumptions about the data, since they arise as solutions to the maximum entropy problem subject to linear constraints.
The exponential family assumption may be relaxed with respect to primary data (with some loss of computational efficiency); for purposes of introducing the method we maintain it throughout.  

We briefly summarize relevant properties of exponential family distributions that play a central role in our analysis.
The reduction of primary data to \textit{sufficient statistics} and the tractability of the resulting posterior-predictive likelihood are of particular importance.
The former motivates the output of the auxiliary data mapping while the latter enables use of various Bayesian learning objectives. 

The primary data $y_i=\{y_{i1},\cdots,y_{iN_i}\}$ in Figure~\ref{fig:LDF-gmgen} are drawn from a known exponential family distribution with unknown latent parameters $\theta_i$,
\begin{align}
  p(y_{ij} \given \theta_i) 
  = h_{\sfy_0}(y_{ij}) \exp\left\{ \eta(\theta_i)^\top t_{\sfy_0}(y_{ij}) - A(\theta_i) \right\}
  \label{eq-likelihood-expfam}
\end{align}
where $\eta(\theta_i)$ are the \textit{natural parameters},
$h_{\sfy_0}(y_{ij})$ is the \textit{base measure}, $t_{\sfy_0}(y_{ij})$ is the
\textit{sufficient statistics}, and $A(\theta_i)
\triangleq
\log \int h_{\sfy_0}(y_{ij}) \exp\left\{ \eta(\theta_i)^\top t_{\sfy_0}(y_{ij}) \right\} \dif y_{ij}$ is the \textit{log-partition function}.
Clearly, the joint distribution of a set of $N_i$ conditionally-independent observations $y_i = \{y_{i1}, \ldots, y_{iN_i}\}$ is also in the exponential family with identical natural parameter $\eta(\theta_i)$, sufficient statistics $t_\sfy(y_i) = \sum_j t_{\sfy_0}(y_{ij})$, base measure $h_\sfy(y_i) = \prod_j h_{\sfy_0}(y_{ij})$, and log-partition function $N_i A(\theta_i)$:
\begin{align}
 	p(y_i \given \theta_i)
        = f(y_i; \theta_i)
        \triangleq h_\sfy(y_i) \exp\left\{ \eta(\theta_i)^\top t_\sfy(y_i) - N_i A(\theta_i)\right\}.
\end{align}
We use a conjugate prior distribution on $\theta_i$  \citep{diaconis1979} with density
\begin{align}\label{eq-prior-expfam}
  p(\theta_i)
  = \pi(\theta_i; \lambda_0)
  \triangleq
  h_{\sftheta}(\theta_i) \exp\left\{
  \begin{bmatrix}
  \tau_0 \\ \nu_0
  \end{bmatrix}^\top
  \begin{bmatrix}
  \eta(\theta_i) \\ -A(\theta_i)
  \end{bmatrix} - \log Z(\tau_0, \nu_0) 
  \right\},
\end{align}
that is also in the exponential family with natural parameters $\lambda_0 \triangleq [\tau_0,\: \nu_0]^\top$, sufficient statistic
$[\eta(\theta_i),\: -A(\theta_i)]^\top$, base measure $h_{\sftheta}(\theta_i)$, and
partition function 
\begin{align}
Z(\tau, \nu) 
    \triangleq \int h_{\sftheta}(\theta_i) \exp
    \left\{
 \begin{bmatrix}
 \tau & \nu
 \end{bmatrix}
 \begin{bmatrix}
 \eta(\theta_i) & -A(\theta_i)
 \end{bmatrix}^\top
 \right\}
    \dif\theta_i. \label{eq-prior-partition-expfam}
\end{align}
While $Z(\tau, \nu)$ is not guaranteed to be finite, it is for most exponential family likelihoods.

\begin{figure}[t]
  \centering
  \begin{tabular}{cc}
  \resizebox{0.75\textwidth}{!}{%
%

\begin{tikzpicture}
%
%
\tikzstyle{box}+=[ rounded corners = 5pt,
                       align           = left,
                       font            = \footnotesize,
                       text width      = 2.45cm]

\tikzstyle{every path}+=[thin,>=latex];
\tikzstyle{plate}+=[thin,sharp corners,inner sep=5pt,outer sep=0pt]; 

%
\draw[very thin, color=gray!0](-0.5, 0) grid (8.825, 0);
%

%
%
\def \xmncolor {red!70}
\def \thetcolor {cyan!70!black}
\def \zmncolor {yellow!70!black}
\def \betcolor {green!70!black}

%
%
%
\tikzstyle{latent}+=[node distance=1.25 and 1.5, on grid];
\tikzstyle{obs}+=[fill=blue!30,node distance=1.25 and 1.5,on grid];
\tikzstyle{obsaux}=[latent,fill=red!30,node distance=1.25 and 1.5,on grid];
\tikzstyle{obsauximp}=[latent,fill=green!30,node distance=1.25 and 1.5,on grid];
\tikzstyle{const}+=[node distance=1.25 and 1.5,on grid];

%
%

%
%

\node[latent] (theta1) {$\theta_1$};
\node[obs, right = 1.5 of theta1] (y1j) {$y_{1j}$};
\node[obs, right = 1.5 of y1j] (Ty1) {$T_\mathsf{y}(y_1)$};
%
%
\edge {theta1} {y1j}; 
\edge {y1j} {Ty1}; 

\plate {yplate1} {(y1j)} {$N_1$}; 

\node[const,right=4.25 of theta1] (cdots) {$\cdots$};

%
%
\node[latent,right=1 of cdots] (thetaM) {$\theta_M$};
\node[obs, right = 1.5 of thetaM] (yMj) {$y_{Mj}$};
\node[obs, right = 1.5 of yMj] (TyM) {$T_\mathsf{y}(y_M)$};

%
%
\edge {thetaM} {yMj}; 
\edge {yMj} {TyM}; 

\plate {yplateM} {(yMj)} {$N_M$}; 

%
%
\node[const,above=2 of theta1] (thetaparam) {$\lambda_0$};
\draw[->,>=latex,shorten <=2pt] (thetaparam) to [out= -90,in=90] (theta1);
\draw[->,>=latex,shorten <=2pt] (thetaparam) to [out=0,in=135] (thetaM);


\tikzstyle{latent}+=[densely dashed];
\tikzstyle{plate}+=[color=black];

%
%





\end{tikzpicture}

} \\
  (a) \\
  \resizebox{0.75\textwidth}{!}{%
%

\begin{tikzpicture}
%
%
\tikzstyle{box}+=[ rounded corners = 5pt,
                       align           = left,
                       font            = \footnotesize,
                       text width      = 2.45cm]

\tikzstyle{every path}+=[thin,>=latex];
\tikzstyle{plate}+=[thin,sharp corners,inner sep=5pt,outer sep=0pt]; 

%
\draw[very thin, color=gray!0](-0.5, 0) grid (8.825, 0);
%

%
%
\def \xmncolor {red!70}
\def \thetcolor {cyan!70!black}
\def \zmncolor {yellow!70!black}
\def \betcolor {green!70!black}

%
%
%
\tikzstyle{latent}+=[node distance=1.25 and 1.5, on grid];
\tikzstyle{obs}+=[fill=blue!30,node distance=1.25 and 1.5,on grid];
\tikzstyle{obsaux}=[latent,fill=red!30,node distance=1.25 and 1.5,on grid];
\tikzstyle{obsauximp}=[latent,fill=green!30,node distance=1.25 and 1.5,on grid];
\tikzstyle{const}+=[node distance=1.25 and 1.5,on grid];

%
%

%
%

\node[latent] (theta1) {$\theta_1$};
\node[obs, right = 1.5 of theta1] (Ty1) {$T_\mathsf{y}(y_1)$};
\node[obs, right = 1.5 of Ty1] (y1j) {$y_{1j}$};

%
%
\edge {theta1} {Ty1}; 
\edge {Ty1} {y1j}; 

\plate {yplate1} {(y1j)} {$N_1$}; 

\node[const,right=4.25 of theta1] (cdots) {$\cdots$};

%
%
\node[latent,right=1 of cdots] (thetaM) {$\theta_M$};
\node[obs, right = 1.5 of thetaM] (TyM) {$T_\mathsf{y}(y_M)$};
\node[obs, right = 1.5 of TyM] (yMj) {$y_{Mj}$};

%
%
\edge {thetaM} {TyM}; 
\edge {TyM} {yMj}; 

\plate {yplateM} {(yMj)} {$N_M$}; 

%
%
\node[const,above=2 of theta1] (thetaparam) {$\lambda_0$};
\draw[->,>=latex,shorten <=2pt] (thetaparam) to [out= -90,in=90] (theta1);
\draw[->,>=latex,shorten <=2pt] (thetaparam) to [out=0,in=135] (thetaM);


\tikzstyle{latent}+=[densely dashed];
\tikzstyle{plate}+=[color=black];

%
%





\end{tikzpicture}

} \\
  (b) \\
  \resizebox{0.75\textwidth}{!}{%
%

\begin{tikzpicture}
%
%
\tikzstyle{box}+=[ rounded corners = 5pt,
                       align           = left,
                       font            = \footnotesize,
                       text width      = 2.45cm]

\tikzstyle{every path}+=[thin,>=latex];
\tikzstyle{plate}+=[thin,sharp corners,inner sep=5pt,outer sep=0pt]; 

%
\draw[very thin, color=gray!0](-1.9, 0) grid (7.425, 0);
%

%
%
\def \xmncolor {red!70}
\def \thetcolor {cyan!70!black}
\def \zmncolor {yellow!70!black}
\def \betcolor {green!70!black}

%
%
%
\tikzstyle{latent}+=[node distance=1.25 and 1.5, on grid];
\tikzstyle{obs}+=[fill=blue!30,node distance=1.25 and 1.5,on grid];
\tikzstyle{obsaux}=[latent,fill=red!30,node distance=1.25 and 1.5,on grid];
\tikzstyle{obsauximp}=[latent,fill=green!30,node distance=1.25 and 1.5,on grid];
\tikzstyle{const}+=[node distance=1.25 and 1.5,on grid];

%
%

%
%

\node[latent] (theta1) {$\theta_1$};
\node[obs, right = 1.5 of theta1] (Ty1) {$T_\mathsf{y}(y_1)$};

%
%
\edge {Ty1} {theta1}; 

%

\node[const,right=1.375 of Ty1] (cdots) {$\cdots$};

%
%
\node[latent,right=1 of cdots] (thetaM) {$\theta_M$};
\node[obs, right = 1.5 of thetaM] (TyM) {$T_\mathsf{y}(y_M)$};

%
%
\edge {TyM} {thetaM}; 


%
%
\node[const,above=2 of theta1] (thetaparam) {$\lambda_0$};
\draw[->,>=latex,shorten <=2pt] (thetaparam) to [out= -90,in=90] (theta1);
\draw[->,>=latex,shorten <=2pt] (thetaparam) to [out=0,in=135] (thetaM);


\tikzstyle{latent}+=[densely dashed];
\tikzstyle{plate}+=[color=black];

%
%





\end{tikzpicture}

} \\
  (c)
  \end{tabular}
  \caption{\small{Three equivalent subgraphs showing inference over $\theta_1, \ldots, \theta_M$ based on  \textit{\textcolor{blue!60}{primary} data only}.  These graphs are equivalent because $T_\sfy(y_i)$ is \textit{sufficient} for $y_i$ w.r.t. inferences about $\theta_i$.  
  \textbf{(a)} Inference for $p(\theta_i \given y_i)$. 
  \textbf{(b)} $p(\theta_i \given T_\sfy(y_i)))$ that is equivalent in distribution to the former by sufficiency.  Given a sufficient statistic $T_\sfy$ the Markov property shows that $\theta_i$ is conditionally independent from the actual data realization $y_i$.
  \textbf{(c)} The original data $y_i$ may be discarded, as inference only relies on the sufficient statistics $T_\sfy(y_i)$.}}
   \label{fig:LDF-gmsuffstats}
\end{figure}
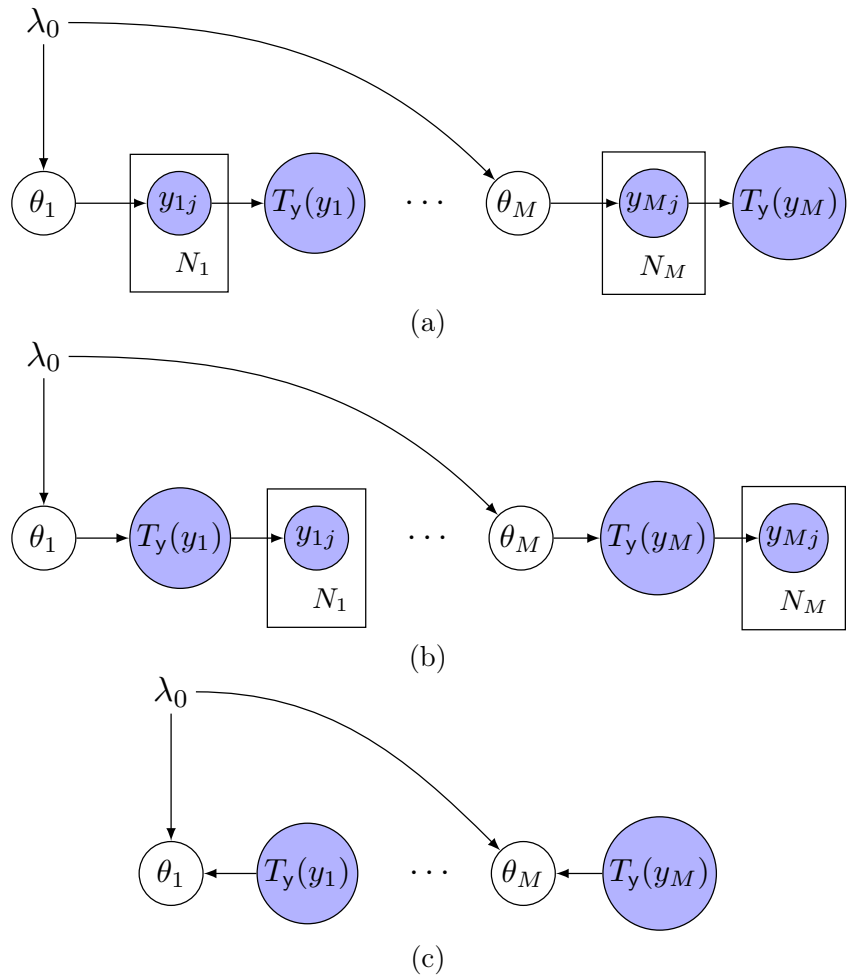

Consequently, the posterior distribution conditioned on $y_i$ is given by
\begin{align}
p(\theta_i \given y_i)
  &= \pi(\theta_i; \lambda_0 + T_\sfy(y_i)) \\
  &= h_{\sftheta}(\theta_i) \exp\left\{ 
\begin{bmatrix}
  \tau_0 + t_\sfy(y_i)  \\  \nu_0 +N_i
\end{bmatrix}^\top
\begin{bmatrix}
\eta(\theta_i) \\ -A(\theta_i)
\end{bmatrix}
- \log Z(\tau_0 + t_\sfy(y_i), \nu_0 + N_i)
\right\}
\label{eq-posterior-expfam}
\end{align}
where $T_\sfy(y_i) = [t_\sfy(y_i);\,N_i]$ are the aggregated sufficient statistics.  We refer to $T_\sfy(y_i)$ as the \textit{aggregated} sufficient statistics to disambiguate it from the sufficient statistics function $t_\sfy(y_i)$ in the exponential family form.
Prior $p(\theta_i)$ is referred to as conjugate because the posterior $p(\theta_i \given y_i)$ takes the same form but with updated natural
parameters $[\tau_0 + t_\sfy(y_i),\: \nu_0 + N_i]^\top$.

It is well known that the aggregated sufficient statistics $T_\sfy(y_i) = [t_\sfy(y_i),\:N_i]$ completely summarize the information about the parameter $\theta_i$ in the primary data $y_i$.
The \textit{aggregated} sufficient statistics $T_\sfy(y_i)$ are Bayes sufficient for $y_i$ w.r.t. inferences about $\theta_i$, i.e.
\begin{align}
    p(\theta_i \given y_i) 
        = p(\theta_i \given T_\sfy(y_i)).
\end{align}

This property of sufficiency is further shown in Figure~\ref{fig:LDF-gmsuffstats} that shows three \textit{equivalent} subgraphs that are part of Figure~\ref{fig:LDF-gmgen} when viewed for inferences about $\theta = (\theta_1, \ldots, \theta_M)$.  
Figure~\ref{fig:LDF-gmsuffstats}(a) shows the standard generative model of primary data; the equivalent model in Figure~\ref{fig:LDF-gmsuffstats}(b) shows that given the aggregated sufficient statistics $T_\sfy(y_i)$, the latent variable $\theta_i$ and the data $y_i$ are independent.  
Thus, for inference, one only requires access to the sufficient statistics $T_\sfy(y_i)$---\textit{not} the original data $y_i$, as shown in Figure~\ref{fig:LDF-gmsuffstats}(c).

Conjugate priors and sufficiency impact the posterior predictive distribution of the data.  Given $p(y_i \given \theta_i)$ and the conjugate prior $p(\theta_i)$ as above, the marginal can be written as
\begin{align}
p(y_i)
= \int p(y_i \given \theta_i) p(\theta_i) \dif\theta_i
= h_\sfy(y_i) \dfrac{Z(\tau_0 + t_\sfy(y_i), \nu_0 + N_i)}{Z(\tau_0, \nu_0)} 
\label{eq-marginal-expfam}
\end{align}
so long as $Z(\tau, \nu)$ is finite.
While analytically tractable, this distribution is not usually in the exponential family.
We can additionally write the posterior predictive distribution of a new primary datum $\tilde{y}_i = \{\tilde{y}_{i1}, \ldots,\tilde{y}_{i\tilde{N}_i}\}$ as:
\begin{align}
p(\tilde{y}_i \given y_i)
    = \int p(\tilde{y}_i \given \theta_i) p(\theta_i \given T_\sfy(y_i)) \dif\theta_i
    &= h_\sfy(\tilde{y}_{i})\, \dfrac{Z(\tau_0 + t_\sfy(y_i) + t_\sfy(\tilde{y}_i), \nu_0 + N_i + \tilde{N}_i)}{Z(\tau_0 + t_\sfy(y_i), \nu_0 + N_i)}
\label{eq:postpredict}
\end{align}
which takes the same form as the marginal $p(y_i)$ but with updated parameters $\lambda_0 + T_\sfy(y_i)$.
This property will play a pivotal role in learning and inference for LDF models.

\subsection{Conjugate Mappings of Auxiliary Data}
\label{sec:conj-aux}

Utilizing the properties above, we consider transforming auxiliary data $x_i = (x_{i1}, \ldots, x_{iP})$, subject to $\theta_i$ by some unknown forward model $p(x_i \given \theta_i)$, to a representation that is interpretable as a sufficient statistic (akin to primary data).  
In many applications, the lack of a forward model precludes posterior inference.
Here we bypass the forward model by focusing on the sufficient statistics. 
The methodology is predicated on the following two assumptions:

\begin{enumerate}
\item \textbf{Sufficiency:} there exists a statistic $T^*_\sfx(x_i) = [t^*_\sfx(x_i), n^*_\sfx(x_i)]$ that is Bayes sufficient for $x_i$ w.r.t. $\theta_i$ in the same way that $T_\sfy(y_i)$ is sufficient for $y_i$:
\begin{align}
    p(\theta_i \given x_i) 
    = p(\theta_i \given T^*_\sfx(x_i))
    = \pi(\theta_i; \lambda_0 + T^*_\sfx(x_i)).
\end{align}

\item \textbf{Learnability:} one can learn an approximation to $T^*_\sfx(x_i)$ given enough joint primary/auxiliary data observations and a suitably expressive function class $T_\sfx(x_i; \phi)$.
\end{enumerate}
Consequently, the posterior distribution takes the form
\begin{align}
  p(\theta_i &\given x_i; \phi)
    = \pi(\theta_i; \lambda_0 + T_\sfx(x_i; \phi)) \\
    &= h_{\sftheta}(\theta_i) \exp\!\left\{\!
  \begin{bmatrix}
    \tau_0 + t_\sfx(x_i; \phi) \\ \nu_0 + n_\sfx(x_i; \phi)
  \end{bmatrix}^\top
  \begin{bmatrix}
    \eta(\theta_i) \\ -A(\theta_i)
  \end{bmatrix}
  - \log Z(\tau_0 + t_\sfx(x_i; \phi), \nu_0 + n_\sfx(x_i; \phi))
  \!\right\}\!
  \label{eq-posterior-trfexpfam}
\end{align}
which shows that conditioning on the auxiliary data $x_i$  simplifies to calculating aggregated sufficient statistics $T_\sfx(x_i; \phi)$ and performing a conjugate update, analogous to inference given the primary data.  
It is for this reason that we refer to $T_\sfx(x_i; \phi)$ as a \textit{conjugate mapping} of the auxiliary data $x_i$.  As with the primary data, this model implies $p(\theta_i \given x_i; \phi) = p(\theta_i \given T_\sfx(x_i; \phi))$.

Neither assumption represents a departure from Bayesian methodology.  The existence of sufficient statistics is well-established, as is the existence of \textit{minimal} sufficient statistics under mild regularity conditions.  Other methods learn transformations of input data and/or assume a conjugate surrogate likelihood as we describe in Sec.~\ref{sec-related}.

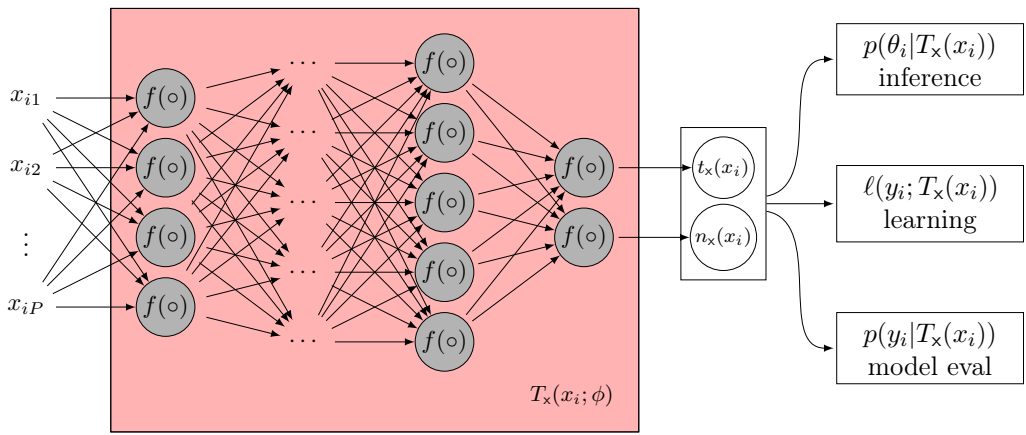
\begin{figure}[t]
  \centering
  \resizebox{0.9\textwidth}{!}{\begin{tikzpicture}
\def \layersep{2}  
\tikzstyle{box}+=[ rounded corners = 5pt,
                       align           = left,
                       font            = \footnotesize,
                       text width      = 2.45cm]

\tikzstyle{every path}+=[thin,>=latex];
\tikzstyle{plate}+=[thin,sharp corners,inner sep=10pt,outer sep=5pt];

\tikzstyle{latent}+=[node distance=1.25 and 1.5, on grid];
\tikzstyle{obs}+=[fill=blue!30,node distance=1.25 and 1.5,on grid];
\tikzstyle{obsaux}=[latent,fill=black!30,node distance=1.25 and 1.5,on grid];
\tikzstyle{obsauximp}=[latent,fill=green!30,node distance=1.25 and 1.5,on grid];
\tikzstyle{const}+=[node distance=1.25 and 1.5,on grid];

%
%
\foreach \name / \y in {1/1,2/2,P/4}
  \node[latent,draw=white] (I\name) at (0,-\y) {$x_{i\name}$};

\node[at= {($(I2)!.5!(IP)$)}]  {$\vdots$};

%
%
 \foreach \name / \y in {1,...,4}
 \node[obsaux] (Ha\name) at (2,-\y) {\small $f(\circ)$};
 
%
%
 \foreach \name / \y in {1,...,5}
 \path[yshift=0.5cm] node[latent,fill=red!30,draw=red!30] (Hm\name) at (4,-\y) {\small $\cdots$};
 
%
%
 \foreach \name / \y in {1,...,5}
   \path[yshift=0.5cm] node[obsaux] (Hb\name) at (6,-\y) {\small $f(\circ)$};

%
%
 \foreach \name / \y in {1,...,2}
   \path[yshift=-1cm]
            node[obsaux] (O\name) at (8,-\y) {\small $f(\circ)$};

%
\foreach \src in {1,2,P}
   \foreach \dst in {1,...,4}
       \draw[->,>=latex,shorten <=2pt] (I\src) to  (Ha\dst);

%
\foreach \src in {1,...,4}
   \foreach \dst in {1,...,5}
     \draw[->,>=latex,shorten <=5pt] (Ha\src) to  (Hm\dst);
%
\foreach \src in {1,...,5}
   \foreach \dst in {1,...,5}
     \draw[->,>=latex,shorten <=2pt] (Hm\src) to  (Hb\dst);
   
%
\foreach \src in {1,...,5}
   \foreach \dst in {1,...,2}
   \draw[->,>=latex,shorten <=2pt] (Hb\src) to  (O\dst);

%
%
\node[latent,right=2 of O1] (s1) {\scriptsize $t_\sfx(x_i)$};
\node[latent,right=2 of O2] (s2) {\scriptsize $n_\sfx(x_i)$};

%
%



%
%
\node[draw, fit=(s1) (s2)] (suffbox) {};

%
%
\node[rectangle, minimum width = 7em,draw,align=center,right=1 of
suffbox] (learnbox)  {\parbox[c][][c]{5em}{\centering $\ell(y_i; T_\sfx(x_i))$\newline
    learning
  }};

\node[rectangle, minimum width = 7em,draw,align=center,above= of
learnbox] (inferbox)  {\parbox[c][][c]{5em}{\centering
    $p(\theta_i \given T_\sfx(x_i))$\newline
    inference
  }};

\node[rectangle, minimum width = 7em,draw,align=center,below= of
learnbox] (evalbox)  {\parbox[c][][c]{5em}{\centering $p(y_i \given T_\sfx(x_i))$\newline
    model eval
  }};

\draw[->,>=latex,shorten <=2pt] (O1) to  (s1);
\draw[->,>=latex,shorten <=2pt] (O2) to  (s2);

\draw[->,>=latex,shorten >=1pt] (suffbox) to [out=10,in=180]  (inferbox);

\draw[->,>=latex,shorten >=1pt] (suffbox) to [out=0,in=180] (learnbox);

\draw[->,>=latex,shorten >=1pt] (suffbox) to [out=-10,in=180] (evalbox);

\begin{pgfonlayer}{background}
\plate[fill=red!30] {nnplate} {(Ha1)(Hb1)(Hb5)(O1)(O2)} {$T(\circ;\phi)$};
\end{pgfonlayer}
\plate {nnplate} {(Ha1)(Hb1)(Hb5)(O1)(O2)} {$T_\sfx(x_i;\phi)$};

\end{tikzpicture}}
  \caption{\small{Detail of NN used for mapping
        (architecture varies by application) transforming
        \textcolor{red!60}{auxiliary} data to sufficient statistics
        used for inference, learning, and model validation.}
    \label{fig:LDF-transformation}}
\end{figure}

The conjugate mapping $T_\sfx(x_i; \phi)$  must be suitably flexible to capture any relevant information about $\theta_i$ from the auxiliary data $x_i$. 
We use NNs to represent the functions $t_\sfx(x_i; \phi)$ and $n_\sfx(x_i; \phi)$, e.g. the network shown in Figure~\ref{fig:LDF-transformation}, but other nonlinear functions may also be used.
Note that this only a notional figure, and that in practice the network outputs $t_\sfx(x_i; \phi)$ and $n_\sfx(x_i; \phi)$ may differ by far more than one linear transformation.
By modern standards, even small networks are adequately expressive to extract relevant information about $\theta_i$ from $x_i$.  
Training such networks is generally straightforward in modern deep learning frameworks (e.g. TensorFlow and PyTorch) that do not require the specification of gradients for many reasonable loss functions and make these models an attractive option even for non-experts.
We describe details of training in Sec.~\ref{sec-learning-and-inference}.

Importantly for learning, the mappings $T_\sfx(x_i; \phi)$ are shared across all $M$ models much like the aggregate sufficient statistics $T_\sfy(y_i) = [t_\sfy(y_i);\, N_i]$ that compresses primary data.
The learned sufficient statistics functions $t_\sfx(x_i; \phi)$ and $n_\sfx(x_i; \phi)$ are directly interpretable as the corresponding quantities for primary data, enabling reasoning over the quality of information contributed by each data source.
Table~\ref{tab-examples} shows the interpretations for several common primary data types.  Additional details can be found in Appendix~A.

\begin{table}[t]
\centering
\caption{{\small Interpretations of the different conjugate mappings of auxiliary data under a selection of common primary data likelihoods.}}
{\small
\begin{tabular}{lll}
\hline 
Primary Data & Conjugate Prior & Auxiliary data interpretation \\ 
\hline 
Bernoulli/binomial & Beta &
number of trials $n_\sfx(x_i; \phi) \geq 0$, \\ &&
success rate $\mu(x_i; \phi) \in [0, 1]$ \\ 
Multinoulli/multinomial & Dirichlet & 
number of trials $n_\sfx(x_i; \phi) \geq 0$, \\ &&
outcome probabilities $\delta(x_i; \phi) \in \Delta^{d-1}$ \\
Poisson & Gamma & 
number of arrivals $a(x_i; \phi) \geq 0$, \\ &&
number of intervals $b(x_i; \phi) \geq 0$ \\
Multivariate Gaussian & NIW &
potential vector $h_\sfy(x_i; \phi) \in \bbR^d$, \\ &&
precision matrix lower triangle $L(x_i; \phi) L(x_i; \phi)^\top \succeq 0$ \\
\hline 
\end{tabular} 
}
\label{tab-examples}
\end{table}

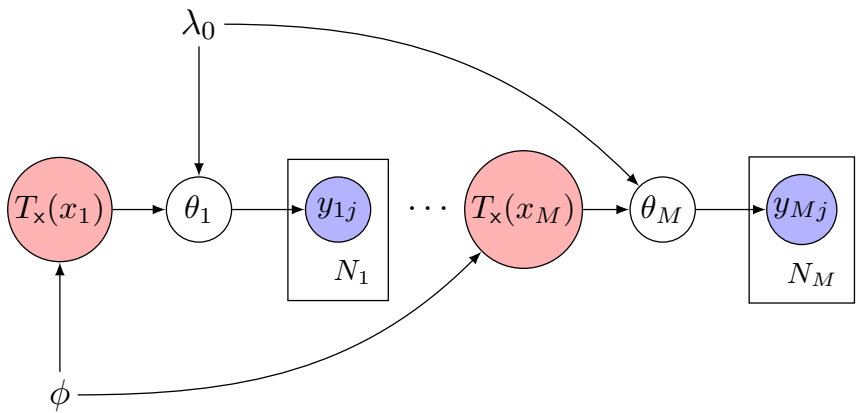
\begin{figure}[t]
  \centering
  \resizebox{0.75\textwidth}{!}{%
%

\begin{tikzpicture}
%
%
\tikzstyle{box}+=[ rounded corners = 5pt,
                       align           = left,
                       font            = \footnotesize,
                       text width      = 2.45cm]

\tikzstyle{every path}+=[thin,>=latex];
\tikzstyle{plate}+=[thin,sharp corners,inner sep=5pt,outer sep=0pt]; 
%
%

%
%
\def \xmncolor {red!70}
\def \thetcolor {cyan!70!black}
\def \zmncolor {yellow!70!black}
\def \betcolor {green!70!black}

%
%
%
\tikzstyle{latent}+=[node distance=1.25 and 1.5, on grid];
\tikzstyle{obs}+=[fill=blue!30,node distance=1.25 and 1.5,on grid];
\tikzstyle{obsaux}=[latent,fill=red!30,node distance=1.25 and 1.5,on grid];
\tikzstyle{obsauximp}=[latent,fill=green!30,node distance=1.25 and 1.5,on grid];
\tikzstyle{const}+=[node distance=1.25 and 1.5,on grid];

%
%

%
%
\node[obsaux] (Ttildx1) {$T_\sfx(x_1)$};
\node[latent,right=1.5 of Ttildx1] (theta1) {$\theta_1$};
\node[obs,right = 1.5 of theta1] (y1j) {$y_{1j}$};
%
%
\edge {Ttildx1}{theta1};
\edge {theta1} {y1j}; 

\plate {yplate1} {(y1j)} {$N_1$}; 

\node[const,right=4 of Ttildx1] (cdots) {$\cdots$};

%
%
\node[obsaux,right=1 of cdots] (TtildxM) {$T_\sfx(x_M)$};
\node[latent,right=1.5 of TtildxM] (thetaM) {$\theta_M$};
\node[obs,right = 1.5 of thetaM] (yMj) {$y_{Mj}$};
%
%
\edge {TtildxM} {thetaM};
\edge {thetaM} {yMj}; 

\plate {yplateM} {(yMj)} {$N_M$}; 

%
%
\node[const,above=2 of theta1] (thetaparam) {$\lambda_0$};
\draw[->,>=latex,shorten <=2pt] (thetaparam) to [out= -90,in=90] (theta1);
\draw[->,>=latex,shorten <=2pt] (thetaparam) to [out=0,in=135] (thetaM);

\node[const,below=2 of Ttildx1] (LDFparam) {$\phi$};
\draw[->,>=latex,shorten <=2pt] (LDFparam) to [out=90,in=-90] (Ttildx1);
\draw[->,>=latex,shorten <=2pt] (LDFparam) to [out=0,in=-135] (TtildxM);

\tikzstyle{latent}+=[densely dashed];
\tikzstyle{plate}+=[color=black];

%
%





\end{tikzpicture}

}
  \caption{\small{PGM incorporating a mapping $T_\sfx(x; \phi) = [t_\sfx(x; \phi),\, N_\sfx(x; \phi)]$ with \textbf{shared} hyperparameters $\phi$ \textit{learned} from jointly observed \textcolor{red!60}{auxiliary} and \textcolor{blue!60}{primary} data instances $x_i, y_i$ generated by diverse and unobserved $\theta_i$. After learning and validation of $\phi$, we are solely interested in the role of  $T_\sfx(x_i; \phi)$ for \textit{posterior} inference over $\theta_i$. Note that LDF mappings are applicable to a variety of PGM structures beyond the model depicted here.}}
   \label{fig:LDF-gmdisc}
\end{figure}

Under the conjugate mappings assumptions, inference in the model of Figure~\ref{fig:LDF-gmgen} can be viewed equivalently as Figure~\ref{fig:LDF-gmdisc}.  The dependence on $x_i$ has been replaced by dependence on $T_\sfx(x_i; \phi)$, made possible by sufficiency as in Figure~\ref{fig:LDF-gmsuffstats}, and the arrow reversed to indicate that we always condition on the conjugate mapping $T_\sfx(x_i; \phi)$ instead of the original data $x_i$.

The form of the posterior $p(\theta_i \given x_i; \phi)$ in Equation~\ref{eq-posterior-trfexpfam} is in the same exponential family as our original conjugate prior $p(\theta_i)$, resulting in convenient forms for two conditionals critical for learning and inference in both simple models (\eg~Fig.~\ref{fig:LDF-gmgen} and Sec.~\ref{sec-electrification}) and more complex hierarchical models where $\gamma \neq \emptyset$, (\eg~Sec.~\ref{sec-homicide}).
Specifically, $p(\theta_i \given x_i; \phi)$ remains conjugate to the likelihood of \textit{primary} data $y_i$, Equation~\ref{eq-likelihood-expfam}, yielding a full posterior
\begin{equation}\label{eq-posterior-full-expfam}
  p(\theta_i \given x_i, y_i; \phi)
    = \pi(\theta_i; \lambda_0 + T_\sfx(x_i; \phi) + T_\sfy(y_i)),
\end{equation}
that is in the same family as Eq.~\ref{eq-prior-expfam} with updated natural parameters $[\tau_0 + t_\sfx(x_i; \phi) + t_\sfy(y_i),\: \nu_0 + n_\sfx(x_i; \phi) + N_i]$.  This quantity will be central to inference over the latent variables $\theta$.
Additionally, for the models where $Z(\tau, \nu)$ is finite we obtain a closed-form conditional distribution $p(y_i \given x_i; \phi)$:
\begin{align}\label{eq-predictive-trfexpfam}
p(y_i \given x_i; \phi)
    &= \int f(y_i; \theta_i) \pi(\theta_i; \lambda_0 + T_\sfx(x_i; \phi))  \dif\theta_i \nonumber\\
    &= h_\sfy(y_i) \dfrac{Z(\tau_0 + t_\sfx(x_i; \phi) +  t_\sfy(y_i), \nu_0 + n_\sfx(x_i; \phi) + N_i)}{Z(\tau_0 + t_\sfx(x_i; \phi), \nu_0 + n_\sfx(x_i; \phi))}
\end{align}
that has the same form as the marginal $p(y_i)$ and posterior predictive (Eqs.~\ref{eq-marginal-expfam}, \ref{eq:postpredict}) with updated parameters $[\tau_0 + t_\sfx(x_i; \phi),\: \nu_0 + n_\sfx(x_i; \phi)]$.  As with the posterior we can equivalently write $p(y_i \given x_i; \phi) = p(y_i \given T_\sfx(x_i; \phi))$ by sufficiency.
This convenient form for the primary data posterior predictive, a strikingly simple function of the mappings $t_\sfx(x_i; \phi)$ and $n_\sfx(x_i; \phi)$, plays a central role in learning the mapping parameters $\phi$.

\subsection{Conjugate Mappings in More Complex Models}

Our modeling approach, and the subsequent learning and inference methods, extend naturally to more complex models with global structure.
For models with global variables $\gamma \sim p(\gamma)$, the remainder of our model is
\begin{align}
p(\theta_i \given \gamma, x_i; \phi)
    &= \pi(\theta_i; \lambda_i(\gamma) + T_\sfx(x_i; \phi)),  &
p(y_i \given \theta_i)
    &= f(y_i; \theta_i), &
    & i = 1, \ldots, M
\end{align}
where $\pi(\cdot)$ and $f(\cdot)$ are conjugate and the functions $\lambda_i(\gamma) = [\tau_i(\gamma);\, \nu_i(\gamma)]$ relate the natural parameters of $\theta_i$ to the global variables.

The conditioning and marginalization properties from the prior subsections carry over to more complex model structures.  The distributions over $\theta_i$ remain in the same exponential family distribution $\pi(\cdot)$ \textit{regardless} of the conditioning set:
\begin{align}
p(\theta_i \given \gamma)
    &= \pi(\theta_i; \lambda_i(\gamma)) \\
p(\theta_i \given \gamma, x_i; \phi)
    &= \pi(\theta_i; \lambda_i(\gamma) + T_\sfx(x_i; \phi))  \\
p(\theta_i \given \gamma, x_i, y_i; \phi)
    &= \pi(\theta_i; \lambda_i(\gamma) + T_\sfx(x_i; \phi) + T_\sfy(y_i))
\end{align}
where $T_\sfy(y_i) = [t_\sfy(y_i);\, N_i]$ and $i = 1, \ldots, M$ as before.
Additionally, we can still marginalize out $\theta_i$ in closed-form for the primary data posterior predictive
\begin{align}
p(y_i \given \gamma, x_i; \phi) 
    &= \int f(y_i; \theta_i) \pi(\theta_i; \lambda_i(\gamma) + T_\sfx(x_i; \phi)) \dif\theta_i \\
    &= h_\sfy(y_i) \dfrac{Z(\tau_i(\gamma) + t_\sfx(x_i; \phi) +  t_\sfy(y_i), \nu_i(\gamma) + n_\sfx(x_i; \phi) + N_i)}{Z(\tau_i(\gamma) + t_\sfx(x_i; \phi), \nu_i(\gamma) + n_\sfx(x_i; \phi))},
\end{align} 
which now depends on the global latent variables $\gamma$.


\section{Learning and Inference with Conjugate Mappings}
\label{sec-learning-and-inference}

\begin{algorithm}[t]
\KwIn{primary data $y = \{y_i\}_{i=1}^M$, auxiliary data $x = \{x_i\}_{i=1}^M$, model $\calM$, objective $\calL$}
\KwOut{mapping parameters $\hat{\phi}$; posterior $p(\theta, \gamma \given x, y; \hat{\phi})$}
Learn mapping parameters $\hat{\phi} = \arg\max_\phi \calL(\{p(y_i \given T_\sfx(x_i; \phi))\}_{i=1}^M)$ \\
Compute auxiliary data sufficient statistics $T_\sfx \gets \{T_\sfx(x_i; \hat{\phi})\}_{i=1}^M$ \\
Compute primary data sufficient statistics $T_\sfy \gets \{T_\sfy(y_i)\}_{i=1}^M$ \\
Infer posterior $p(\theta, \gamma \given x, y; \hat{\phi}) \gets \textsc{ApproximateInference}(\calM, T_\sfx, T_\sfy)$
\caption{\textsc{LightweightDataFusion}: a learning and inference procedure returning the posterior over global latent variables $\gamma$ and local latent variables $\theta = (\theta_1, \ldots, \theta_M)$ and a point estimate $\hat{\phi}$ of conjugate mapping parameters.}
\label{alg-learning-and-inference}
\end{algorithm}

The LDF methodology, described in Algorithm~\ref{alg-learning-and-inference}, provides a framework for posterior inference $p(\theta, \gamma \given x, y)$ where $p(x_i \given \theta_i)$ is unknown. 
The conjugate mappings technique of Section~\ref{sec-mappings} reduces this to the task of learning a suitable conjugate mappings function $T_\sfx(x_i; \phi)$ and then doing inference:
\begin{align}
    p(\theta, \gamma \given x, y; \phi)
        \propto p(\gamma) \prod_{i=1}^M p(\theta_i \given \gamma, T_\sfx(x_i; \phi)) p(y_i \given \theta_i).
    \label{eq-fusion-mappings}
\end{align}
LDF treats the transformation parameters $\phi$ as hyperparameters and learns a point estimate $\hat{\phi}$ as discussed in Sec.~\ref{sec-general-learning}.
Aggregated sufficient statistics are computed from the primary data $\{T_\sfy(y_i)\}_{i=1}^M$ and auxiliary data $\{T_\sfx(x_i; \hat{\phi})\}_{i=1}^M$ using the learned value of $\hat{\phi}$.
Posterior inference subsequently calculates the posterior $p(\theta, \gamma \given x, y; \hat{\phi})$ using the primary and auxiliary data sufficient statistics as detailed in Sec.~\ref{sec-general-inference}.
LDF's novelty is in mapping \textit{auxiliary} data into aggregated sufficient statistics that facilitate efficient conditioning.
Converting primary data to sufficient statistics is standard practice for inference in exponential families.

We denote distributions that depend only on the sufficient statistics of the auxiliary data $T_\sfx(x_i; \phi)$ rather than the original data $x_i$ explicitly to emphasize dependence on mapping parameters $\phi$, e.g. $p(\theta_i \given T_\sfx(x_i; \phi))$ and $p(y_i \given T_\sfx(x_i; \phi))$ rather than $p(\theta_i \given x_i; \phi)$ and $p(y_i \given x_i; \phi)$.

\subsection{Learning Conjugate Mappings}
\label{sec-general-learning}

We begin by discussing learning conjugate mapping parameters in models with no global structure ($\gamma = \emptyset$); we subsequently provide strategies for learning in more complex model structures.
The optimization problem
\begin{align}
  \hat{\phi} 
  &= \arg \max_{\phi} \sum_{i=1}^M w_i \calL(y_i; T_\sfx(x_i; \phi))
  \label{eq:GenBayesLoss}
\end{align}
formulates a method for learning the mapping parameters $\phi$.  Weights $w_i$ specify the contribution of model $i$ in the sum and $\mathcal{L}(y_i; T_\sfx(x_i; \phi))$ is \textit{any} general Bayesian objective function providing a score of the observed primary data $y_i$ parameterized by transformed auxiliary data $T_\sfx(x_i; \phi)$.
We omit an explicit regularization term from our objective but typically use an L2 penalty on the mapping parameters $\phi$.
In large data regimes it suffices to optimize an \textit{unbiased estimator} of the objective in Eq.~\ref{eq:GenBayesLoss}, as in stochastic optimization methods \citep{duchi2011adaptive, tieleman2012lecture, kingma2014adam}.

Here we consider the broad class of objective criteria that are functions $\ell(\cdot)$ of the primary data posterior predictive log-likelihood $\log p(y_i \given T_\sfx(x_i; \phi))$
\begin{align}
    \calL(y_i; T_\sfx(x_i; \phi)) = \ell(\log p(y_i \given T_\sfx(x_i; \phi))).
    \label{eq-general-loss-objective}
\end{align}
This ensures that information extracted from the auxiliary data $x_i$ about the primary data $y_i$ is mediated by their relation to the latent variable $\theta_i$, as evident from 
\begin{align}
    p(y_i \given T_\sfx(x_i; \phi)) 
    = \int p(y_i \given \theta_i) p(\theta_i \given T_\sfx(x_i; \phi)) \dif\theta_i.
\end{align}

The maximum-likelihood (ML) criterion
\begin{align}
   \calL_\mathrm{ML}(y_i; T_\sfx(x_i; \phi)) 
   \triangleq \log p(y_i \given T_\sfx(x_i; \phi))
   \label{eq-loss-maximum-likelihood}
\end{align}
is a natural choice of learning objective with appealing asymptotic properties. It is widely understood that
one can interpret the ML criterion as learning a transformation $T_\sfx(x_i; \phi) \approx T_\sfx^{*\!}(x_i)$ that makes the auxiliary data as informative as possible about the primary data $y_i$ under the model.
To see this, consider the data as a set of i.i.d. samples from the true joint distribution $\{x_i, y_i\}_{i=1}^M \sim p^{*\!}(x, y)$ under the reduced model setting, $\gamma = \emptyset$ and uniform weights $w = (1/M, \ldots, 1/M)$.  Then
\begin{align}
    \hat{\phi}
    &= \arg\max_\phi \lim_{M\rightarrow\infty} \dfrac{1}{M} \sum_{i=1}^M \log p(y_i \given T_\sfx(x_i; \phi)) \\
    &= \arg\max_\phi \bbE_{p^{*\!}(x, y)}\!\left[ \log p(y \given T_\sfx(x; \phi)) \right] &     &\longleftarrow  -\bbH_{p^{*\!}}(p(y \given T_\sfx(x; \phi)) \label{eq-learning-entropy} \\
    &= \arg\max_\phi \bbE_{p^{*\!}(x, y)}\!\left[ \log p(y \given T_\sfx(x; \phi)) - \log p(y) \right] &
        \\
    &= \arg\max_\phi \bbE_{p^{*\!}(x, y)}\!\left[ \log \dfrac{p(y \given T_\sfx(x; \phi))}{p(y)} \right] &
        &\longleftarrow  \mathrm{I}_{p^{*\!}}(y; T_\sfx(x; \phi)). \label{eq-learning-mi}
\end{align}
Eq.~\ref{eq-learning-entropy} shows that maximizing the ML objective minimizes the entropy of the posterior predictive distribution; Eq.~\ref{eq-learning-mi} shows that it also maximizes the mutual information between the primary data and the transformed auxiliary data.  It is additionally straightforward to show from (Eq.~\ref{eq-learning-entropy}) that maximizing the ML objective minimizes the KL divergence of the primary data posterior predictive from the true (unknown) distribution $p^{*\!}(y \given x)$: $\KL(p^{*\!}(y \given x) \:\|\: p(y \given T_\sfx(x; \phi)))$---simply add $\log p^{*\!}(y \given x) - \log p^{*\!}(y \given x)$ to the argument of the expectation and reorganize terms appropriately.

\begin{figure}[t]
  \vskip 0.1in
  \begin{center}
    \centerline{
    \includegraphics[width=\linewidth, trim={0 0 0 0}, clip]{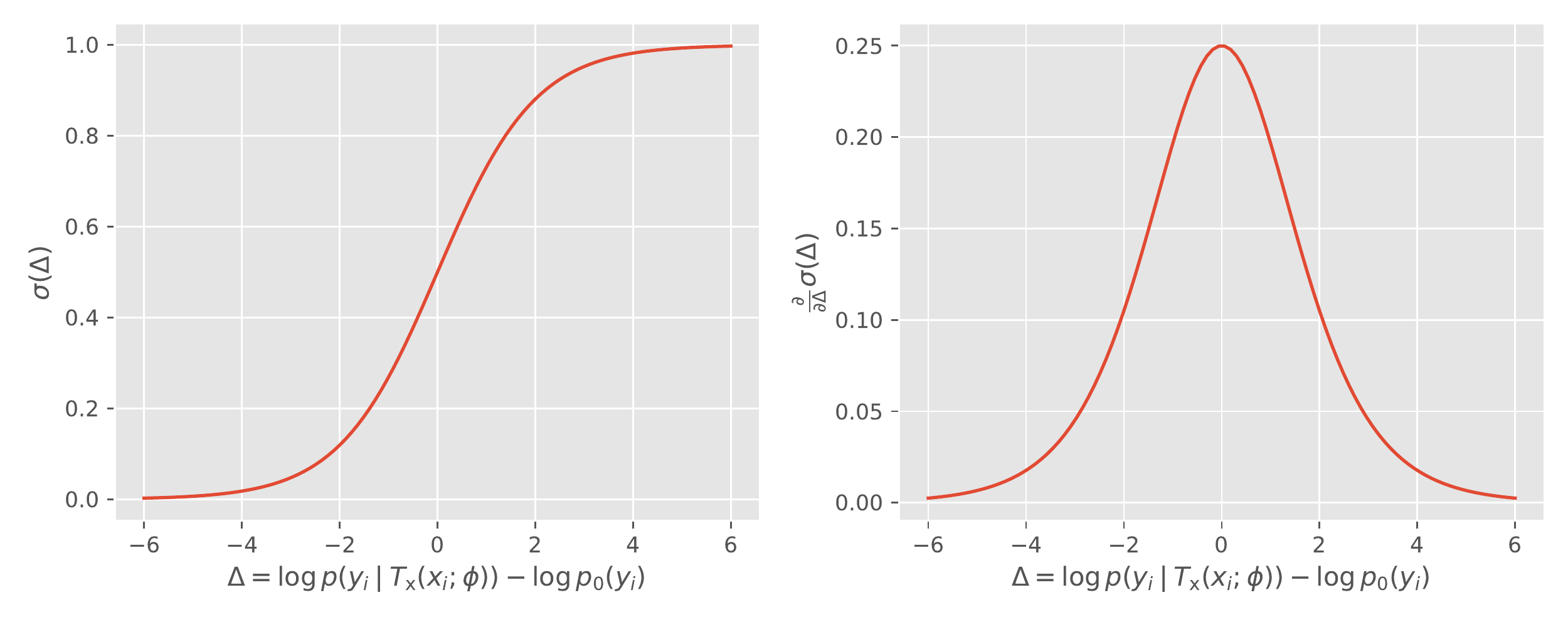}    
    }
    \caption{\small \textbf{Model selection objective} in terms of the difference $\log p(y_i \given T_\sfx(x_i; \phi)) - \log p_0(y_i)$ between the learned model and a reference model $P_0$.  \textit{(Left)} the objective function value saturates for extreme values of the difference; \textit{(Right)} the derivative of the objective w.r.t. the difference between the two models.
    }
    \label{fig-model-selection-loss}
  \end{center}
  \vspace*{-5mm}
\end{figure}

Despite its appealing properties there are potential issues with the ML criterion.  
This objective emphasizes \textit{average} performance across $M$ models and can thus yield a solution that arbitrarily increases the likelihood for some instances to the detriment of others. 
While proper weighting and regularization can mitigate such effects, an alternative criterion encodes this explicitly (and doesn't preclude the use of either as appropriate).
Consider a model selection (MS) criterion that maximizes the probability (on average) that the learned model $p(y_i \given T_\sfx(x_i; \phi))$ is a better explanation for the primary data $y_i$ than a \textit{reference model} $p_0(y_i)$:
\begin{align}
    \calL_\mathrm{MS}(y_i; T_\sfx(x_i; \phi))
        & \triangleq \frac{p(y_i \given T_\sfx(x_i; \phi))}{p(y_i \given T_\sfx(x_i; \phi)) + p_0(y_i)} \nonumber\\ 
        & = \frac{1}{1+\exp\left\{ - (\log p(y_i \given T_\sfx(x_i; \phi)) - \log p_0(y_i)) \right\}} \nonumber\\
        & = \sigma\!\left(\log p(y_i \given T_\sfx(x_i; \phi)) - \log p_0(y_i))\right)
            \label{eq-loss-model-selection}
\end{align}
where $\sigma(\cdot)$ is the logistic sigmoid function.
This reference model $p_0(y_i)$ could be derived from the prior distribution over $\theta_i$ or learned over a reduced-order model class.
The MS objective and its derivative are shown in Figure~\ref{fig-model-selection-loss} as a function of the log difference $\log p(y_i \given T_\sfx(x_i; \phi)) - \log p_0(y_i)$.
This objective clearly saturates as $\log p(y_i \given T_\sfx(x_i; \phi))$ exceeds $\log p_0(y_i)$, leading to a solution that emphasizes more \textit{uniform} performance across the $M$ different models rather than the \textit{average} performance emphasized by the ML criterion.
This helps avoid reaching potentially degenerate states of the parameter space, e.g. transformations that can increase the posterior predictive likelihood of a specific training example $y_i$ arbitrarily high by increasing the number of pseudo-observations $n_\sfx(x_i; \phi)$.
Other choices of suitable objective criteria exist and may be motivated by the application domain and the structure of the problem.

Regardless of the choice of objective criteria, the gradients are computed via back-propagation within TensorFlow or PyTorch and do not need to be derived to implement our approach.
However, the objective function gradients provide insights about models learned under the ML and MS criteria.  
For objective criteria taking the form of Equation~\ref{eq-general-loss-objective} we have
\begin{align*}
\nabla_{\!\phi} \calL(y_i; T_\sfx(x_i; \phi))
    &= \nabla_{\!\phi} T_\sfx(x_i; \phi) 
    \cdot \nabla_{\!T_{\sfx}} \log p(y_i \given T_\sfx(x_i; \phi)) 
    \cdot \nabla_{\!\log p} \ell(\log p(y_i \given T_\sfx(x_i; \phi)))
\end{align*}
where the gradients (from left to right) are the gradient of the aggregated sufficient statistics, the gradient of the posterior predictive log-likelihood, and the gradient of the objective.
For the ML criteria (Equation~\ref{eq-loss-maximum-likelihood}) the gradient of the objective for the $i$-th instance $(x_i, y_i)$ is
\begin{align}
\nabla_{\!\log p} \ell_\mathrm{ML}(\log p(y_i \given T_\sfx(x_i; \phi))
    &= 1
\end{align}
which contributes equally to the full gradient \textit{regardless} of how well the primary data is fit by the model.  In contrast, the gradient of the objective for the MS criteria (Equation~\ref{eq-loss-model-selection}) is
\begin{align}
\nabla_{\!\log p} \ell_\mathrm{MS}(\log p(y_i \given T_\sfx(x_i; \phi)))
    &= \dfrac{
    \exp\left\{-\left[\log p(y_i \given T_\sfx(x_i; \phi)) - \log p_0(y_i)\right]\right\}}{
    \left(1 + \exp\left\{-\left[\log p(y_i \given T_\sfx(x_i; \phi)) - \log p_0(y_i)\right]\right\}\right)^2}
\end{align}
which is shown in Figure~\ref{fig-model-selection-loss} (right) in terms of the difference $\log p(y_i \given T_\sfx(x_i; \phi)) - \log p_0(y_i)$ between the learned model and a reference model $p_0(y_i)$.  Instances $(x_i, y_i)$ where one model clearly outperforms the other contribute negligibly to the full gradient and minimally impact the optimization procedure.  Instead, optimization is guided by those instances where the LDF model $p(y_i \given T_\sfx(x_i; \phi))$ performs similarly to the reference model $p_0(y_i)$ and increases its objective by making $T_\sfx(x_i; \phi)$ more informative about the primary data.

For both the ML and MS criteria the ability to calculate the primary data posterior predictive likelihood $p(y_i \given T_\sfx(x_i; \phi)) $ is paramount.
Under the simple model structure of Figure~\ref{fig:LDF-gmdisc} it can be computed in closed form using Eq.~\ref{eq-predictive-trfexpfam}.
Under the general form of Equation~\ref{eq-fusion-mappings} where $\gamma \neq \emptyset$ this likelihood requires marginalization of the global variables $\gamma$:
\begin{align}
p(y_i &\given T_\sfx(x_i; \phi))
    = \int_\gamma p(\gamma) \int_{\theta_i} p(\theta_i \given \gamma, T_\sfx(x_i; \phi)) p(y_i \given \theta_i) \dif\theta_i \dif\gamma \\
    &= h(y_i) \int_\gamma p(\gamma) \left(\dfrac{Z(\tau_i(\gamma) + t_\sfx(x_i; \phi) +  t_\sfy(y_i), \nu_i(\gamma) + n_\sfx(x_i; \phi) + N_i)}{Z(\tau_i(\gamma) + t_\sfx(x_i; \phi), \nu_i(\gamma) + n_\sfx(x_i; \phi))}\right) \dif\gamma
\end{align}
where the inner integral is computable in closed form and takes the same form as the prior predictive.  The outer integral can be approximated without overly complicating the optimization w.r.t. $\phi$.  For example, when approximating marginalization of $\gamma$ using Monte Carlo integration
\begin{align}
p(y_i &\given T_\sfx(x_i; \phi))
    \approx h(y_i) \dfrac{1}{S} \sum_{s=1}^S \dfrac{Z(\tau_i(\gamma^{(s)}) + t_\sfx(x_i; \phi) +  t_\sfy(y_i), \nu_i(\gamma^{(s)}) + n_\sfx(x_i; \phi) + N_i)}{Z(\tau_i(\gamma^{(s)}) + t_\sfx(x_i; \phi), \nu_i(\gamma^{(s)}) + n_\sfx(x_i; \phi))}
\end{align}
the samples $\gamma^{(s)}$ for $s = 1, \ldots, S$ may be reused across multiple iterations of gradient descent since their sampling distribution $p(\gamma)$ does not depend on $\phi$. 
Since Monte Carlo integration provides a consistent estimator the overall objective (Eq.~\ref{eq:GenBayesLoss}) is also consistent.
We demonstrate an alternative approach, based on the expectation-maximization (EM), in Section~\ref{sec-homicide}.

\subsection{Latent Variable Posterior Inference}
\label{sec-general-inference}

We now consider posterior inference of $p(\gamma, \theta \given x, y; \phi)$ conditioning on the learned model parameters $\hat{\phi}$.  This generally requires approximate inference, i.e. MCMC or variational methods.  
We show that significant simplifications arise under the construction of Section~\ref{sec-mappings} for sampling-based inference procedures.  Similar properties extend to coordinate-ascent variational inference methods.
In both cases, inference relies not on the original data $x_i$ and $y_i$, but on the aggregate sufficient statistics $T_\sfy(y_i) = [t_\sfy(y_i);\, N_i]$ and $T_\sfx(x_i; \hat{\phi}) = [t_\sfx(x_i; \hat{\phi});\, n_\sfx(x_i; \hat{\phi})]$ for $i = 1, \ldots, M$.

For now, consider Gibbs sampling under the general model where $\gamma \neq \emptyset$.  Inference requires sampling from the complete conditional distributions for each latent variable.  For $\theta_i$, $i = 1, \ldots, M$, this is
\begin{align}
p(\theta_i &\given \theta_{\setminus i}, \gamma, x, y; \phi)
    = p(\theta_i \given \gamma, x_i, y_i; \phi) \nonumber\\
    &\propto p(\theta_i \given \gamma, T_\sfx(x_i; \phi)) p(y_i \given \theta_i) \nonumber\\
    &\propto \pi(\theta_i \given \tau_i(\gamma) + t_\sfx(x_i; \phi) + t_\sfy(y_i), \nu_i(\gamma) + n_\sfx(x_i; \phi) + N_i) \label{eq-gibbs-conditional-theta}
\end{align}
which is the full posterior (Eq.~\ref{eq-posterior-full-expfam}) from Section~\ref{sec-mappings}, except that the hyperparameters $\tau_0$ and $\nu_0$ have been replaced with model-$i$-specific functions $\tau_i(\gamma)$ and $\nu_i(\gamma)$ on account of the global structure.  The distributions (Equation~\ref{eq-gibbs-conditional-theta}) are independent across the $\theta_1, \ldots, \theta_M$, so they can be sampled in parallel for efficient inference.
For the global latent variables $\gamma$, the complete conditional is
\begin{align}
p(\gamma &\given \theta, x, y; \phi) = p(\gamma \given \theta)
\label{eq-gibbs-conditional-beta}
\end{align}
since $\gamma$ is conditionally independent from the data $x$ and $y$ given $\theta$.

If $\gamma = \emptyset$ then sampling Equation~\ref{eq-gibbs-conditional-beta} and Equation~\ref{eq-gibbs-conditional-theta} can be skipped entirely, as the full posterior factorizes as
\begin{align}
p(\theta \given x, y; \phi)
     = \prod_{i=1}^M p(\theta_i \given x_i, y_i; \phi)
\end{align}
where the individual factors have the form of Equations~\ref{eq-gibbs-conditional-theta}~and~\ref{eq-posterior-full-expfam} with natural parameters computable in closed-form.

It is now clear how our LDF method results in \textit{lightweight} posterior inference w.r.t. the auxiliary data.
Given the sufficient statistics $T_\sfx(x_i; \hat{\phi})$ learned from the primary data, full posterior inference $p(\gamma, \theta \given x, y; \hat{\phi})$ is clearly no more computationally-demanding than inference based on primary data only, $p(\gamma, \theta \given y)$.
The computational complexity of primary data posterior inference $p(\theta, \gamma \given y)$ is dominated by the interior dependence structure of the global latent variables $\gamma$, along with their relationship to the $\{\theta_i\}_{i=1}^M$.
This is apparent for Gibbs-sampling-based inference in Equation~\ref{eq-gibbs-conditional-beta}.
The relative cost of updating the natural parameters of the local latent variables $\theta_i$ with the corresponding sufficient statistics $T_\sfy(y_i)$ and $T_\sfx(x_i; \hat{\phi})$ is comparatively low.
The main computational cost of our approach is in the process of learning the conjugate mapping parameters $\hat{\phi}$.
Alternative approaches, e.g. those which formulate a non-conjugate likelihood model of $p(x_i \given \theta_i)$, would significantly increase the cost of doing inference by precluding the use of conjugate updates via sufficient statistics.


\section{Related Work}
\label{sec-related}

LDF addresses a perennial problem in statistical data analysis: making use of data from uncalibrated or poorly-understood sources that we denote as auxiliary data $x$.
The existence of primary data $y$ is a central feature of our setting and one that invites comparisons to a variety of existing methods.
Classical methods view auxiliary data $x$ and primary data $y$ as \textit{covariates} and \textit{response variables}, respectively.
Transforming covariates to better match statistical assumptions, e.g. through basis expansions, have long been explored~\citep{Hastie2011}, but typically require extensive hand-tuning of features.  
Data whitening techniques, e.g. PCA or factor analysis, avoids feature tuning but implicitly assumes Gaussianity~\citep{Bishop2006}.

Under the maximum-likelihood loss function, LDF bears similarities to generalized linear models (GLMs) that combine nonlinear transformations with exponential families~\citep{Nelder1972, Mccullagh1989}, however, there are significant differences.
For one, the class of transformations considered in GLMs is more limited, amounting to a linear transformation composed with a single non-linearity (link function) that provides a scalar output. GLMs using canonical link functions learn the natural parameter of the response variable likelihood.
Equivalently we can view GLMs as learning the \textit{sufficient statistics} $t_\sfx(x_i; \phi)$ from a set of prior observations.
In contrast, LDF learns significantly richer and more expressive transformations of the auxiliary data owing to its use of flexible NN mappings, which include the GLM transformation as a special case.
GLM mappings can be recovered from an LDF-style NN mapping with zero hidden layers whose output activation function matches the GLM's link function.

A more subtle yet critical difference reflects how latent variables are interpreted under each model.  Consider the LDF learning under the ML objective (with minimal model structure and uniform weights), 
\begin{align}
    \log p(y \given x) 
        = \sum_{i=1}^M \log p(y_i \given x_i; \phi)
        = \sum_{i=1}^M \log \int p(y_i \given \theta_i) p(\theta_i \given x_i; \phi) \dif\theta_i,
\end{align}
which is equivalent to standard GLM learning objectives with a few caveats.  
While both models share the primary data likelihood $p(y_i \given \theta_i)$, the underlying latent variable model for LDF is fundamentally different from the notion of latent variables in GLMs. 
LDF transforms the auxiliary data $x_i$ into parameters of the distribution $p(\theta_i \given x_i; \phi)$ whereas GLMs predict a \textit{degenerate} distribution centered at $\hat{\theta}(x_i; \phi)$ that is a nonlinear transformation of $\phi^\top x_i$.
Unlike in GLMs, the marginal $p(y_i \given x_i; \phi)$ is \textit{not} in the exponential family for LDF models due to a non-degenerate representation of $p(\theta_i \given x_i; \phi)$ that admits uncertainty about $\theta_i$.
Aside from this interpretation latent variables are completely absent from standard GLMs, in addition to any further hierarchical latent variable structure.  
GLM latent variables typically refer to the parameters of the underlying linear transformation $\phi^\top x_i$, which have uninformative or Gaussian priors, and are akin to LDF's conjugate mapping parameters.


Other closely-related regression models, e.g. beta-binomial and negative-binomial regression, account for cases where the observed variance in the response variable are over-dispersed relative to their likelihood $p(y_i \given \theta_i)$.
These approaches learn a dispersion parameter shared by all data instances, in addition to the linear transformation parameters from the underlying GLM.
This implicitly defines non-degenerate posteriors $p(\theta_i \given x_i; \phi)$ where the dispersion parameter controls \textit{how} the distribution concentrates about $\hat{\theta}(x_i; \phi)$.
Under this framing, many GLMs and related approaches can be shown to be special cases of LDF. 
LDF can express the corresponding GLM or over-dispersed variant for certain choice of conjugate mapping   $T_\sfx(x_i; \phi) = [t_\sfx(x_i; \phi);\,n_\sfx(x_i; \phi)]$: the over-dispersed variant is obtained for a single shared number of pseudo-observations $n_\sfx(x_i; \phi)$ is learned for all instances; the standard GLM is achieved in the limit as $n_\sfx(x_i; \phi) \rightarrow \infty$.
This holds for both applications, Section~\ref{sec-electrification} and \ref{sec-homicide}, with details located in the corresponding appendices.

Modern generative deep learning methods learn latent variable representations encoded as NN architectures dating as far back as~\citep{neal1990learning, frey1999variational}.
These are often applied in cases where the data generating process is unknown, e.g. deep latent Gaussian models/variational autoencoders (VAEs)~\citep{Rezende2014, Kingma2014}, generative adversarial networks (GANs)~\citep{Goodfellow2014}, deep exponential families~\citep{Ranganath2015}, and exponential family embeddings~\citep{Rudolph2016}, among others~\citep{edwards2016towards, Li2018, Tran2019}.  
However, the resulting latent representations lack interpretability and require complex training procedures, in contrast to LDF.

LDF is most closely related to \textit{structured VAEs} (SVAEs)~\citep{Johnson2016}, but with substantive differences. 
Both VAEs and SVAEs propose a generative model where the data are nonlinear transformations of the latent space.
Training learns a pair of nonlinear transformations: an \textit{encoder} that maps a data realization into the latent space and a \textit{decoder} that maps a latent state into data space.
For SVAEs the latent space is structured as a PGM where all distributions are in the exponential family and conditionally-conjugate.  
The encoder learns a mapping subject to a surrogate likelihood form that results in a conjugate potential for inference.
LDF, by contrast, learns a \textit{single} nonlinear transformation of the data into aggregate sufficient statistics that parameterize a closed-form posterior in the exponential family. 
This leads to easily-tractable approximate inference procedures, even in more complex models.
Unlike in SVAEs, outside of the distributions concerning $(y_i, \theta_i)$ the PGM structures represented in LDF do not have any required form and are \textit{not} constrained to be in the exponential family.

Another critical difference is that SVAEs treat \textit{all} of the data as auxiliary data $x$ from our treatment.
Incorporating primary data with known exponential family likelihood into the SVAE framework requires a non-trivial extension and care to ensure the resulting model reflects the intended relationship between the local latent variables $\theta_i$, the primary data $y_i$, and the auxiliary data $x_i$. 
The impact of primary data $y$ on the ability to learn and make inferences about the latent space cannot be overstated.
Primary data ensures that LDF learns posteriors of the underlying probabilistic model that are interpretable and meaningful w.r.t. the data and in the context of the PGM.
Without primary data, training can learn effectively meaningless latent variable posteriors and a suitably complex data likelihood (decoder) that still manages to adequately model the data; such posteriors would ultimately fail at any downstream reasoning tasks.
In LDF the primary data directly conditions on latent variables in the PGM:  this ensures that posterior learning and inference can model the primary data $y$ in a manner orthogonal to the conjugate mapping of auxiliary data $x$.
Finally, the primary data $y$ dramatically reduces the difficulty of the learning problem.
SVAEs require a large amount of data for what amounts to unsupervised learning of the encoder and decoder models.
LDF instead learns the conjugate mapping parameters for the posterior $p(\theta_i \given T_\sfx(x_i;\phi))$ using the primary data as noisy labels with a known noise pattern, $p(y_i \given \theta_i)$.
This aspect of LDF learning enables its application in much smaller data regimes than SVAEs.

Finally, the LDF methodology produces full probability distributions of the latent variables, rather than simple point estimates.
While NN outputs are often interpreted as probability distributions (e.g. softmax activations used in classification problems), such distributions tend to provide poorly-calibrated estimates of uncertainty.
Most existing literature on uncertainty quantification in the context of deep networks generally concerns \textit{parameter} uncertainty \citep{Kendall2017, Gal2016} and synthesis (e.g. GANs), rather than the latent variables of interest.
In contrast, LDF enables uncertainty quantification with respect to \textit{latent} variables in a way that leverages the power of NNs and preserves many of the elegant properties of Bayesian models.


\section{Case Study: Electrification Access Analysis}
\label{sec-electrification}

We demonstrate the analyses enabled by LDF on the example problem of inferring granular rates of electricity access in developing nations.  
Obtaining accurate information about electrification at a local level is critical for both governments and NGOs.
At spatially-granular levels well-characterized primary data---direct measures of electrification status---are rare and acquiring large-scale high-fidelity data is cost-prohibitive \citep{Leesj2018}.  
Auxiliary data of phenomena \textit{related} to electrification, but which lack a well-characterized statistical model, are comparatively cheap and ubiquitous.   The auxiliary data used here are more commonly accessible and are indicative of the types, variability, and quality of data available for many developing nations.  
The stochastic nature of the limited primary data, combined with the uncalibrated nature of the auxiliary data and the importance of accurately characterizing prediction uncertainty, preclude the use of many methods brought to bear on similar problems, e.g.~\citep{Jean2016}.

\begin{figure}[t!]
  \vskip 0.1in
  \begin{center}
    \centerline{
    \includegraphics[width=\linewidth, trim={0cm 0cm 0cm 0cm},clip]{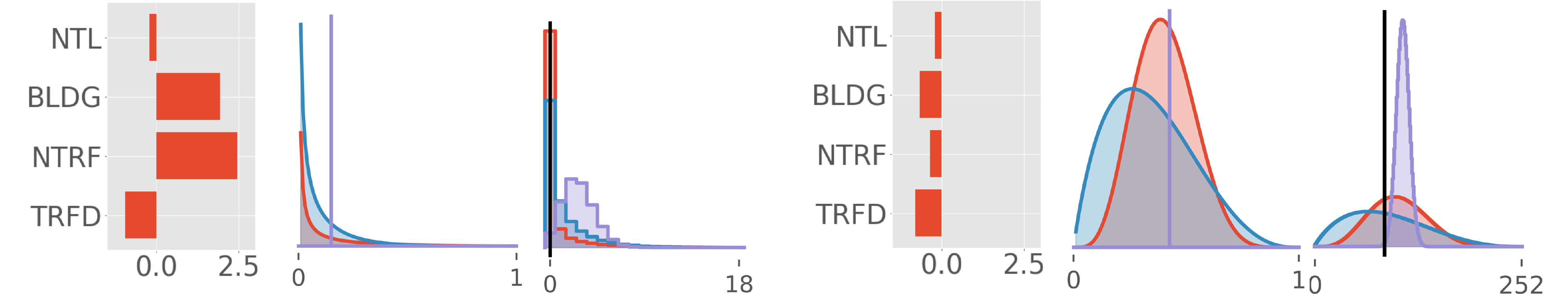}
    }
    \caption{\small \textbf{Features and inferences for Kayonza.}  Inferences show LDF as red, beta-binomial regression as blue, and binomial regression as purple. (Left Three) A region in the Kayonza district with relatively low electrification probability;  (Right Three) a region of Kayonza relatively high electrification probability.   (Inner; Left to Right) the feature values [night-time lights, building density, transformer density, and transformer distance]; the posterior $p(\theta_i \given x_i; \phi)$ for LDF and the baselines; and the posterior predictive distributions $p(y_i \given x_i; \phi)$ with the observed value shown in black.
    }
    \label{fig-elec-posterior-breakout}
  \end{center}
  \vspace*{-5mm}
\end{figure}

Beyond the simple predictive capabilities of standard regression approaches, our LDF approach enables analyses including direct inspection of the posteriors and any derived statistics, posterior predictive checking, and assessing model calibration.
Figure~\ref{fig-elec-posterior-breakout} shows a variety of facets of the analysis pipeline.  Columns 1 and 4 show the different auxiliary features for two different local regions.  The electrification rate posteriors $p(\theta_i \given x; \phi)$ (columns 2 and 5) are directly accessible under our model and support uncertainty-aware downstream decision making.  Finally, the posterior predictive distributions $p(y_i \given x_i; \phi)$ for two regions (columns 3 and 6) enable model checking and calibration analysis.

\subsection{Data}

We produce results for the nearly 2,000 $\kilometers^2$ Kayonza district of Rwanda that had a 16.1\% electrification rate as of 2012  \citep{nisr2014rwanda}. 
We divide the Kayonza district into a grid of 2,236 local regions, each of size $30$ arc-seconds $\times$ $30$ arc-seconds (approximately $1$ km $\times$ $1$ km).  The electrification rate within each region is characterized a latent r.v. characterizing the probability that a randomly-selected building in the region has electricity access.  
In general the primary data are of limited availability are used for training an auxiliary data model that is applicable to regions \textit{lacking} primary data.

\textbf{Primary data} are comprised of well-characterized observations related to the local electrification probabilities $\theta_i$.
While in general these are challenging to obtain at the desired granularity, for the Kayonza district we derive them from two distinct data sources.
The first data source is geo-referenced low-voltage distribution line data from Energy Development Corporation Limited (EDCL), Rwanda; the data set was considered 90\% complete for Kayonza as of 2016 \citep{eucl2020}.
The second data source, also provided by EDCL, was produced in 2013 by acquiring orthophotos for the whole country of Rwanda and then manually identifying 1,704,749 buildings across the country \citep{Sofreco2013}. The data set was released in 2013 using orthophotos taken between 2010 and 2011.

Of the 2,236 regions only $M=854$ have associated primary data from these two sources.  
Primary data for each region consist of the number of \textit{electrified} buildings and the \textit{total} number of buildings and are shown in Appendix~C.
Informed by domain specialists, for our purposes a building is considered to have electricity access if it is within a certain distance of the low-voltage power network \citep{garcia2019reg}.
Standard practice is to treat the number of electrified buildings as a realization of a binomial random variable with unknown probability \citep{andrade2019household}.
Stochasticity arises from errors in both the building extraction process and in the binariziation of power status based on distance to the low-voltage lines.

\textbf{Auxiliary data} are available for all regions from various remote information sources.  For each region, these are comprised of the following measurements whose relationship to the underlying electrification probability is not well-characterized.
\begin{itemize}
    \item \textit{Building density: } The aformentioned data set  \citep{Sofreco2013} also provides building location and is used to compute building density features used in our analysis.
    
    \item \textit{Nighttime lights: } nighttime lights imagery is available globally at 30 arc-second (\textsim~1~\kilometers) pixel resolution from the Defense Meteorological Satellite Program - Operational Linescan System (DMSP-OLS) \citep{NGDC2013}. Annual composite images are used that represent averages over nightly images that have been processed to remove sources of measurement noise \citep{elvidge1997mapping, elvidge2001night}. We use annual composite data for the year 2013, since more recent DMSP-OLS annual composites are not available.  For convenience, the resolution of the grid used in the model is set to match the resolution of this imagery.
    
    \item \textit{Number of transformers: } Distribution transformers are indicative of electricity access. Low voltage lines ultimately connect to these units and geo-referenced transformer data is very often more easily available than low-voltage line data in low-access countries. The transformer location data we use for Kayonza was provided by EDCL and is considered 100\% complete as of 2015. For each region, we include as a feature the number of transformers in each cell.
    
    \item \textit{Distance to the nearest transformer: } Using the same data source as the above, we compute an additional feature for each region: the distance from its center to its nearest transformer.
    
\end{itemize}

Auxiliary data are shown in Appendix~C.  For evaluation purposes we perform standardization of both the training and validation sets using \textit{only} the statistics of the training data.

\subsection{Approach}

We employ the basic model structure of Figure~\ref{fig:LDF-gmdisc} for inferring electrification status in Kayonza.   
This amounts to calculating the full posterior, $p(\theta \given x, y)$ where $\theta = (\theta_1, \ldots, \theta_M)$ are the electrification probabilities, $x = (x_1, \ldots, x_M)$ are the auxiliary data, and $y = (y_1, \ldots, y_M)$ are the primary data for the $M=854$ regions.
Here we describe the model, the learning procedure for the conjugate mapping parameters $\phi$, and the details of posterior inference $p(\theta \given x, y)$.

\subsubsection*{Model}

For each region we set the primary data $y_i$ be the number of the $N_i$ detected buildings that are within range of the low-power network, and interpret these as a binomial trial with success probability $\theta_i$.
We assume a conjugate prior distribution for the electrification probabilities, yielding the primary data model
\begin{align}
p(\theta_i)
    &= \betapdf(\theta_i; \alpha_0, \beta_0) & 
p(y_i \given \theta_i)
    &= \binomialpmf(y_i; N_i, \theta_i) & i &= 1, \ldots, M
\end{align}
where the prior is shared by all regions and the hyperparameters $\alpha_0$ and $\beta_0$ are interpreted as sufficient-statistics from prior observations with $\alpha_0$ prior successes and $\beta_0$ prior failures.
Consequently, the primary data posterior
\begin{align}
p(\theta_i \given y_i)
&= \betapdf(\theta_i; \alpha_0 + y_i, \beta_0 + N_i - y_i) 
& i &= 1, \ldots, M
\end{align}
is well-known.  The primary data are summarized by the sufficient statistics 
$T_\sfy(y_i) = [y_i;\:N_i]$ 
and the posterior entropy decreases with the size of the primary data $N_i$.

We learn a conjugate mapping of auxiliary data $x_i$ into a set of sufficient statistics $T_\sfx(x_i; \phi) = [t_\sfx(x_i; \phi);\:n_\sfx(x_i; \phi)]$ using our approach of Sections~\ref{sec-mappings} and \ref{sec-learning-and-inference}.
The impact of $x_i$ on the posterior is equivalent to $n_\sfx(x_i; \phi) \geq 0$ primary data trials with $t_\sfx(x_i; \phi) \in [0, n_\sfx(x_i; \phi)]$ (potentially non-integral) successes.
Before proceeding, we define several other quantities that are sufficient statistics for $x_i$ when paired with the number of pseudo-observations $n_\sfx(x_i; \phi)$:
\begin{itemize}
\item The number of successes: $a(x_i; \phi) = t_\sfx(x_i; \phi) \geq 0$
\item The number of failures: $b(x_i; \phi) = n_\sfx(x_i; \phi) - t_\sfx(x_i; \phi) \geq 0$
\item The success rate: $\mu(x_i; \phi) = t_\sfx(x_i; \phi) / n_\sfx(x_i; \phi) \in [0, 1]$ that is an auxiliary-data-based MLE of the parameter $\theta_i$.
\end{itemize}

It is convenient to discuss the derived distributions in terms of the number of successes $a(x_i; \phi)$ and failures $b(x_i; \phi)$.
Conditioning on auxiliary data occurs through the following form,
\begin{align}
p(\theta_i \given x_i; \phi) 
    &= \betapdf(\theta_i; \alpha_0 + a(x_i; \phi), \beta_0 + b(x_i; \phi)), & i &= 1, \ldots, M,
\label{eq-elec-posterior}
\end{align}
which remains conjugate to the primary data likelihood.  Therefore, the full posterior is
\begin{align}
p(\theta_i \given x_i, y_i; \phi)
    &= \betapdf(\theta_i;  
    \alpha_0 + a(x_i; \phi) + y_i, 
    \beta_0 + b(x_i; \phi) + N_i - y_i), & i &= 1, \ldots, M.
\label{eq-elec-full-posterior}
\end{align}
The primary data posterior predictive is
\begin{align}
p(y_i &\given x_i; \phi) 
    = \int p(y_i \given \theta_i) p(\theta_i \given x_i; \phi) \dif\theta_i 
    = \betabinomialpmf(y_i; N_i, \alpha_0 + a(x_i; \phi), \beta_0 + b(x_i; \phi))
\label{eq-elec-predictive}
\end{align}
for $i = 1, \ldots, M$, where $\betabinomialpmf(\cdot)$ denotes the beta-binomial compound distribution.

\subsubsection*{Learning}

\begin{figure}[t!]
  \vskip 0.1in
  \begin{center}
    \centerline{
    \includegraphics[width=0.5\linewidth, trim={0.5cm 0.3cm 0.6cm 0.5cm},clip]{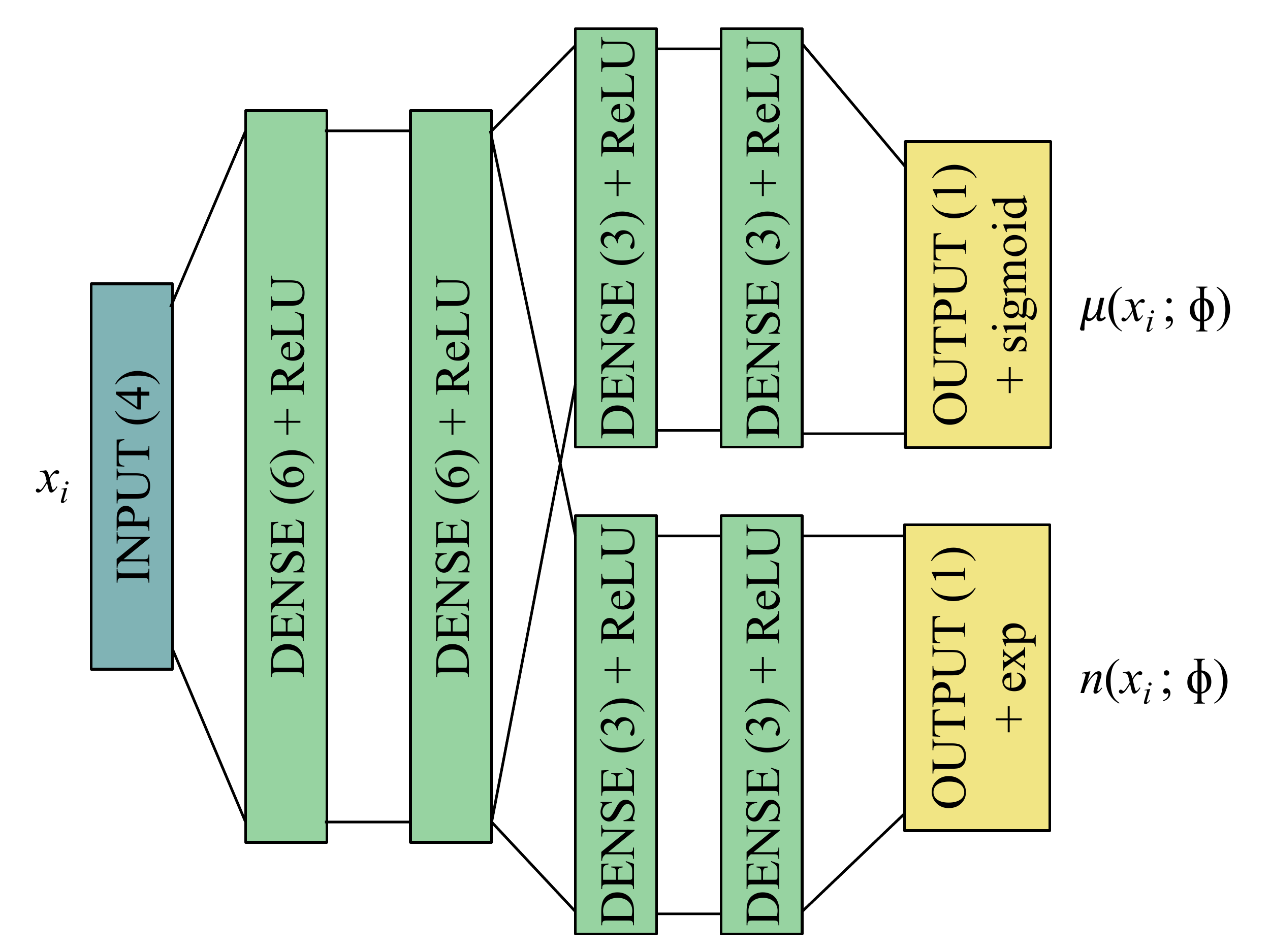}    
    }
    \caption{\small \textbf{NN architecture used as a conjugate mapping for electrification.} Connections between layers are standard dense linear projections (with biases), as per standard practice.  Layer sizes and activation functions are explicitly denoted.
    }
    \label{fig-elec-architecture}
  \end{center}
  \vspace*{-5mm}
\end{figure}

For learning it is convenient to specify the conjugate mapping in terms of the functions $n_\sfx(x_i; \phi) \geq 0$ and $\mu(x_i; \phi) = t_\sfx(x_i; \phi)/n_\sfx(x_i; \phi) \in [0, 1]$, defined as a NN with parameters $\phi$ shown in Fig.~\ref{fig-elec-architecture}.
This branched network architecture enables LDF to learn a common, shared representation of the auxiliary data-information space before specializing to what is important for learning the number of pseudo-counts $n_\sfx(x_i; \phi) \geq 0$ and pseudo-rate $\mu(x_i; \phi) \in [0, 1]$ independently.
While the network is fairly small, with only 146 trainable parameters, it is sufficiently expressive to extract relevant information from the auxiliary data.

Network parameters are learned to minimize the model-selection loss (Equation~\ref{eq-loss-model-selection}) relative to a trained beta-binomial regression reference model (c.f. Appendix~C).
Class-balancing is incorporated thru the weights $w_i$ based on five quantiles of the empirical rate, $y_i / N_i$.

We trained our networks in TensorFlow using the Adam optimizer with a batch size of 256.  
To help avoid terminating in highly-suboptimal local minima we employ early stopping with a patience of 100 epochs (using a randomly-selected $10\%$ of the training data), L2 weight decay, and 10 random initializations (with the best model selected based on the log-likelihood of a randomly-selected $10\%$ of the training data).

\subsubsection*{Inference}

Posterior inference over the latent variables $\theta$ given the data $x$ and $y$ is immediate.  The full posterior 
\begin{align}
p(\theta \given x, y; \phi)
    &= \prod_{i=1}^M p(\theta_i \given x_i, y_i; \phi)
\end{align}
completely factorizes into a product of terms given by Equation~\ref{eq-elec-full-posterior}.

\subsection{Experimental Results}

We compare against both binomial and beta-binomial regression that are special cases of our LDF approach applied to the model with no global variables, $\gamma = \emptyset$.
For more details on the baselines refer to Appendix~C.
They both focus on learning a \textit{single} transformation of the auxiliary data, 
\begin{align}
\mu(x_i; \phi) = \exp(\phi^T \tilde{x}_i)
\end{align}
for some $\phi \in \bbR^{p+1}$ where $\tilde{x}_i$ is the auxiliary data $x_i$ in homogeneous coordinates.  Beta-binomial regression additionally learns a constant number of pseudo-trials $n_\sfx(x_i; \phi) = n_\sfx \geq 0$ that are shared across all instances $i$.  Consequently, the number of pseudo-successes can be given by $t_\sfx(x_i; \phi) = \mu(x_i; \phi) n_\sfx(x_i; \phi)$.

As with LDF, both regression schemes encode the response variable/primary data $y_i$ as binomially-distributed with success probability $\theta_i$ and number of trials $N_i$.
Beta-binomial regression encodes a beta-distributed posterior over the latent success probability taking the same form as Equation~\ref{eq-elec-posterior}.  
The main distinction between LDF for models with $\gamma = \emptyset$ and beta-binomial regression is in the flexibility of our learned transformations and our ability to vary the number of pseudo-trials as a function of the data.  
Binomial regression is a departure from both, as it implies a degenerate distribution over $\theta_i$ that is a Dirac delta at $\mu(x_i; \phi)$. 
This is observed by noting that its objective
\begin{align}
p_\mathrm{Bin}&(y_i \given x_i; \phi) \nonumber \\
    &= \lim_{n_\sfx \rightarrow \infty} \int \binomialpmf(y_i; N_i, \theta_i) \betapdf(\theta_i; \alpha_0 + n_\sfx \mu(x_i; w), \beta_0 + n_\sfx (1-\mu(x_i; \phi))) \dif\theta_i \\
    &= \binomialpmf(y_i; N_i, \mu(x_i; \phi)) 
\end{align}
effectively assigns \textit{infinite} information content to the auxiliary data and precludes subsequent conditioning on the \textit{primary} data.

All of the predictive results are the result of 10-fold cross-validation.
In all of the plots provided we have \textit{not} trained on the region being predicted on, that is, each region shown was in the held-out set when its posterior parameters were being calculated.  The split is stratified on a digitized version of the parameter $y_i / N_i$ to ensure a representative sample.  
Hyperparameters were obtained by randomly sampling configurations on a log scale and were set to the following values: the prior hyperparameters  $\alpha_0 = 0.005316$ and $\beta_0=1.55$, the L2 weight decay penalty $1.74 \times 10^{-7}$, and the learning rate for the Adam optimizer $0.005418707$.

LDF compares favorably to the baselines along three different methods of comparison: interpretability, predictive accuracy, and the capacity for model checking.

\subsubsection*{Interpretable mappings}

\begin{figure}[t!]
  \vskip 0.1in
  \begin{center}
    \centerline{
    \includegraphics[width=0.75\linewidth, trim={0.5cm 0.55cm 0.55cm 0.275cm},clip]{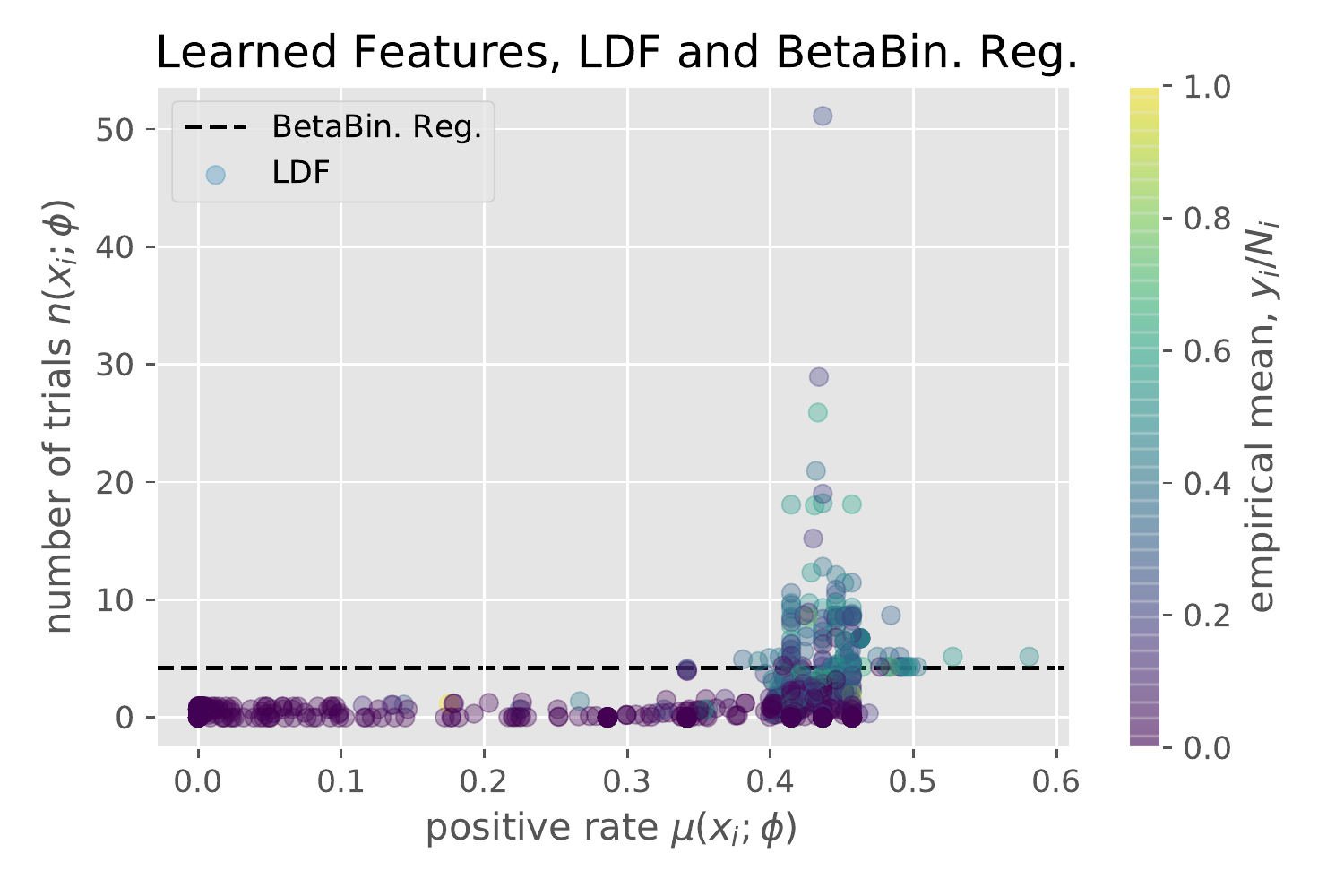}    
    }
    \caption{\small \textbf{Interpretable mappings.}  Circles denote the sufficient statistics learned for each auxiliary datum $x_i$ under LDF.  The equivalent statistics are shown for beta-binomial regression with the black dashed line showing the single pseudo-count shared by all data points.
    }
    \label{fig-elec-learned statistics}
  \end{center}
  \vspace*{-5mm}
\end{figure}

The closed-form nature of the exponential family posterior updates allow direct analysis of the information provided by the auxiliary data.  This is computable from the hyperparameters and the learned sufficient statistics.
In Fig.~\ref{fig-elec-learned statistics} we show on a per-region basis the success rate $\mu(x_i; \phi)$ contributed compared to the number of pseudo-trials $n_\sfx(x_i; \phi)$.  

Beta-binomial regression extracts a fixed number of pseudo-surveys due to fitting a single value shared by all regions.   Binomial regression, from the standpoint of the posterior $p_\textrm{bin}(\theta_i \given x_i)$, obtains $n_\sfx(x_i; \phi) = \infty$ for all $i$ and is omitted from this plot.
By comparison, LDF flexibly attributes less influence to data in less informative regimes and greater influence for data in more informative regimes.
This enables more accurate integration of the auxiliary data. 
For observations corresponding to low electrification probabilities there is relatively little confidence being gained by the LDF model from the auxiliary data; regions where the data contributes evidence of moderate electrification rates often contribute the equivalent of over eight primary data observations.  
In many regions with only one or two primary data observations the auxiliary data can prove at least as informative---despite the lack of a known forward model!

\subsubsection*{Compact posteriors} 

\begin{figure}[t!]
  \vskip 0.1in
  \begin{center}
    \centerline{
    \includegraphics[width=0.65\linewidth, trim={0.375cm 1cm 0.375cm 1.5cm},clip]{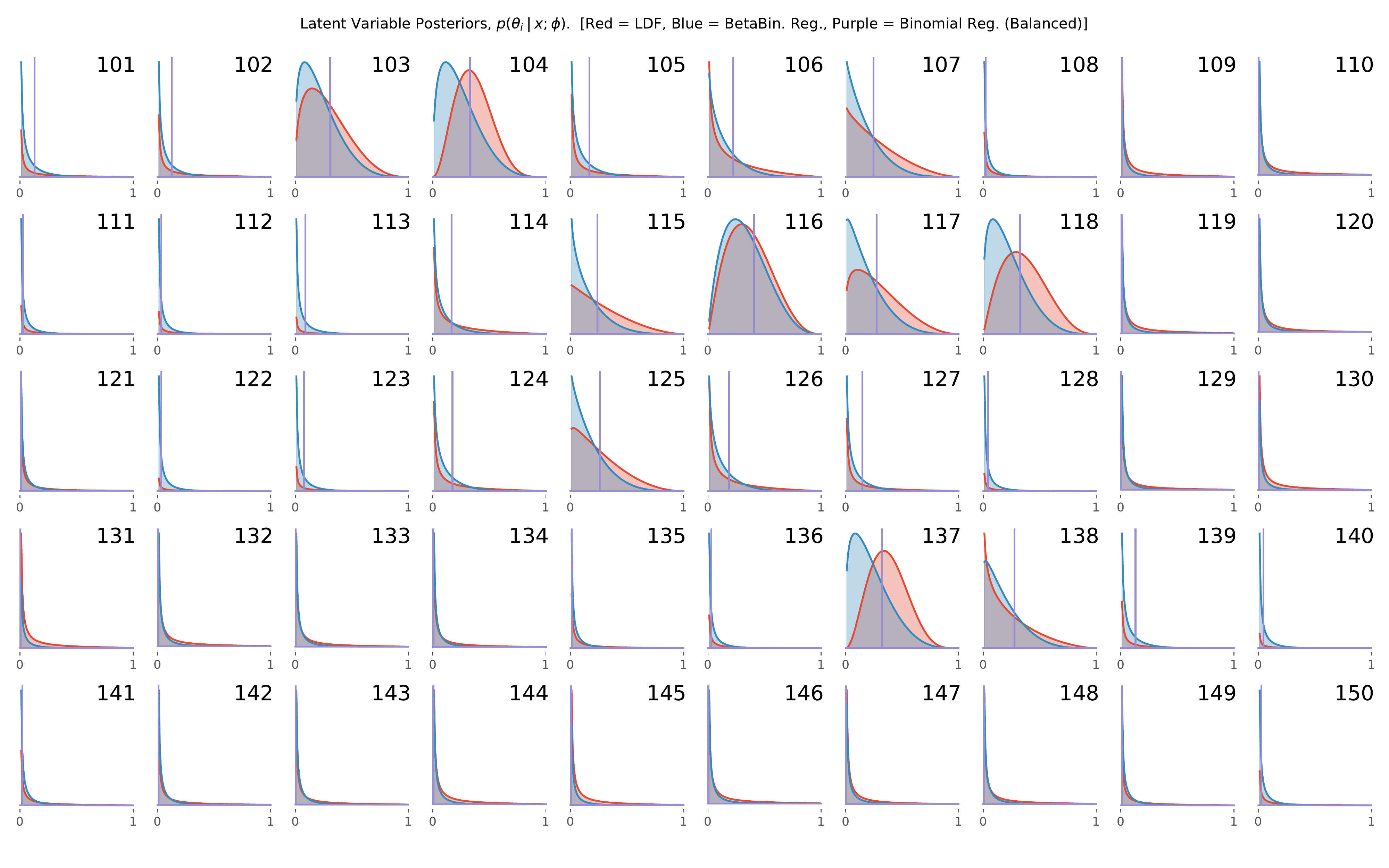}
    \includegraphics[width=0.35\linewidth, trim={0.375cm 0.45cm 0.9cm 0.125cm},clip]{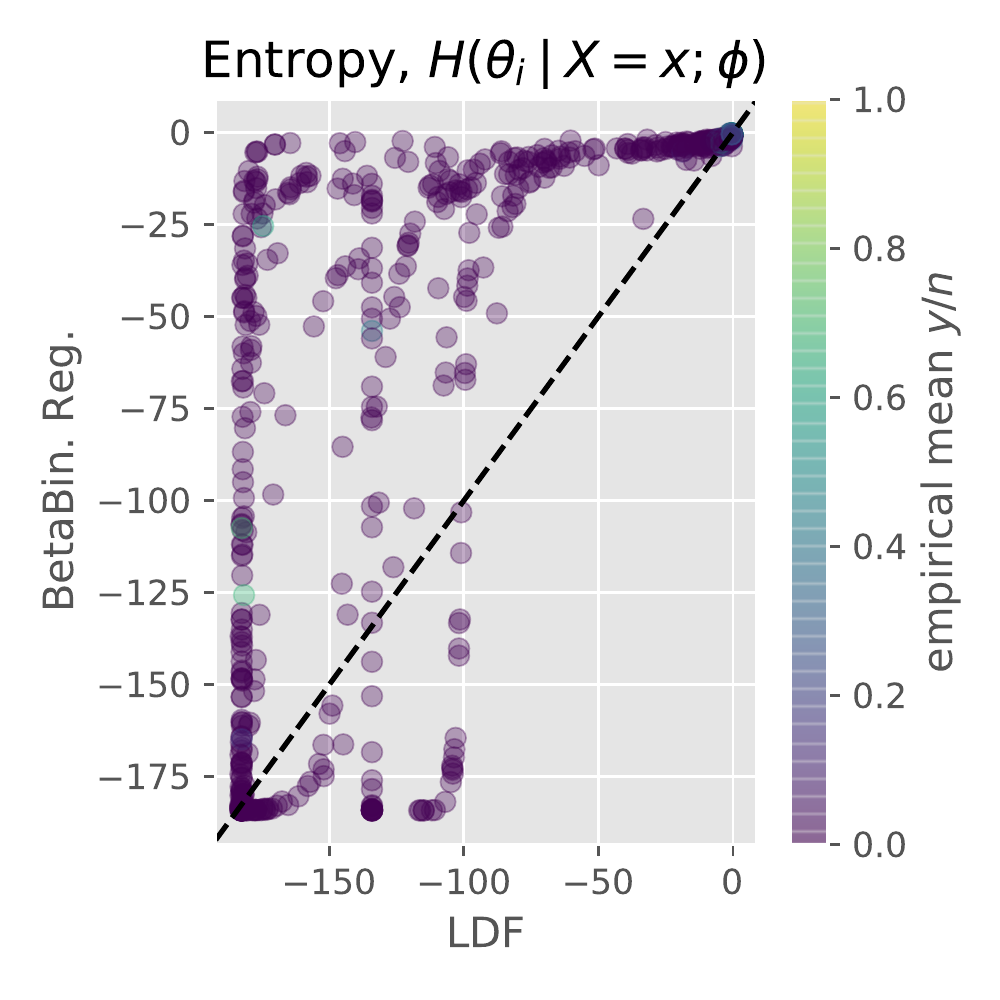}
    }
    \caption{\small \textbf{Latent variable posteriors.}  
        (Left)  A sample of the posteriors $p(\theta_i \given x; \phi)$ for the electrification application.  LDF is red, beta-binomial regression is blue, and binomial regression is purple.
        (Right) Comparison of the differences in posterior entropies $\bbH(\theta_i \given X = x; \phi)$ for each region between Beta-Binomial regression and LDF.  Positive values indicate higher entropy under the baseline; LDF has lower-entropy posteriors for $59.5\%$ of the regions.  Colored points represent individual local regions and are colored by the empirical mean of the primary data, $y_i / N_i$. 
    }
    \label{fig-elec-posterior-summaries}
  \end{center}
  \vspace*{-5mm}
\end{figure}

Our method produces full posterior distributions of the latent electrification probability $\theta_i$ for all $M=854$ regions.
A selection of these are shown in Figure~\ref{fig-elec-posterior-summaries} (left) for regions $101$ thru $150$ for their interesting variety of visual features, including regions whose probability of electrification are close to and far above zero and regions whose quantity of primary data observations are both high and low.

Any additional summary statistic of the posterior is readily accessible, e.g. posterior entropies $\bbH(\theta_i \given X = x; \phi)$ that quantify the uncertainty in the posterior $p(\theta_i \given x; \phi)$.  
We omit a comparison to binomial regression that only provides a point-estimate of $\theta_i$ whose entropy is 0.  
Fig.~\ref{fig-elec-posterior-summaries} (right) compares the entropy $\bbH(\theta_i \given X = x; \phi)$ between LDF and the Beta-binomial baseline, where $X$ and $x$ denote the r.v. and the realization, respectively.  
LDF produces more confident posterior distributions that extract more information than the baseline about the underlying electrification probabilities $\theta_i$.  
A large percentage of regions ($59.5\%$) reside in the upper triangle, where the models produced by our method are \textit{more} confident and exhibit less \textit{uncertainty} relative to the baseline.

\subsubsection*{Superior model fit relative to baselines}  

\begin{figure}[t!]
  \vskip 0.1in
  \begin{center}
    \centerline{
    \includegraphics[width=0.65\linewidth, trim={0.375cm 1cm 0.375cm 1.5cm},clip]{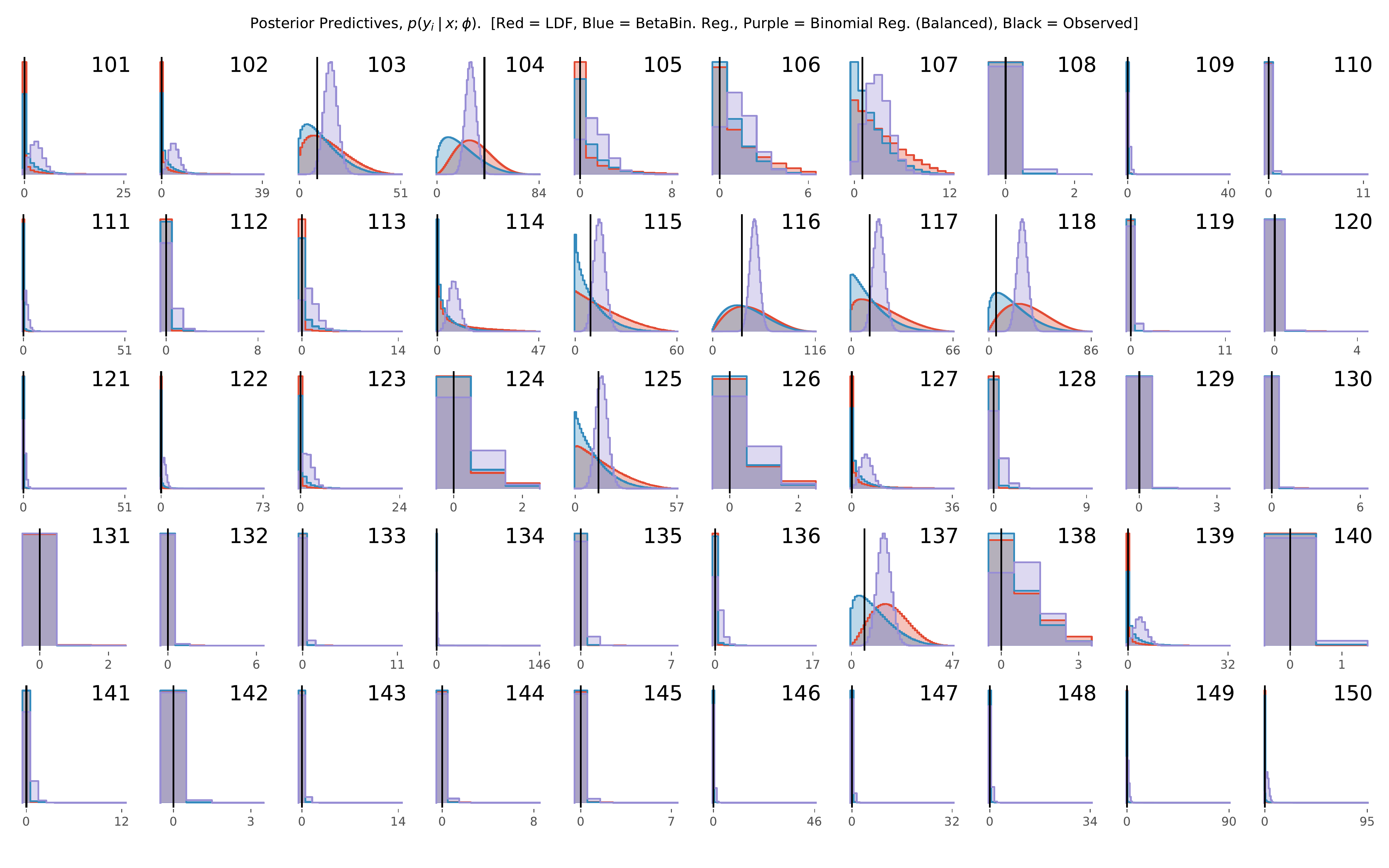} \:
    \includegraphics[width=0.35\linewidth, trim={0.375cm 0.325cm 0.9cm 0.125cm},clip]{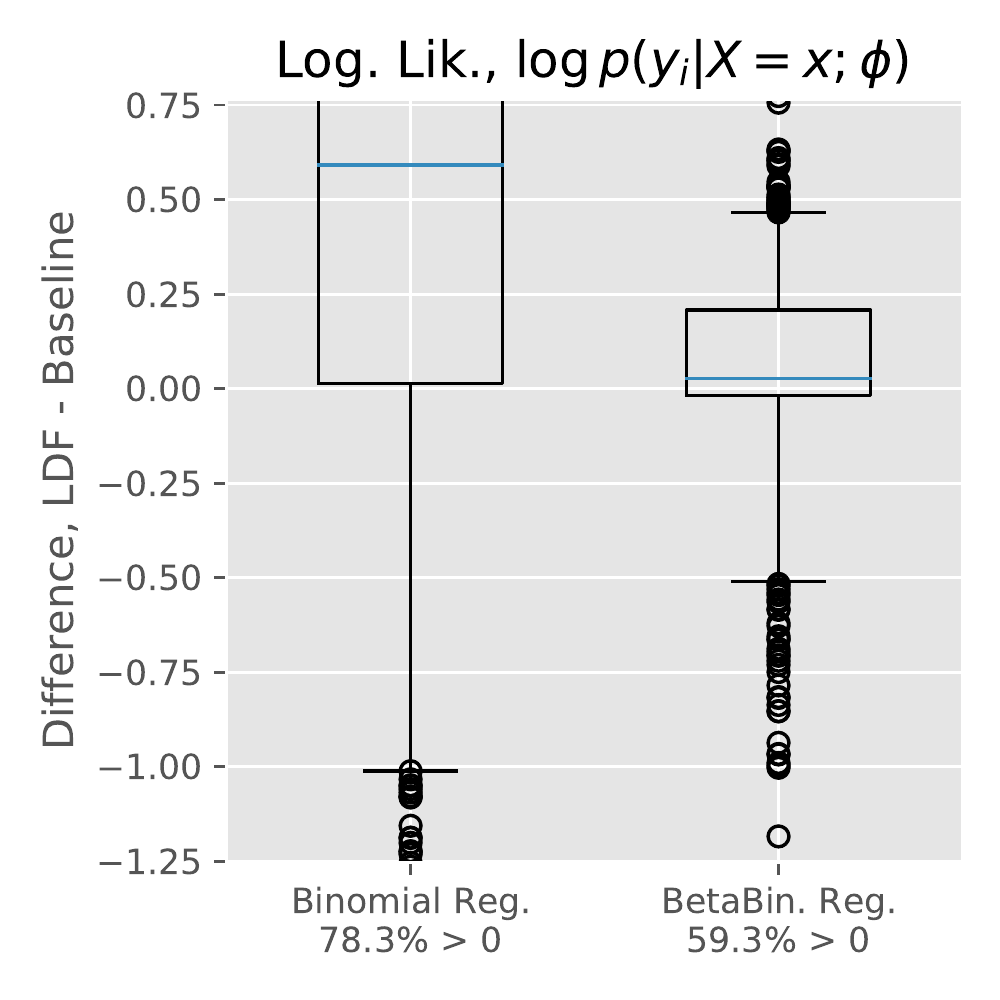}
    }
    \caption{\small \textbf{Primary data posterior predictives.}  
        (Left) A subset of the predictive likelihoods $p(y_i \given x; \phi)$ for the electrification application.  LDF is red, beta-binomial regression is blue, binomial regression is purple, and the observed value is black.
        (Right) Comparison of the difference in predictive log-likelihoods $p(y_i \given x_i; \phi)$ for each region between LDF and the baselines.  Positive values indicate the point is more likely under LDF; the 75th-percentile for binomial regression is $3.24$.   Colored points represent individual local regions and are colored by the empirical mean of the primary data, $y_i / N_i$.  
    }
    \label{fig-elec-model-checking}
  \end{center}
  \vspace*{-5mm}
\end{figure}

In addition to posteriors of the latent variable $\theta_i$, LDF also provides posterior predictive distributions $p(y_i \given x_i; \phi)$ that provide a better fit of held-out survey data.
Figure~\ref{fig-elec-model-checking} (left) shows these distributions for the same 50 regions from Fig.~\ref{fig-elec-posterior-summaries}.
Model checking on credible regions indicates that our model is slightly conservative w.r.t. uncertainty quantification, with over $95\%$ of the primary data residing within it's region's minimum-width $90\%$ credible region.
Other common techniques for Bayesian models can be also applied, e.g. \citep{gelman1996posterior}.

Fig.~\ref{fig-elec-model-checking} (right) compares the per-region difference of predictive log-likelihood $\log p(y_i \given x; \phi)$ of the held-out data between our method and the baseline.  LDF provides a better fit compared to the baselines for $78.3\%$ (binomial) and $59.3\%$ (beta-binomial) of the observed regions, indicated by the positive values.

\section{Case Study: Mixture Models for Homicide Rate Analysis}
\label{sec-homicide}

\begin{figure}[t!]
	\vskip 0.1in
	\begin{center}
		\begin{subfigure}[b]{\textwidth}
			\centering
			\includegraphics[width=1.0\linewidth, trim={0cm 0cm 0cm 0cm},clip]{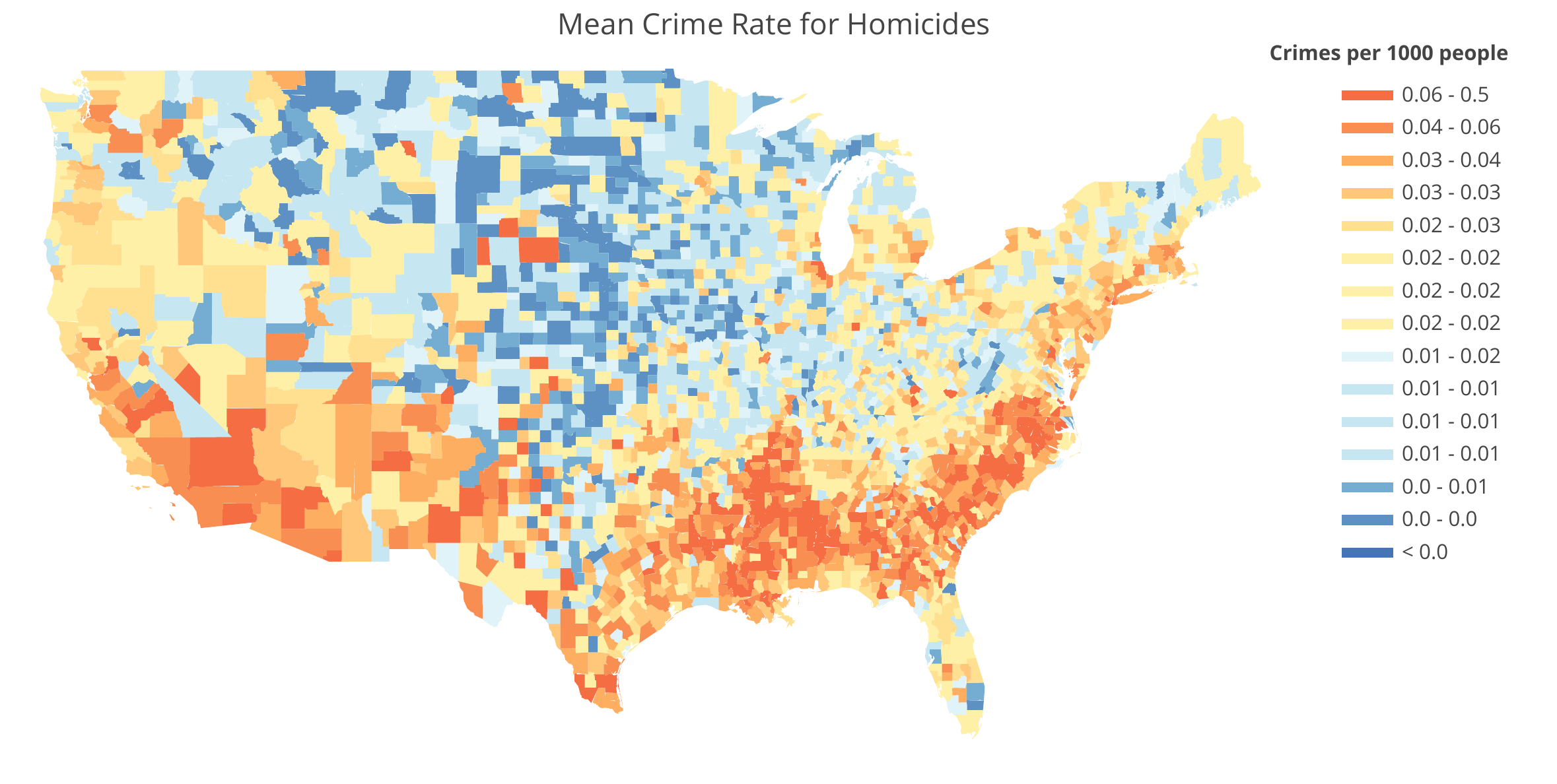}
		\end{subfigure}
		\caption{\small \textbf{Posterior mean per-capita homicide rate} $\bbE[\theta]$. Statistics characterizing posterior uncertainties, e.g. entropies $\bbH(\theta_i \given x_i; \phi)$, are also available under our approach.}
		\label{fig-geomap-mean-theta}
	\end{center}
	\vspace*{-5mm}
\end{figure}

Here we demonstrate our LDF methodology in a more complex model setting.  
A key feature of LDF is that it is easily embedded in more complex-structured PGMs.
Our approach enables posterior reasoning over \textit{all latent variables} in the graphical model.
This includes computing expectations or other statistics of the local latent variables $\theta_i$ or any other latent variables in the model.

We model per-capita homicide rates for each U.S. county, using 2010 crime statistics as primary data and socio-economic features as auxiliary data.
Figure~\ref{fig-geomap-mean-theta} shows the posterior mean homicide rate for each county in the U.S.
Critically, our analysis anonymizes all U.S. Census Bureau features based on ethnicity to prohibit the casual reader from drawing false conclusions.  
We do not predict causal relationships nor do we account for information redundancy among features.  
We aim primarily to illustrate the application of our technique to more complex model structures.

\subsection{Data}

We learn a joint primary-auxiliary data model for homicide rates using a variety of publicly-available data sources.

\textbf{Primary data} are the observed number of homicides from the year 2010, as reported in the U.S. Federal Bureau of Investigation's (FBI) Uniform Crime Reporting Program Data \citep{ICPSR2010}.  These are provided for each county but are treated as noisy realizations, \textit{not} ground-truth estimates of the true per-capita homicide rate, following the convention from relevant literature \citep{hepburn2004firearm, nielssen2008rates, williams1984economic, fernandes2018race, Osgood2000}.
As in Sec.~\ref{sec-electrification}, generally the primary data are of limited availability enable training an auxiliary data model that is more broadly applicable to settings without primary data.

\textbf{Auxiliary data}  are a set of 47 different county-level socio-economic features from the U.S. Census Bureau's 2010 Census of Population and Housing \citep{Census2011}. 
The specific features are tabulated in Appendix~D.
Several of these include: current population as of 2010, per-capita unemployment rate, the proportion of the population that is male, the proportion that voted Democratic in 2008, and a set of features based on income bracket and ethnicity.

In many cases the raw features were converted into per-capita rates.  This was done for features presented by the U.S. Census Bureau presented as county-wide \textit{counts}.
This information is also provided in the appendix.  
All auxiliary data was standardized to have mean 0 and standard deviation 1.  As in Section~\ref{sec-electrification}, this is done based on the statistics of the training set for each fold of cross-validation.

\subsection{Approach}

For consistency with existing literature, we interpret the primary data as exposure-adjusted Poisson observations, where the population represents the degree of exposure.
Concretely, the number of occurrences $y_i \given \theta_i \sim \poissonpdf(n_i \theta_i)$ in each of $M=3,\!137$ counties.  The county population $n_i$ is given in thousands, known from the 2010 census statistics.  The per-capita homicide rate $\theta_i$ is in terms of occurrences per thousand people.  We use a conjugate prior for the homicide rates $\theta_i \sim \gammapdf(\alpha_0, \beta_0)$ in shape-rate parameterization, where $\alpha_0$ corresponds to a prior number of occurrences in $\beta_0$ prior degree of exposure. 
Each county further has auxiliary data $x_i$ that we learn to condition on using conjugate mappings.
 
Here we show an extension of LDF onto models with additional PGM structure, i.e. $\gamma \neq \emptyset$.
In particular, we consider a PGM that is a mixture over different conjugate mappings that vary across counties.

\subsubsection*{Model}

Consider a model where instead of a single mapping $T_\sfx(x_i; \phi)$ being used to condition on auxiliary data $x_i$ we have \textit{multiple} conjugate mappings $T_\sfx(x_i; \phi_k)$ of auxiliary data into aggregated sufficient statistics, where $k = 1, \ldots, K$ indexes the mapping.  Different models $i = 1, \ldots, M$ express different relationships between the auxiliary data $x_i$ and the underlying latent parameter $\theta_i$.
Our notation assumes for convenience that all $K$ of the conjugate mappings fall into the same parametric class and are sufficiently differentiated by indexing their parameter sets $\phi_1, \ldots, \phi_K$; this is simply for convenience and \textit{not} a limitation of the approach.

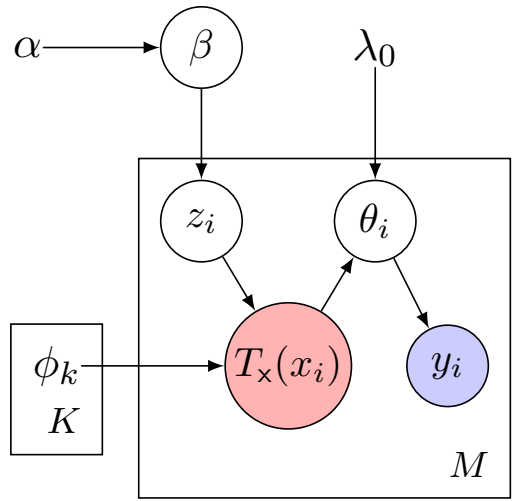
\begin{figure}[t!]
     \vskip 0.1in
     \centering
    \resizebox{0.45\columnwidth}{!}{%
%

\begin{tikzpicture}
%
%
\tikzstyle{box}+=[ rounded corners = 5pt,
                       align           = left,
                       font            = \footnotesize,
                       text width      = 2.45cm]

\tikzstyle{every path}+=[thin,>=latex];
\tikzstyle{plate}+=[thin,sharp corners,inner sep=5pt,outer sep=0pt]; 

%
%

%
%
\def \xmncolor {red!70}
\def \thetcolor {cyan!70!black}
\def \zmncolor {yellow!70!black}
\def \betcolor {green!70!black}

%
%
%
\tikzstyle{latent}+=[node distance=1.25 and 1.5, on grid];
\tikzstyle{obs}+=[fill=blue!20,node distance=1.25 and 1.5,on grid];
\tikzstyle{obsaux}=[latent,fill=red!30,node distance=1.25 and 1.5,on grid];
\tikzstyle{const}+=[node distance=1.25 and 1.5,on grid];
\tikzstyle{plate}+=[color=black];


\node[const] (alpha) {$\alpha$};
\node[latent, right = of alpha] (pi) {$\beta$};
\node[latent, below = 1.5 of pi] (zi) {$z_i$};
\node[latent, right = of zi] (thetai) {$\theta_i$};
\node[obs, below right= 1.25 and 0.625 of thetai] (yi) {$y_i$};
\node[obsaux, below left= 1.25 and 0.75 of thetai] (xi) {$T_\sfx(x_i)$};
\node[const, left = 2.0 of xi] (phik) {$\phi_k$};
\node[const, right = of pi] (lambda) {$\lambda_0$};


\edge {alpha} {pi};
\edge {pi} {zi};
\edge {lambda} {thetai};
\edge {thetai} {yi};
\edge {xi} {thetai};
\edge {phik} {xi};
\edge {zi} {xi};


\plate {dataplate} {(zi)(yi)(xi)} {$M$};
\plate {atomplate} {(phik)} {$K$};

\end{tikzpicture}

}
    \caption{\small{A mixture model over latent parameters $\{\phi_k\}_{k=1}^K$.}}
    \label{fig-mm-phi-dis-gm}
\end{figure}

Consider a primary data likelihood $f(\cdot)$ and its conjugate prior $\pi(\cdot)$, in our example Poisson and gamma distributions respectively.  We define a mixture model over conjugate mappings shown in Figure~\ref{fig-mm-phi-dis-gm}.  Assuming a shared conjugate prior $p(\theta_i) = \pi(\theta_i; \lambda_0)$ across counties we have
\begin{align}
p(\beta) &= \dirpdf(\beta; \alpha), &
p(z_i \given \beta) &= \catpmf(z_i; \beta), \nonumber \\
p(\theta_i \given x_i, z_i; \phi) &= \pi(\theta_i; \lambda_0 + T_\sfx(x_i; \phi_{z_i})), &
p(y_i \given \theta_i) &= f(y_i; \theta_i).
\end{align}
Hyperparameter $\alpha \in \bbR_+^K$ is a vector of prior pseudo-counts and $\lambda_0$ are the hyperparameters of the prior on $\theta_i$.  Mapping weights vector $\beta \in \Delta^{K-1}$ is on the probability simplex and gives rise to the (unobserved) mapping assignments.  The marginal (over $z_i$) posterior distribution $p(\theta_i \given x_i; \phi)$ is now a mixture distribution, however, given the conjugate mapping assignment $z_i$ the posterior $p(\theta_i \given x_i, z_i; \phi)$ and the posterior predictive
\begin{align}
p(y_i \given x_i, z_i; \phi) = \int f(y_i; \theta_i) \pi(\theta_i; \lambda_0 + T_\sfx(x_i; \phi_{z_i}))\dif\theta_i
\label{eq-mixtures-predictive}
\end{align}
remains closed-form conditioned on $z_i$.

While this model bears some similarities to a standard Bayesian mixture model there are critical differences.
This mixture is over the conjugate mappings $T_\sfx(\cdot)$, \textit{not} the latent variables $\theta_i$.  As such, different models (indexed by $i$) each have their own latent variable $\theta_i$, unlike in standard mixture models where the latent variables are shared across multiple data points. 
Instead, the $i$-th model has a hard assignment via the indicator variable $z_i$ to one of the $K$ conjugate mappings that govern how the auxiliary data $x_i$ influences the parameter $\theta_i$.
In our setting this mixture model implies that different counties have different relationships between the socio-economic features of the auxiliary data $x_i$ and the per-capita homicide rate, dictated by the mapping assignment $z_i$.
When $K = 1$ this model contains only a single conjugate mapping and is equivalent to the $\gamma=\emptyset$ model of Sec.~\ref{sec-mappings}.

Absent the auxiliary data, the posterior distribution of the per-capita rate $\theta_i$ given primary data $y_i$ is
\begin{align}
    p(\theta_i \given y_i) 
    &= \gammapdf(\theta_i; \alpha_0 + y_i, \beta_0 + 1) \label{eq-gapo-primary-posterior}
\end{align}
where the primary data are summarized by aggregate sufficient statistics $T_\sfy(y_i) = [y_i;\, 1]$.  These are interpretable as the number of occurrences and a number of multiples of the exposure $n_i$.  
Analogously, the $k$-th conjugate mapping $T_\sfx(x_i; \phi_k)$ converts the auxiliary data vector $x_i$ into a number of occurrences $a(x_i; \phi_k) \triangleq t_\sfx(x_i; \phi_k)$ that occur within $b(x_i; \phi_k) \triangleq n_\sfx(x_i; \phi_k)$ multiples of the exposure.
We could additionally cast the conjugate mapping output in terms of the number of intervals $b(x_i; \phi)$ and the empirical rate, $r(x_i; \phi) \triangleq a(x_i; \phi) / b(x_i; \phi)$, the number of occurrences per interval.
It is trivial to show that the posterior parameters can be written as a weighted combination of the prior hyperparameters and the outputs of the conjugate mapping functions. 

Conditioned on the mapping assignment $z_i$ the above model provides the auxiliary data posterior
\begin{align}
p(\theta_i \given x_i, z_i; \phi) 
&= \gammapdf(\theta_i; \alpha_0 + a(x_i; \phi_{z_i}), \beta_0 + b(x_i; \phi_{z_i})), \label{eq-gapo-joint}
\end{align}
and the posterior predictive
\begin{align}
p(y_i \given x_i, z_i; \phi) 
    &= \int_0^\infty \poissonpdf(y_i; n_i \theta_i) \gammapdf(\theta_i; \alpha_0 + a(x_i; \phi_{z_i}), \beta_0 + b(x_i; \phi_{z_i})) \dif\theta_i \nonumber \\
    &= \gammapoissonpmf(y_i; \alpha_0 + a(x_i; \phi_{z_i}), n_i^{-1}(\beta_0 + b(x_i; \phi_{z_i}))) \label{eq-gapo-predictive}  
\end{align}
that is the Gamma-Poisson compound distribution.\footnote{We denote by Gamma-Poisson distribution the pmf with shape parameter $a > 0$ and rate parameter $b > 0$:
\begin{equation*}
p(y \given a, b) = \int_0^\infty \poissonpmf(y; \theta) \gammapdf(\theta; a, b) \dif\theta = \dfrac{\Gamma(a + y)}{y! \Gamma(a)} \left(\dfrac{1}{b + 1}\right)^y \left(\dfrac{b}{b + 1}\right)^a.
\end{equation*}
For integer $\alpha$ this distribution is a reparametrization of the Negative Binomial distribution commonly used in failure-rate analysis.  A r.v. $X \sim \textrm{NB}(r, p)$ describes the total number of trials before $r= \alpha \in \bbN$ failures are observed, after which the experiment is stopped, and trials have success probability $p = 1 / (1 + \beta) \in (0, 1)$. Given parameters $r$ and $p$ we can convert back to Gamma-Poisson parameters via $\alpha = r$ and $\beta = (1 - p) / p > 0$.
} 
Both of these terms clearly depend on the mapping assignment $z_i$that dictates how to condition on the auxiliary data $x_i$.

\subsubsection*{Learning}

A critical quantity in learning the conjugate mapping parameters $\phi = \{\phi_1, \ldots, \phi_K\}$ is the primary data posterior predictive, $p(y_i \given x_i;\phi)$. 
Unlike in Sec.~\ref{sec-electrification}, this is not directly available from the model and instead requires marginalizing over the latent variables $\gamma = (\beta, z)$.
In general, as in this model, this marginalization cannot be done entirely in closed form.
Under the Bayesian framework of Sec.~\ref{sec-general-learning} there are a variety of approaches available for learning transformation parameters in more complex models, including the following methods for the ML criterion:
\begin{itemize}
    \item \textit{Approximate the marginal.}
The marginal likelihood $p(y_i \given x; \phi)$ may be approximated by sampling-based methods.  We can of course marginalize out the local latent variables $\theta_i$ in closed form given its parents in the PGM by construction.
For this model we obtain distributions taking the general form of Eq.~\ref{eq-mixtures-predictive} (Gamma-Poisson models, c.f. Eq.~\ref{eq-gapo-predictive}) that condition on mapping assignments $z_i$.
Similarly, due to the form of the mixture model we can easily marginalize over the mixture assignments $z$ conditioned on $\beta$.   Unfortunately, marginalizing out the remaining latent variables $\beta$ requires approximate methods.  A reasonable choice is Monte Carlo integration:
\begin{align}
p(y_i \given x; \phi) 
    &= \int_\beta p(\beta) \sum_k \beta_k p(y_i \given x_i, z_i = k; \phi) \dif\beta \\
    &\approx \dfrac{1}{S} \sum_{s=1}^{S} \sum_k \beta_k^{(s)} p(y_i \given x_i, z_i = k; \phi), 
\end{align}
where $\beta^{(s)} \iidsim p(\beta)$ for $s = 1, \ldots, S$.  
Notably, this formulation renders the global variables statistically-independent of the auxiliary data.  This enables drawing a large number $S$ samples from the prior $p(\beta)$ \textit{once} and reusing them across every optimization loop.  Modern deep learning frameworks accommodate such loss functions.

\item \textit{Optimize a lower-bound on the true loss.}
An alternative approach is to maximize a lower bound on the true loss, e.g. the approach taken under generalized Expectation Maximization (EM) \citep{dempster1977maximum}.
For the experiments presented here we integrate $\theta = (\theta_1, \ldots, \theta_M)$ out of the objective in closed form via conjugacy.  We maximize a lower-bound on the remainder using an  EM procedure, provided in Appendix~B, that resembles Neural EM \citep{Greff2017}.
Upon convergence, our procedure returns estimates of the mapping weights $\beta \in \Delta^{K-1}$ and the mapping parameters $\phi = \{\phi_1, \ldots, \phi_K\}$ corresponding to a local optimum of the primary data likelihood $p(y \given x; \phi)$.

\end{itemize}

Regardless of the method for learning, each conjugate mapping is represented by a fully connected network architecture comprised of sequential hidden layers with 6-10 widths.
Input data is connected to the first layer of the network using a hyperbolic tangent activation function. The second hidden layer uses a rectified linear unit (ReLU) activation function. The last hidden layer is connected to the output layer producing $a(x_i; \phi)$ and $b(x_i; \phi)$ that are guaranteed positive by the softplus output activation.

Each model is learned using ten-fold cross-validation to ensure that each county appears in exactly one validation set. Stratification is employed during cross-validation based on homicide rates so that training and validation folds are balanced.  As in the experiments of Sec.~\ref{sec-electrification} parameter sweeps are done over L2 and dropout regularization parameters, also on a log-scale. 
Our method is trained for 1000 EM iterations with early stopping on EM done after 50 iterations of no training loss improvement. Within each EM step, training is performed on the component NNs for 1 epoch of batch size 256.

\subsubsection*{Inference}

Inference in such models is surprisingly straightforward when conditioning on the mapping parameters $\phi = \{\phi_1, \ldots, \phi_K\}$.  
As an example, we now provide Gibbs sampling equations for the full posterior $p(\beta, z, \theta \given x, y; \phi)$ of the above model.  We emphasize that this is only one option.
The full conditional distributions for the mapping weights $\beta$ are
\begin{align}
p(\beta \given z)
    = \dirichletpdf(\beta; \alpha_1 + M_1, \ldots, \alpha_K + M_K)
\end{align}
where $M_k = \sum_{i=1}^M \ind(z_i = k)$ is the number of models currently assigned to the $k$-th conjugate mapping.
For the mapping assignments $z_i$ the full conditional are a set of discrete distributions
\begin{align}
p(z_i = k \given \beta, \theta_i, x_i; \phi)
    \propto \beta_k \pi(\theta_i; \lambda_0 + T_\sfx(x_i; \phi_k))
\end{align}
that can all be computed and sampled from in parallel due to the conditional independence structure in the graph.  Notably, $z_i$ is conditionally independent of the primary data $y_i$ given the auxiliary data $x_i$ and the latent variable $\theta_i$.
Finally the local latent variables $\theta_i$ have full conditional
\begin{align}
p(\theta_i \given x_i, y_i, z_i; \phi) 
    = \pi(\theta_i; \lambda_0 + T_\sfx(x_i; \phi_{z_i}) + T_\sfy(y_i)) \label{eq-mixtures-theta-full-conditional}
\end{align}
by conjugacy; these can again be sampled in parallel across models $i = 1, \ldots, M$.

Hence, drawing samples from the posterior $p(\beta, z, \theta \given x, y; \phi)$ via Gibbs sampling results in an efficient and highly-parallelizable inference procedure.  Additional variations, e.g. collapsed Gibbs sampling with mapping assignments $z_i$ or local latent variables $\theta_i$ marginalized out, are also straightforward to develop for these types of models.

In the experimental results shown here, we have learned point estimates for both the transformation parameters $\phi$ and mixture weights $\beta$.  Instead of Gibbs sampling as above, we instead consider several useful distributions that can be calculated in closed-form under this model.  For example, the posterior distribution of the mapping assignment,
\begin{align}
p(z_i = k \given x, y; \phi)
    &\propto \beta_k p(y_i \given x_i, z_i=k; \phi)
\end{align}
where $p(y_i \given x_i, z_i=k; \phi)$ is given by Eq.~\ref{eq-gapo-predictive}.  The posterior of the rate parameters $\theta_i$ is also given by Eq.~\ref{eq-mixtures-theta-full-conditional} and specializes to
\begin{align}
p(\theta_i \given x_i, y_i, z_i) 
&= \gammapdf(\theta_i;  \alpha_0 + a(x_i; \phi_{z_i}) + y_i, \beta_0 + b(x_i; \phi_{z_i}) + 1)
\end{align}
for the Gamma-Poisson model.
From these distributions it is straightforward to compute a variety of useful statistics, e.g. the MAP estimate $z_{i\text{MAP}}$ for the mapping assignments or the expected homicide rate, 
\begin{align}
    \bbE[\theta_i \given x_i, y_i; \phi]
    =\bbE_{z_i \sim p(z_i | x_i, y_i, \beta; \phi)}\left[\bbE[\theta_i \given x_i, y_i, z_i; \phi]\right],
\end{align}
by the law of iterated expectations.

\subsection{Experimental Results}

We present results for two LDF models: the conjugate mapping mixture model described above with $K = 3$ and the single conjugate mapping model (i.e. $K=1$).
In subsequent discussion and figures we denote these as LDF MM and LDF models, respectively.
For both LDF models we have adopted the relatively small NN architectures described above to highlight the utility of the mixture model defined here.

We compare our LDF approaches against the Poisson 
and Negative-Binomial regression 
\citep{lawless1987negative} 
methods from the relevant literature.
As with the binomial and beta-binomial regression baselines of Sec.~\ref{sec-electrification} these baselines are special and/or asymptotic cases of our LDF approach.  For more details see Appendix~D.
Baseline hyperparameters are tuned via parameter sweeps on a log-scale for the L2 regularization parameters.

\subsubsection*{Rich and expressive posterior distributions}

Our method enables direct posterior analysis for all latent variables.  We have access to the full posterior of the local latent variables, $p(\theta_i \given x, y)$, in addition to those of the other latent variables, the mapping assignments $z_i$ and the mixture weights $\beta$.  From these we can compute any desired statistic of the posterior distribution.

\begin{figure}[t!]
	\vskip 0.1in
	\begin{center}
		\begin{subfigure}[b]{\textwidth}
			\centering
			\includegraphics[width=1.0\linewidth, trim={0cm 0cm 0cm 0cm},clip]{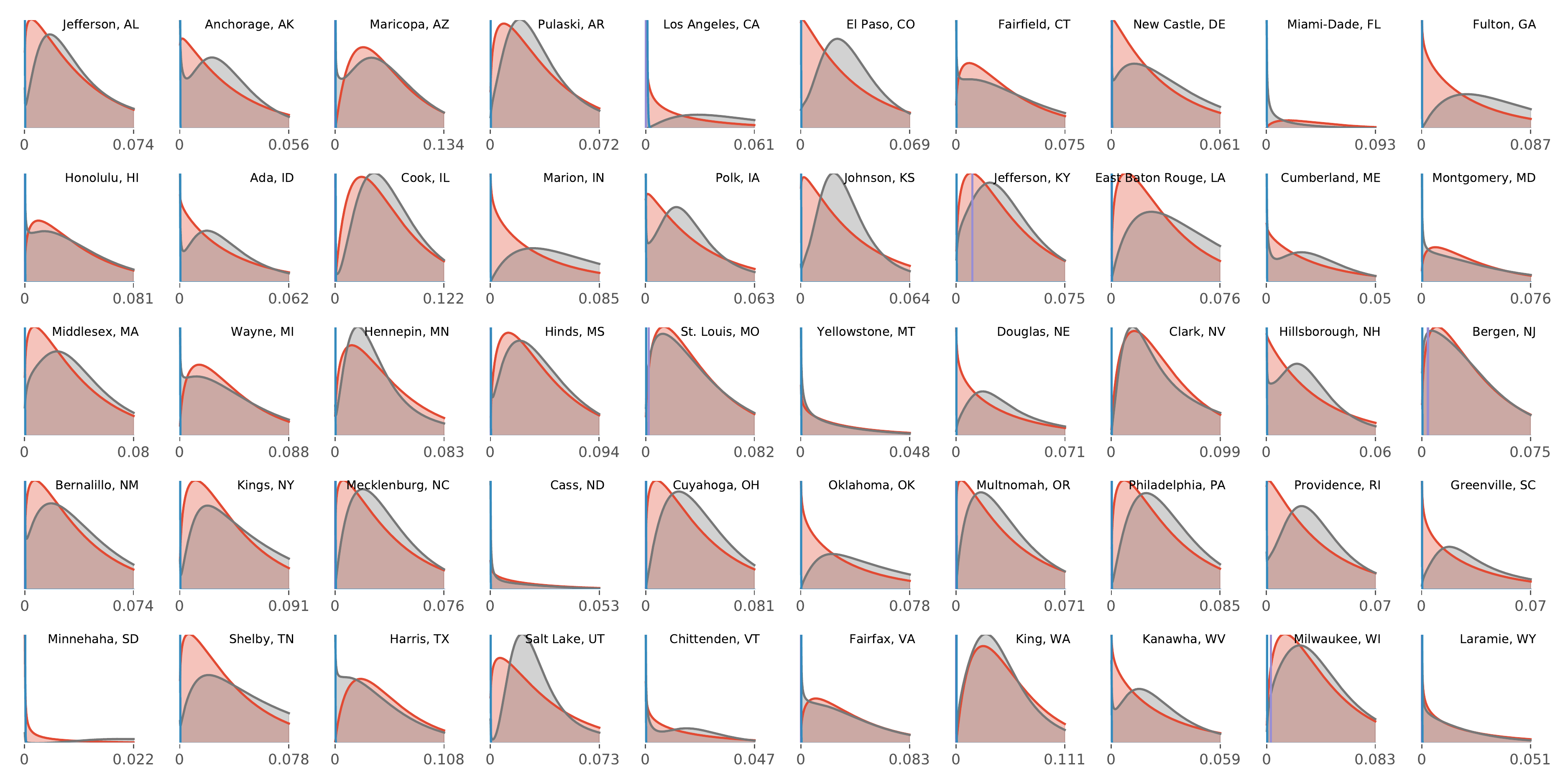}
		\end{subfigure}
		\caption{\small \textbf{Sample posterior distributions}. Posteriors $p(\theta_i|x_i;\phi)$ on homicide rates are shown for one county per state. 
		Distributions corresponding to LDF are in red, LDF MM in gray, gamma-Poisson regression in blue, and Poisson regression in purple.
		The counties displayed are those with the largest population for the given state.} 
		\label{fig-posteriors-homicide}
	\end{center}
	\vspace*{-5mm}
\end{figure}

The posterior distributions of the homicide rate enable both prediction and uncertainty quantification.  
Figure~\ref{fig-posteriors-homicide} shows the posterior distributions of per-capita homicide rate in 50 of the 3,137 U.S. counties, where we have selected the most populous counties from each state.  
We show posterior mean per-capita homicide rate in Figure~\ref{fig-geomap-mean-theta}, corresponding to the mean of the gray distribution.

Under our conjugate mapping mixture model the posteriors $p(\theta_i \given x; \phi) = \bbE_{z_i}[p(\theta_i \given z_i, x; \phi)]$ exhibits multimodal behavior as we see in Figure~\ref{fig-posteriors-homicide}.  This enables greater precision in our estimate of the homicide rate than in the models with only a single conjugate mapping.  In contrast the regression baselines often produce distributions that are highly-concentrated at very low occurrence probabilities; Poisson regression additionally learns a point estimate of the rate function and appears as a Dirac delta.

\begin{figure}[t!]
	\vskip 0.1in
	\begin{center}
		\begin{subfigure}[b]{\textwidth}
			\centering
			\includegraphics[width=1.0\linewidth, trim={0cm 0cm 0cm 0cm},clip]{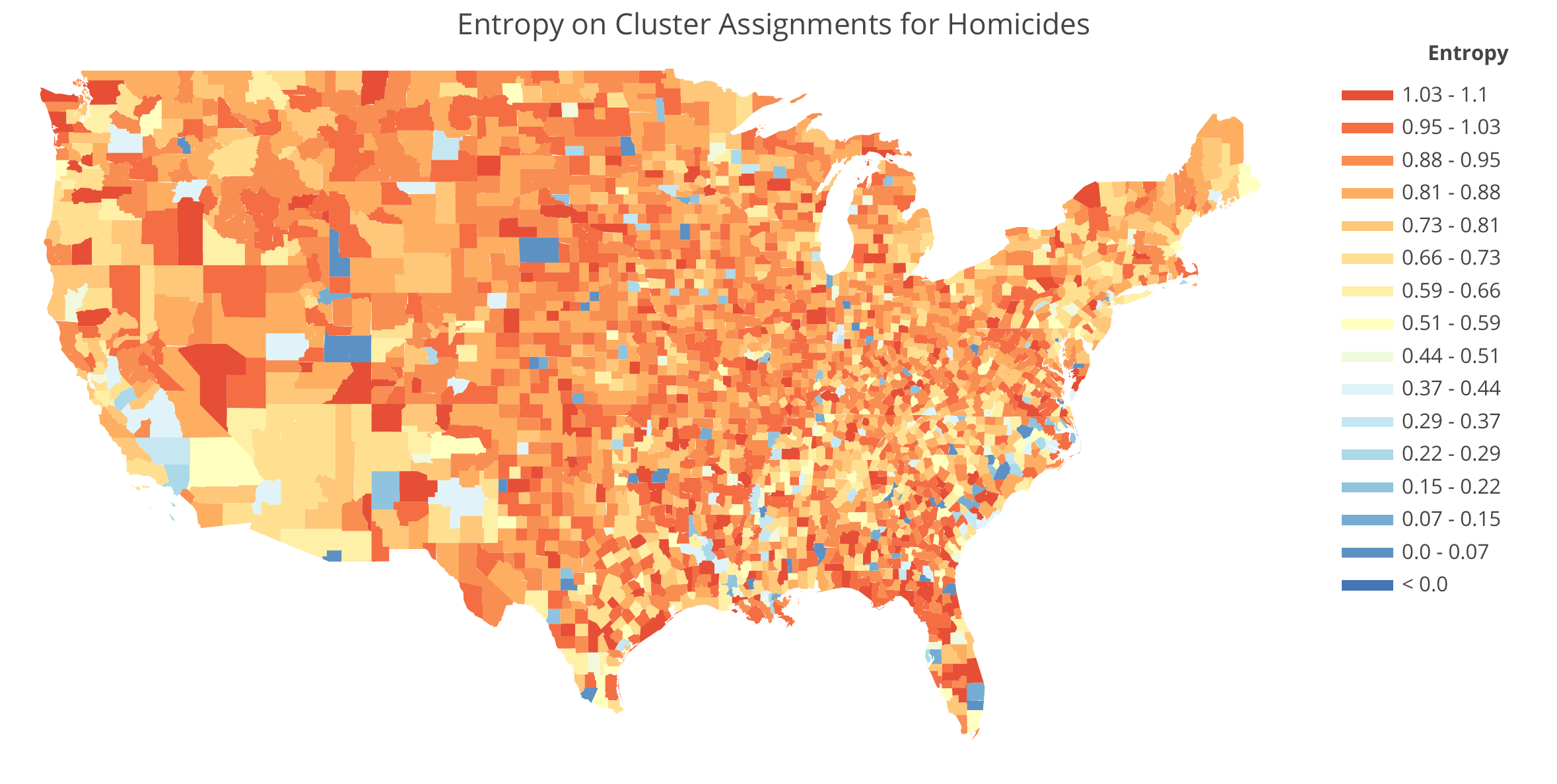}
		\end{subfigure}
		\caption{\small \textbf{Entropy on mapping assignments} $\bbH(z_i \given x_i; \phi)$.  Higher entropies corresponds to more uniform posterior distributions for the mapping assignment, indicating that auxiliary data can be reasonably integrated using \textit{any} of the learned conjugate mappings. 
		}
		\label{fig-geomap-entropy-mapping}
	\end{center}
	\vspace*{-5mm}
\end{figure}

The structured nature of the generative model enables detailed analysis over the way in which auxiliary data impacts the posterior distribution of the per-capita homicide rates.  For example, we show the entropy $\bbH(z_i \given x_i, y_i; \phi)$ of the conjugate mapping assignments in Figure~\ref{fig-geomap-entropy-mapping}.  Higher entropies correspond to more uniform distributions $p(z_i \given x_i, y_i; \phi)$; lower entropies are much more concentrated at the MAP value $z_i^\textrm{MAP}$.  For a three-component mixture model the maximum achievable entropy (in nats) is approximately 1.1, corresponding to a uniform distribution over the three possible mapping assignments.  This occurs for a minority of counties; for the vast majority the three-component mixture provides at least some discriminative power.

\begin{figure}[t!]
	\vskip 0.1in
	\begin{center}
		\begin{subfigure}[b]{\textwidth}
			\centering
			\includegraphics[width=1.0\linewidth, trim={0cm 0cm 0cm 0cm},clip]{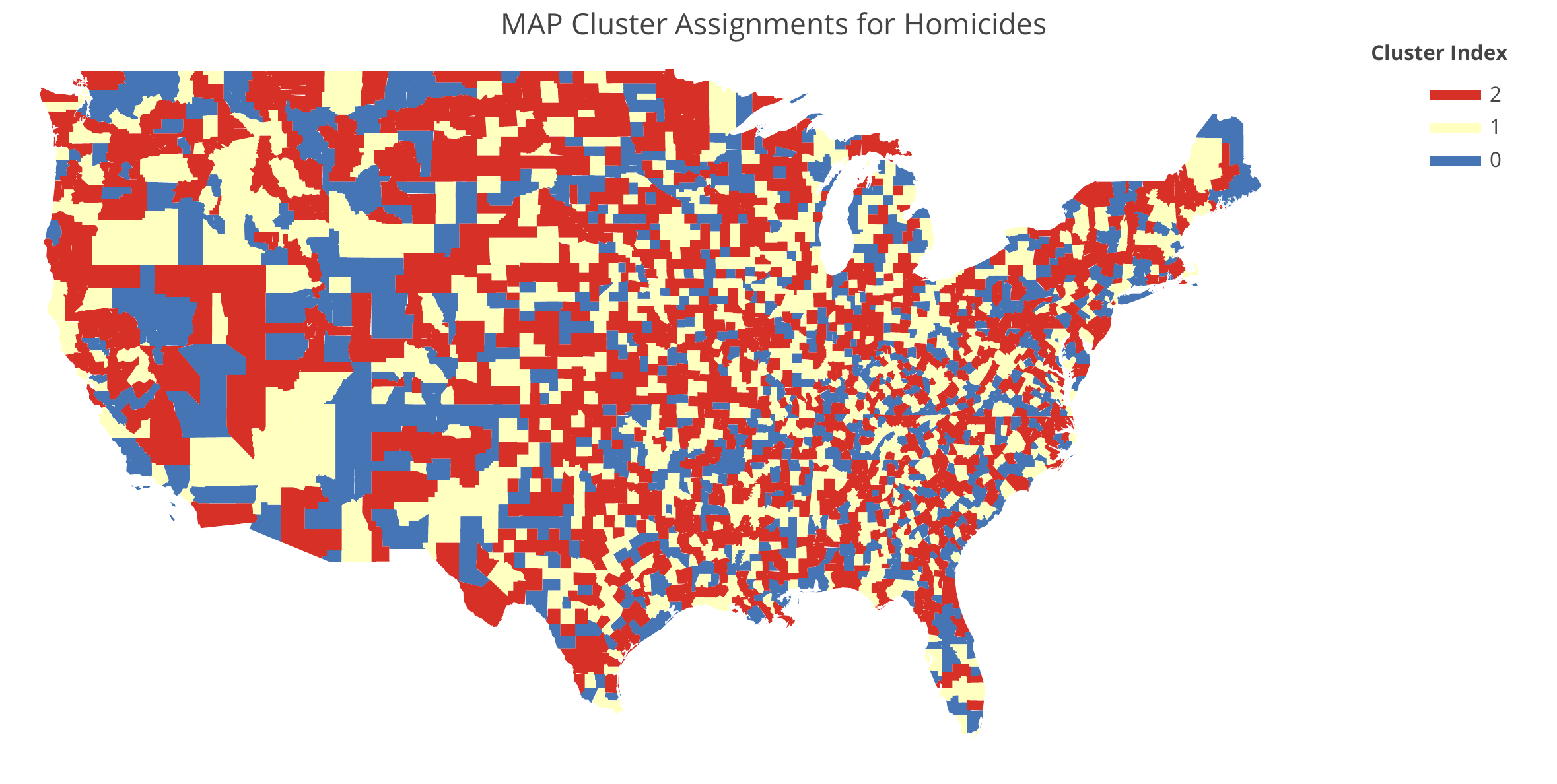}
		\end{subfigure}
		\caption{\small \textbf{MAP conjugate mapping assignments}, showing the mapping assignment with highest posterior probability for each of county.}
		\label{fig-geomap-map-clus-assign}
	\end{center}
	\vspace*{-5mm}
\end{figure}

Figure~\ref{fig-geomap-map-clus-assign} shows the maximum a posteriori (MAP) mapping assignments $z_i^\textrm{MAP}$ for every county.  The clusters are ordered from from fewest (0, blue) to most (2, red) assigned counties.  Large population centers in the U.S. predominantly appear in the white and red clusters, although both of those county types are also the assignment of many rural counties.
This suggests that the discriminative power obtained by the conjugate mapping mixture model is arising from factors other than the direct population.

\subsubsection*{Superior predictive accuracy and model checking}

Our formulation enables inference over the primary data posterior predictives $p(y_i \given x_i; \phi)$ that are critical for assessing model fit and for model checking.

\begin{figure}[t!]
	\vskip 0.1in
	\begin{center}
		\begin{subfigure}[b]{0.44\textwidth}
			\centering
    		\includegraphics[width=\linewidth, trim={0cm 0cm 0cm 0cm},clip]{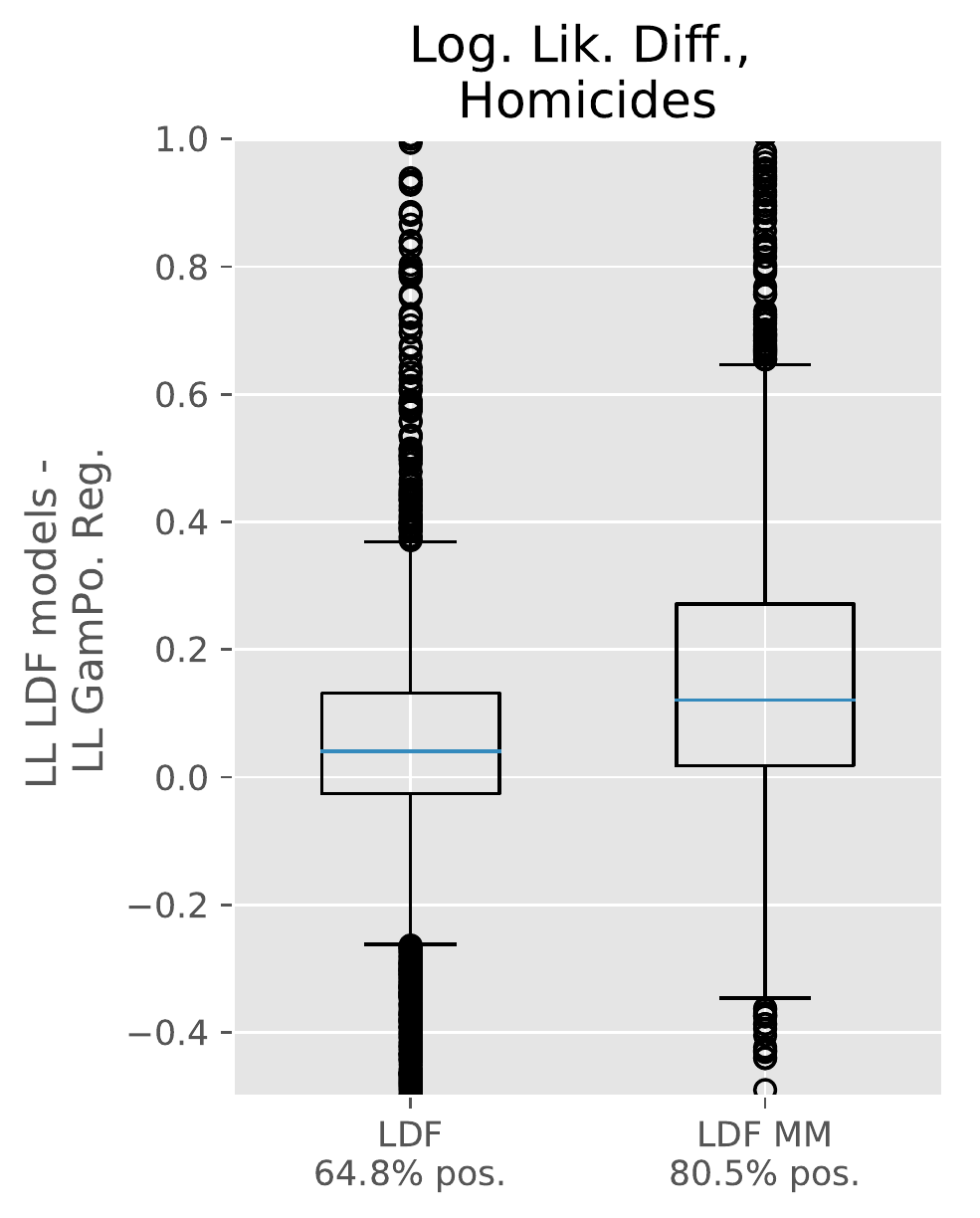}
		\end{subfigure}
		\begin{subfigure}[b]{0.55\textwidth}
			\centering			
			\includegraphics[width=\linewidth, trim={0cm 0cm 0cm 0cm},clip]{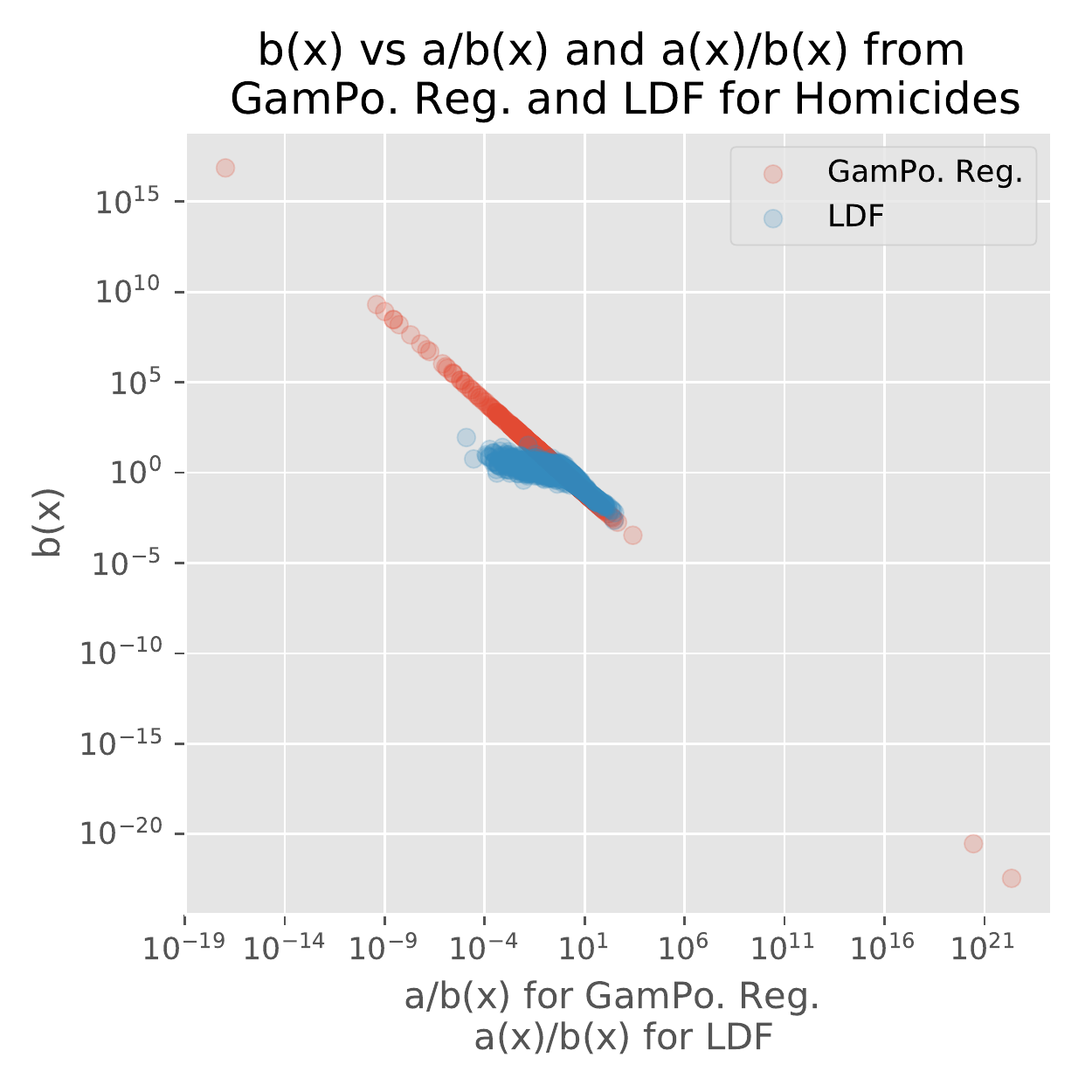}			
		\end{subfigure}	
		\caption{\small 
		\textbf{(Left)} \textbf{Difference in log-likelihood}. Box-plots show difference in log-likelihoods between LDF and LDF MM relative to the gamma-Poisson regression baseline.
		\textbf{(Right)} \textbf{learned parameter scatter plots by county}. Scatter plots showing learned parameters $a(x;\phi)$ vs $a(x;\phi)/b$ and $a(x;\phi)/b(x;\phi)$ from gamma-Poisson regression and LDF.
		}
		\label{fig-diff-ll-box-plots}
	\end{center}
	\vspace*{-5mm}
\end{figure}

Our models provide a much better fit of the primary data than the baselines for both the single conjugate mapping (LDF) and conjugate mapping mixture model (LDF-MM) case. Figure~\ref{fig-diff-ll-box-plots} (left) shows that for homicides our LDF approaches achieve a better held-out log-likelihood for $64.8\%$ (single conjugate mapping) and $80.5\%$ (mixture of conjugate mappings) of the counties.

\begin{figure}[t!]
	\vskip 0.1in
	\begin{center}
		\begin{subfigure}[b]{\textwidth}
			\centering
			\includegraphics[width=1.0\linewidth, trim={0cm 0cm 0cm 0cm},clip]{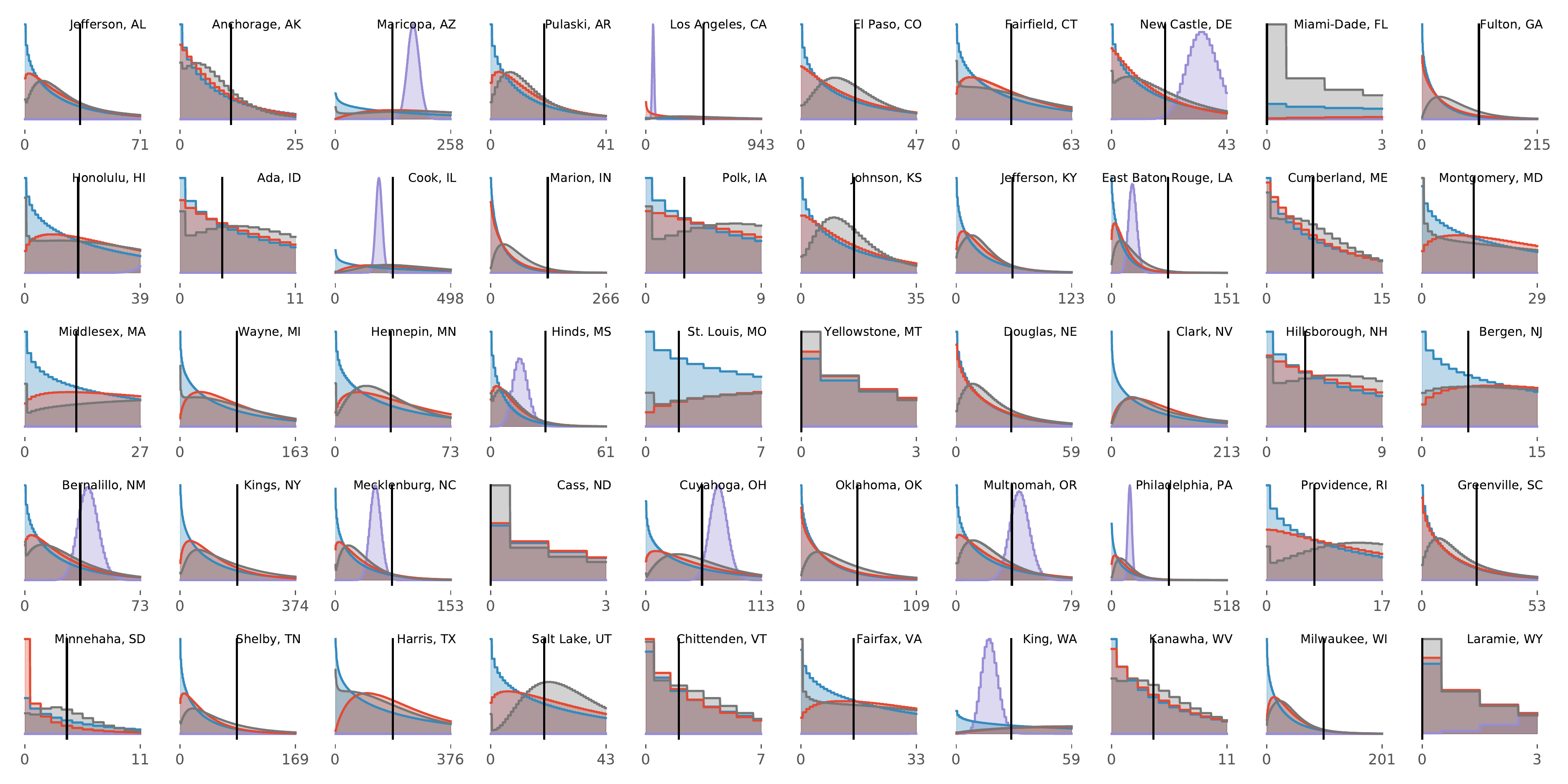}
		\end{subfigure}
		\caption{\small  \textbf{Sample predictive distributions}. Predictive distributions $p(y_i|x_i;\phi)$ on homicide rates are shown for one county per state. 
		Distributions corresponding to LDF are in red, LDF MM in gray, gamma-Poisson regression in blue, and Poisson regression in purple. The empirical rate is shown by the black line.
		The counties displayed are those with the largest population for the given state. 
		} 
	\label{fig-predictives-homicide}
\end{center}
\vspace*{-5mm}
\end{figure}

Model checking is enabled by posterior predictive checks.
Figure~\ref{fig-predictives-homicide} provides posterior predictive distributions, showing the observed number of homicides for the most populous counties in each state.

\subsubsection*{Interpretable mappings of auxiliary data}

For Gamma-Poisson models our LDF approach maps the auxiliary data for each county into two positive scalars: the number of pseudo-occurrences $a(x_i; \phi_k)$ and the number of pseudo-multiples of exposure $n_i$, $b(x_i; \phi_k)$.
The ratio of these is the empirical rate, $a(x_i; \phi_k) / b(x_i; \phi_k)$, a quantity closely related to the mean of the posterior distribution over the homicide rate.
Similarly, the number of pseudo-multiples of exposure $b(x_i; \phi_k)$ is inversely proportional to the posterior variance, so greater exposure conveys higher information content in the auxiliary data.
Because the mean and the variance of our posterior predictive distribution are decoupled through the transformations $a(x_i; \phi_k)$ and $b(x_i; \phi_k)$, we observe that our model has the flexibility to provide accurate models of both predictions and uncertainty.

The Gamma-Poisson regression baseline learns an empirical rate $a(x_i; \phi_k) / b(x_i; \phi_k)$ and a fixed over-dispersion parameter, $\delta$, that is shared across all data points.  
For more details see the supplement.
The Poisson regression is omitted from this comparison as it performed poorly.

Consider, for example, the parameters for predicting homicide rate from Figure~\ref{fig-diff-ll-box-plots} (right).  
Each point represents a specific county and conjugate mapping corresponding to $z_i^\textrm{MAP}$ is used.
For counties corresponding to a relatively low empirical rate (e.g. $a(x_i; \phi) / b(x_i; \phi) < 10^{0}$) the corresponding degree of pseudo-exposure is relatively low, less than $10^{2}$.
This enables the model to learn that it should be \textit{more} uncertain about the occurrence rate for these homicides---a reasonable, if not desirable, trait since the homicides are relatively rare.  In contrast, for the baseline a lower empirical rate is inversely proportional to the number of intervals, resulting in much more overconfident posterior distributions.
Additionally, the empirical rate is linearly related to the degree of pseudo-exposure, as evidenced by the collinearity of the points.


\section{Discussion}
\label{sec-discussion}


We have presented LDF, a methodology for fusing multiple data types that enables downstream posterior inference, reasoning, and uncertainty quantification.
We consider the setting in which some measurements, primary data, have well-characterized forward models and others, auxiliary data, have forward models that are unknown and/or complex.
Central to our approach is the idea of a \textit{conjugate mapping}, which renders the auxiliary data into the form of sufficient statistics of the primary data and enables conjugate updates.
Unlike many approaches, LDF enables learning from the auxiliary data \textit{without} direct access to the underlying latent variable values by relying on the primary data to provide noisy or stochastic labels.

LDF is general and leverages the power of NNs to model complex relationships between the data sources and the well-studied inference algorithms and interpretable posteriors that accompany structured PGMs. 
This method can be applied to all exponential family models and is demonstrated on real-world applications illustrating its flexible framework and robustness to sparse and complex data.  
Additionally, LDF readily extends to a variety of additional settings not explored here.
We have restricted our examples to data sets of relatively small $N$ and learned correspondingly small NN architectures; in large-data regimes even more expressive networks can be learned for extracting sufficient statistics from the auxiliary data.
Although we demonstrate LDF using feed-forward NN architectures the specification of conjugate mappings is general and readily accepts effectively any NN architecture, e.g. long short-term memory (LSTM) networks and other recurrent NNs for capturing temporal dependencies or convolutional NNs to accommodate image-based auxiliary data.
Further, while we have shown MCMC-based inference procedures throughout nothing precludes the use of variational inference methods.
Finally, although our exposition is predicated on the use of conjugate prior distributions our technique and philosophy can still be applied in non-conjugate primary data models.

The methodology of LDF enables efficient inference and is robust to sparse, noisy observations and applicable to many settings that would preclude the use of traditional supervised learning approaches.
By construction LDF facilitates efficient inference and, unlike purely predictive models, the Bayesian formulation enables accurate uncertainty quantification.
Finally, LDF learning can be implemented in modern deep learning frameworks using built-in methods, enabling straightforward application to some of the most challenging problems across a range of disciplines.


\clearpage
\appendix

\section*{Appendix A. Example Mappings for Common Primary Data Types}
\newcommand{\myAppendixPrefix}{A}
\renewcommand{\theequation}{\myAppendixPrefix\arabic{equation}} 
\setcounter{equation}{0} 
\label{sec-appendix-model-types}

Here we list common types of exponential family primary data and the corresponding class of conjugate mapping. 
We provide the auxiliary data posterior $p(\theta_i \given x_i; \phi)$, the primary data posterior predictive likelihood $p(y_i \given x_i)$, and a description of mapping formulations.
These statements follow from conjugacy and inspection of how the primary data sufficient statistics impact the natural parameters of the primary data posterior $p(\theta_i \given y_i)$.  

\textbf{Bernoulli/binomial with known number of trials.}
Consider primary data $p(y_i \given \theta_i) = \binomialpmf(y_i; n_i, \theta_i)$ where $n_i$ is the number of trials.  The conjugate prior over the success probability is beta with parameters $\alpha_0, \beta_0 \geq 0$.  A conjugate mapping of auxiliary data yields
\begin{align}
p(\theta_i \given x_i; \phi) 
    &= \betapdf(\theta_i; \alpha_0 + a(x_i; \phi), \beta_0 + b(x_i; \phi))  \\
p(y_i \given x_i; \phi)
    &= \betabinomialpmf(y_i; n_i, \alpha_0 + a(x_i; \phi), \beta_0 + b(x_i; \phi))
\end{align}
where $a(x_i; \phi), b(x_i; \phi) \geq 0$ are scalars acting as a number of pseudo-successes and -failures, respectively.  These quantities may be derived from an alternative formulation where $n(x_i; \phi) \geq 0$ is the number of pseudo-trials and $\mu(x_i; \phi) \in (0, 1)$ is the pseudo-mean obtained from the auxiliary data, with $a(x_i; \phi) = n(x_i; \phi) \mu(x_i; \phi)$ and $b(x_i; \phi) = n(x_i; \phi) (1 - \mu(x_i; \phi))$.

\textbf{Categorical/multinomial with known number of trials.}
The binomial example above is a special case of multinomial primary data.  Consider primary data $p(y_i \given \theta_i) = \multinomialpmf(y_i; n_i, \theta_i)$ where $n_i$ is the number of trials.  The conjugate prior over the success probability is Dirichlet with concentration parameters $\alpha_0 \in \bbR_{\geq 0}^D$.  A conjugate mapping of auxiliary data yields
\begin{align}
p(\theta_i \given x_i; \phi) 
    &= \dirichletpdf(\theta_i; \alpha_0 + a(x_i; \phi))  \\
p(y_i \given x_i; \phi)
    &= \dirichletmultinomialpmf(y_i; n_i, \alpha_0 + a(x_i; \phi))
\end{align}
where $a(x_i; \phi) \in \bbR_{\geq 0}^D$ act as a number of pseudo-counts for each class.
Alternatively this vector may be derived from a pseudo-trials $n(x_i; \phi) \in \bbR_{\geq 0}$ and a pseudo-mean vector $\mu(x_i; \phi) \in \Delta^{D-1}$ on the $d$-dimensional probability simplex.  Pseudo-counts follow by $a(x_i; \phi) = n(x_i; \phi) \mu(x_i; \phi)$.

\textbf{Poisson with known exposure.}
Consider primary data $p(y_i \given \theta_i) = \poissonpmf(y_i; n_i \theta_i)$ where $n_i$ is the exposure of the $i$-th data point.  A conjugate prior over the rate $\theta_i \in \bbR_{\geq 0}$ is Gamma with shape and rate parameters $\alpha_0, \beta_0 \geq 0$.  A conjugate mapping of auxiliary data provides
\begin{align}
p(\theta_i \given x_i) 
    &= \gammapdf(\theta_i; \alpha_0 + a(x_i; \phi), \beta_0 + b(x_i; \phi))  \\
p(y_i \given x_i)
    &= \gammapoissonpmf(y_i; \alpha_0 + a(x_i; \phi), n_i^{-1} (\beta_0 + b(x_i; \phi)))
\end{align}
where $a(x_i; \phi) \geq 0$ transforms $x_i$ to a number of arrivals occurring within $b(x_i; \phi) \geq 0$ intervals.

\textbf{Linear Gaussian with known covariance.}
Consider primary data $p(y_i \given \theta_i) =  \calN(y_i; A\theta_i + b, \Sigma^{-1})$.  The conjugate prior over the mean $\theta_i \in \bbR^D$ is also Gaussian with mean $\mu_0 \in \bbR^D$ and covariance $J_0^{-1} \in \bbS^D_+$ (the $D$-dimensional positive semi-definite cone).  A conjugate mapping of the auxiliary data yields the posterior and predictive distributions
\begin{align}
p(\theta_i \given x_i; \phi) 
    &= \calN(\theta_i; J^{-1} h, J^{-1}) \\
p(y_i \given x_i; \phi)
    &= \calN(y_i; A J^{-1} h + b, A J^{-1} A^\top + \Sigma^{-1})
\end{align}
where $h = J_0 \mu_0 + h(x_i; \phi)$ and $J = J_0 + L(x_i; \phi) L(x_i; \phi)^\top$.  That is, we learn a mapping from the auxiliary data into a potential vector $h(x_i; \phi) \in \bbR^D$ and the $D(D+1)/2$ nonzero elements of a lower-triangular matrix $L(x_i; \phi) \in \bbR^{D\times D}$.

\section*{Appendix B. EM for Conjugate Mapping Mixture Models}
\renewcommand{\myAppendixPrefix}{B}
\renewcommand{\theequation}{\myAppendixPrefix\arabic{equation}} 
\setcounter{equation}{0} 
\label{sec-appendix-mixture-em}

Here we describe our EM learning approach used to learn point estimates of the mapping parameters $\phi = (\phi_1, \ldots \phi_K)$ and mapping weights $\beta \in \Delta^{K-1}$.  Our model contains auxiliary data $x = (x_1, \ldots, x_N)$, primary data $y = (y_1, \ldots, y_N)$ with an exponential family likelihood $p(y_i \given \theta_i) = f(y_i; \theta_i)$ and a conjugate prior $p(\theta_i \given x_i, z_i; \phi) = \pi(\theta_i; \lambda_0 + T_\sfx(x_i; \phi_{z_i}))$ where mapping assignments $p(z_i \given \beta) = \categoricalpmf(z_i; \beta)$.
Under these circumstances the marginal is generally tractable:
\begin{align}
    p(y_i \given x_i, z_i; \phi)
    = \int f(y_i; \theta_i) \pi(\theta_i; \lambda_0 + T_\sfx(x_i; \phi_{z_i})) \dif\theta_i
    \triangleq g(y_i; \lambda_0 + T_\sfx(x_i; \phi_{z_i})).
\end{align}

EM is an iterative procedure that converges to a local maximum of the marginal data likelihood, $p(y \given x; \phi, \beta)$ where we have marginalized out the \textit{hidden variables}---the mapping assignments $z = (z_1, \ldots, z_N)$.
Here it is useful to consider $z_i$ to be in one-hot-vector notation instead of indicator variable representation.  We define the \textit{complete} data log-likelihood:
\begin{align}
\log p(y, z \given x; \phi, \beta)
	&= \log \prod_{i=1}^N p(z_i \given \beta) p(y_i \given x_i, z_i; \phi)
	= \log \prod_{i=1}^N \prod_{k=1}^K \left\{\beta_k p(y_i \given x_i, z_i; \phi)\right\}^{z_{ik}} \\
	&= \sum_{i=1}^N \sum_{k=1}^K z_{ik} \log \beta_k + \sum_{i=1}^N \sum_{k=1}^K  z_{ik} \log g(y_i; \lambda_0 + T_\sfx(x_i; \phi_{z_i})).
\end{align}

The two steps of EM are:
\begin{itemize}
    \item (Expectation) E-step:  Define $\calQ(\phi, \beta \given \phi^{(t)}, \beta^{(t)})
\triangleq \bbE_{p(z \given x, y; \phi^{(t)}, \beta^{(t)})}\left[\log p(y, z \given x; \phi, \beta)\right]$;
    \item (Maximization) M-step: Maximize $\calQ(\phi, \beta \given \phi^{(t)}, \beta^{(t)})$ w.r.t. $\phi$ and $\beta$.
\end{itemize}
In our setting, the E-step computes the conditional $p(z \given x, y; \phi, \beta)$ which factors here over the $z_i$.  We compute \textit{responsibilities},
\begin{align}
\rho_{ik}^{(t)}
	&\triangleq p(z_i = k \given x_i, y_i; \phi^{(t)}, \beta^{(t)}) \\ 
	&\propto p(z_i = k; \beta^{(t)}) p(y_i \given x_i, z_i=k; \phi^{(t)}) \\
	&= \beta_k^{(t)} g(y_i; \lambda_0 + T_\sfx(x_i; \phi_k^{(t)}))
	\label{eq-em-condx-gamma}
\end{align}
for all $i = 1, \ldots, N$ and $k = 1, \ldots, K$.
The M-step then computes for all $k = 1, \ldots, K$:
\begin{align}
\phi_k^{(t+1)}, \beta_k^{(t+1)}
    &= \argmax_{\phi_k, \beta_k} \calQ(\phi, \beta \given \phi^{(t)}, \beta^{(t)}) \\
    &= \argmax_{\phi_k, \beta_k} \bbE_{p(z \given x, y; \phi^{(t)}, \beta^{(t)})}\left[\log p(y, z \given x; \phi, \beta)\right] \\
    &= \argmax_{\phi_k, \beta_k} \sum_{i=1}^N \sum_{k=1}^K \rho_{ik}^{(t)} \log \beta_k 
    + \sum_{i=1}^N \sum_{k=1}^K \rho_{ik}^{(t)} \log g(y_i; \lambda_0 + T_\sfx(x_i; \phi_k)).
\end{align}
This decomposes into two maximization problems,
\begin{align}
\beta_k^{(t+1)} 
	&= \arg\max_{\beta \in \Delta^{K-1}} \sum_{i=1}^N \rho_{ik}^{(t)} \log \beta_k
	\Longrightarrow
\beta_k^{(t+1)}
	= \dfrac{\sum_{i=1}^N \rho_{ik}^{(t)}}{\sum_{k=1}^K \sum_{i=1}^N \rho_{ik}^{(t)}}, \\
\phi_k^{(t+1)} &= \arg\max_{\phi} \sum_{i=1}^N \rho_{ik}^{(t)} \log g(y_i; \lambda_0 + T_\sfx(x_i; \phi_k)) 
\end{align}
with the latter solved numerically. These values are used to compute $\rho_{ik}^{(t+1)}$, and so on.
The iterative procedure terminates when the complete data log likelihood $\log p(y, z \given x; \phi, \beta)$ stops improving and the current value of $\phi$ is used for inference.  The mapping weights $\beta$ can also be used if desired.

When the conjugate mappings $T_\sfx(x_i; \phi_k)$ are NNs our EM approach appears similar to Neural EM \citep{Greff2017}.
While the M-step of their approach takes only a single step in the direction of the gradient, we fully optimize the expected log likelihood at each iteration.

\section*{Appendix C. Electrification Inference Supplement}
\renewcommand{\myAppendixPrefix}{C}
\renewcommand{\theequation}{\myAppendixPrefix\arabic{equation}} 
\setcounter{equation}{0} 
\label{sec-electrification-supp}
\suppressfloats

In this section we provide additional information and results pertaining to the electrification application.

\subsection*{Data}

\begin{figure}[t]
  \vskip 0.1in
  \begin{center}
    \centerline{\includegraphics[width=0.8\linewidth, trim={0.375cm 0.625cm 0.5cm 0.125cm},clip]{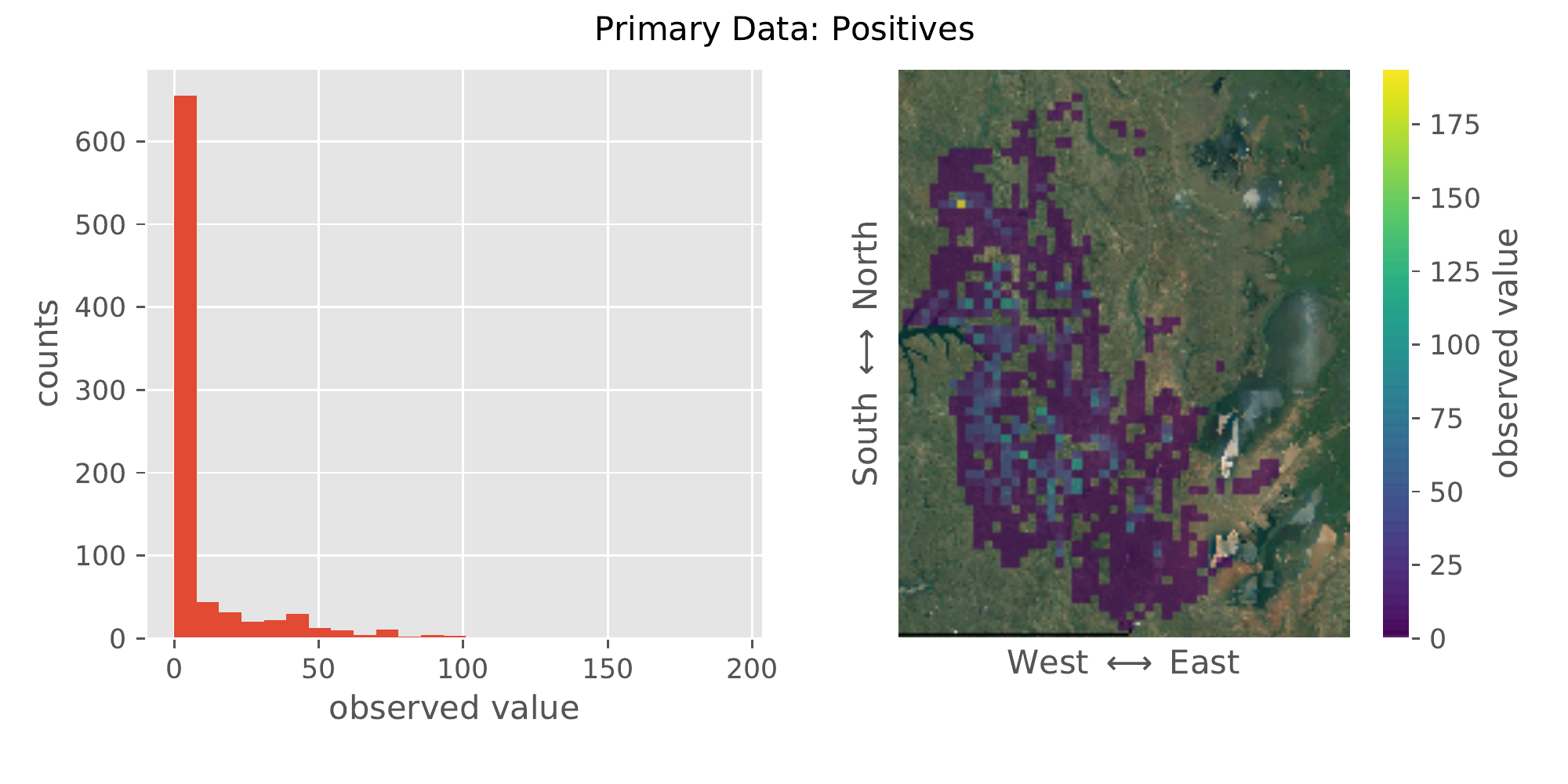}}
    \centerline{\includegraphics[width=0.8\linewidth, trim={0.375cm 0.625cm 0.5cm 0.125cm},clip]{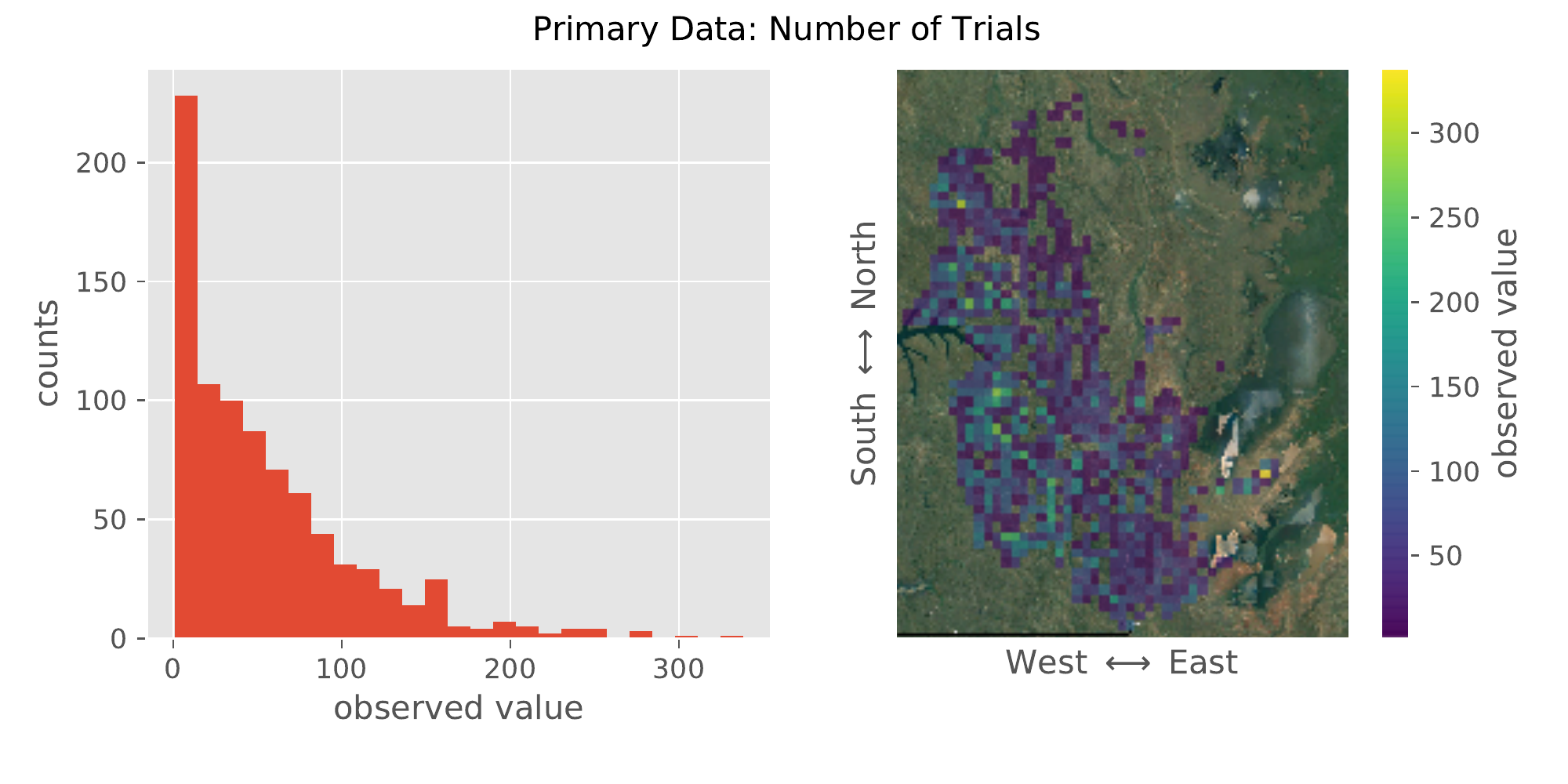}}
    \centerline{\includegraphics[width=0.8\linewidth, trim={0.375cm 0.625cm 0.5cm 0.125cm},clip]{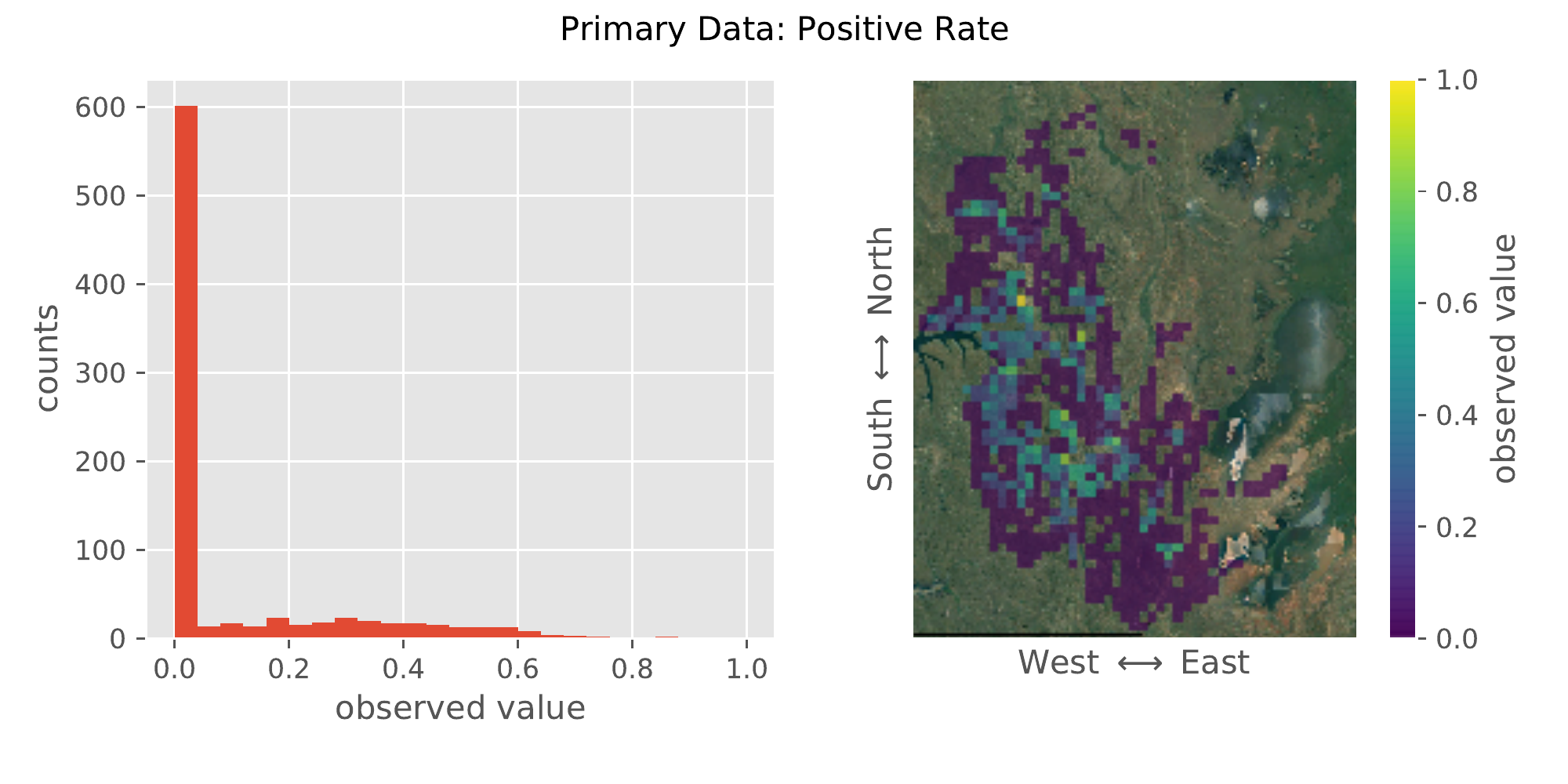}}
    \caption{\small \textbf{Statistics of the primary data}, with model $Y_i \sim \textrm{Bin}(N_i, \theta_i)$.  Histograms and maps are only shown for the $M=784$ regions where $N_i > 0$. (top) histogram and map of the number of successes $Y_i$; (middle) histogram and map of the number of trials $N_i$; (bottom) histogram and map of the empirical estimate of the latent variable $\hat{\theta}_i = Y_i / N_i$.}
    \label{fig-supp-elec-primary-data-hist}
  \end{center}
  \vspace*{-5mm}
\end{figure}

\begin{figure}[t]
  \vskip 0.1in
  \begin{center}
    \centerline{\includegraphics[width=0.7\linewidth, trim={0.375cm 0.625cm 0.5cm 0.125cm},clip]{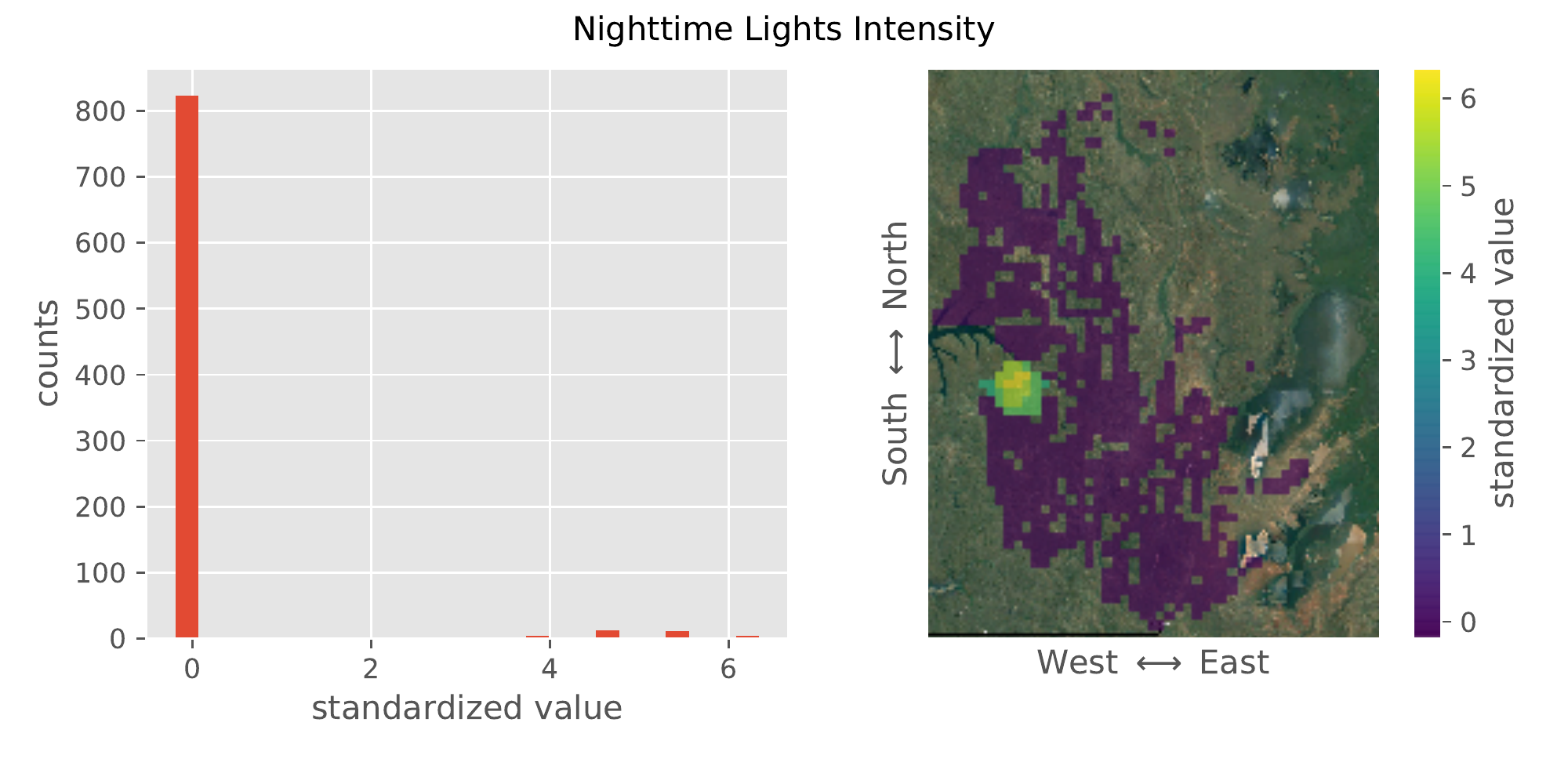}}
    \centerline{\includegraphics[width=0.7\linewidth, trim={0.375cm 0.625cm 0.5cm 0.125cm},clip]{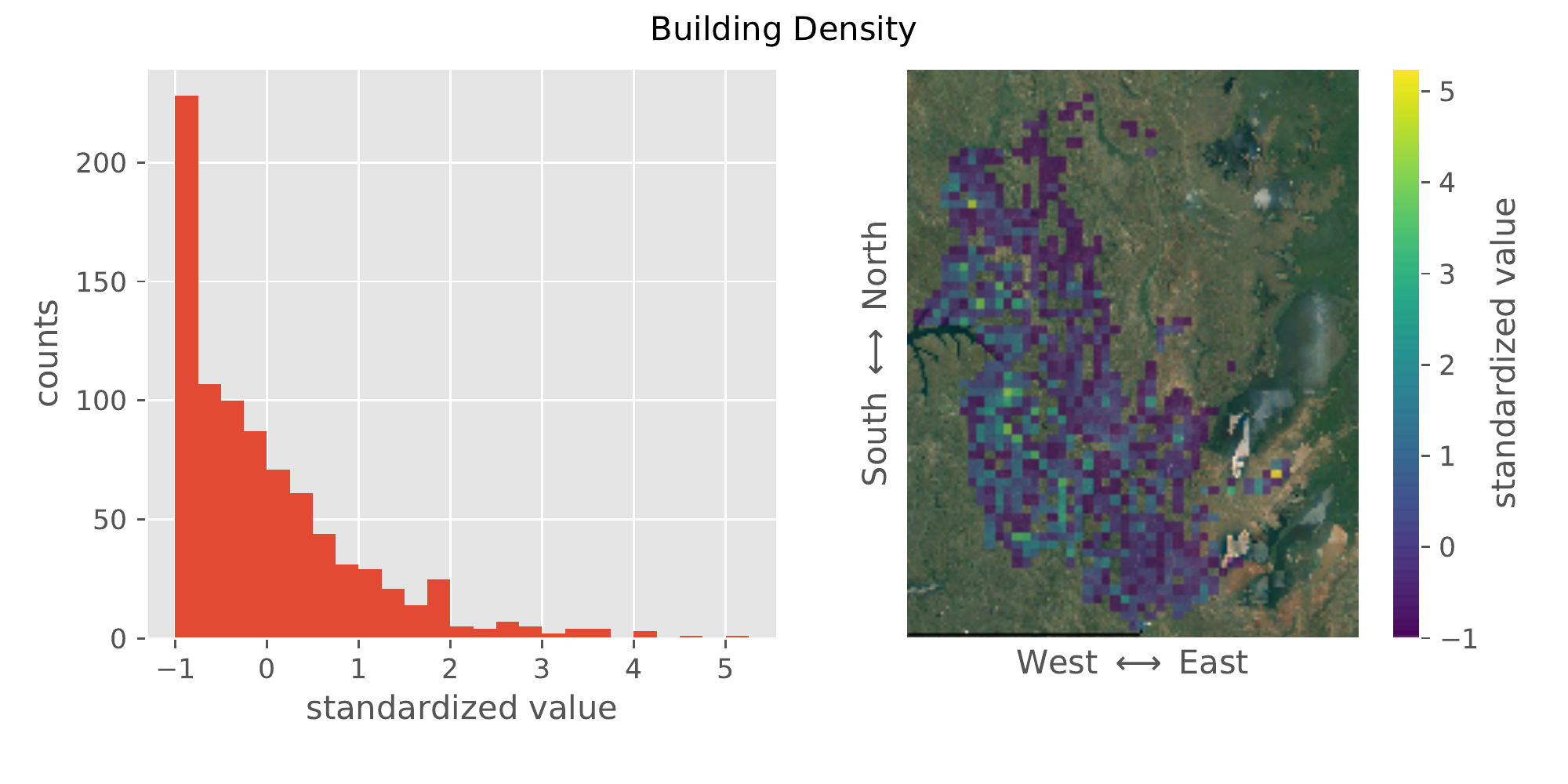}}
    \centerline{\includegraphics[width=0.7\linewidth, trim={0.375cm 0.625cm 0.5cm 0.125cm},clip]{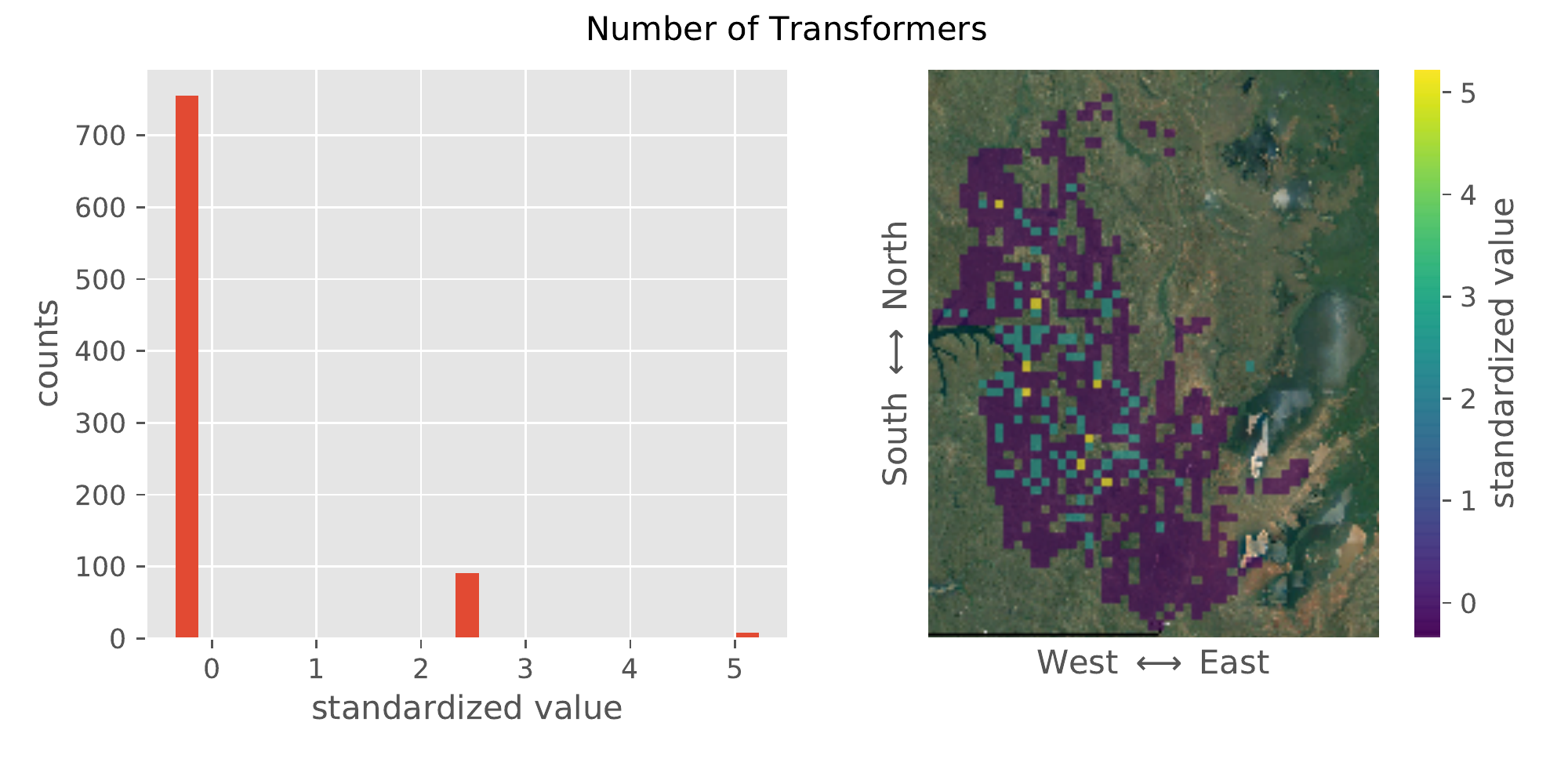}}
    \centerline{\includegraphics[width=0.7\linewidth, trim={0.375cm 0.625cm 0.5cm 0.125cm},clip]{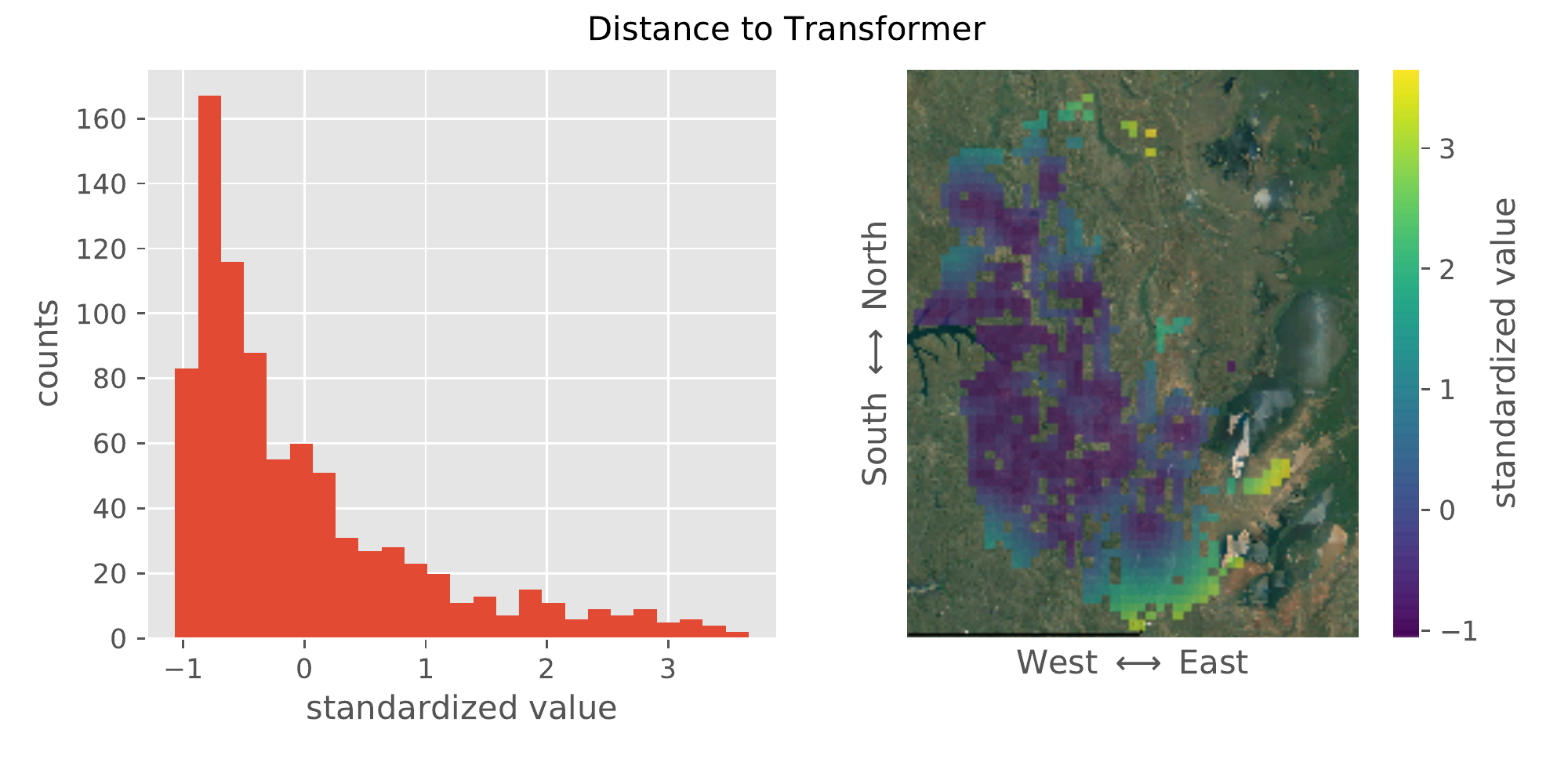}}
    \caption{\small \textbf{Statistics of the auxiliary data}, standardized to mean 0 and standard deviation 1.  For each we show a histogram and map.}
    \label{fig-supp-elec-auxiliary-data-hist}
  \end{center}
  \vspace*{-5mm}
\end{figure}

Primary and auxiliary data measurements appear in Figures~\ref{fig-supp-elec-primary-data-hist} and \ref{fig-supp-elec-auxiliary-data-hist}.  Refer to the main text for more details on sources, interpretations, and processing.

\subsection*{Baseline Approaches}
\label{sec-appendix-elec-baselines}

Binomial and beta-binomial regression, in addition to conforming to the known likelihood $p(y_i \given \theta_i) = \binomialpmf(y_i \given n_i, \theta_i)$ where $n_i$ is known, also model the posterior distribution over $p(\theta_i \given x_i; \phi)$ and the parameters $\phi$ are fit by maximize the log predictive likelihood (or a similar criterion),
\begin{align}
p(y \given x; \phi) = \prod_{i=1}^M p(y_i \given x_i; \phi)
 = \prod_{i=1}^M \int p(y_i \given \theta_i) p(\theta_i \given x_i; \phi) \dif\theta_i
\end{align}
Baseline approaches focus on learning a single transformation
\begin{align}
\mu(x_i; \phi) = \textrm{sigm}(\phi^\top x_i) = (1 + \exp(-\phi^\top x_i))^{-1}
\end{align} 
where the inner product $\phi^\top x_i$ additionally includes a bias term that is omitted for brevity.  Beta-binomial regression additionally learns a sample size parameter $\delta$.  The decomposition of the learning objective for our baseline approaches into the above integral reveals the latent variable posteriors $p(\theta \given x_i)$ that are learned by the baselines, which we can examine relative to our approach.

Binomial regression implies a Dirac posterior.  This yields a point estimate of the latent parameters with an implied uncertainty of zero.  Clearly this representation does not adequately capture posterior uncertainty.  
\begin{align}
p_\textrm{Bin}(y_i \given x_i; \phi) &= \textrm{Bin}(y_i \given n_i, \mu(x_i; \phi)) \\
p_\textrm{Bin}(\theta_i \given x_i; \phi) &= \text{Dirac}(\theta_i - \mu(x_i; \phi))
\end{align}
Further, binomial regression lacks a closed-form expression for the full posterior $p_\textrm{Bin}(\theta_i \given x_i, y_i; \phi)$.  
Beta-binomial regression, on the other hand, embeds a conjugate (w.r.t. $p(y_i \given \theta_i)$) beta prior distribution with mean $\mu(x_i; \phi)$.  The learned parameter $\delta$ is inversely proportional to the posterior uncertainty and is \textit{shared} across all posteriors $p_\textrm{BB}(\theta_i \given x_i; \phi)$.
\begin{align}
p_\textrm{BB}(y_i \given x_i; \phi) &= \textrm{BetaBin}(y_i \given n_i, \alpha_0 + \delta \mu(x_i; \phi), \beta_0 + \delta (1-\mu(x_i; \phi))) \\
p_\textrm{BB}(\theta_i \given x_i; \phi) &= \textrm{Beta}(\theta_i \given \alpha_0 + \delta \mu(x_i; \phi), \beta_0 + \delta (1-\mu(x_i; \phi)))
\end{align}
In contrast, our method provides the flexibility for each posterior to have an auxiliary-data-dependent degree of uncertainty driven by $a(x_i; \phi) + b(x_i; \phi)$.
\begin{align}
p_\textrm{LDF}(y_i \given x_i; \phi) &= \textrm{BetaBin}(y_i \given n_i, \alpha_0 + a(x_i; \phi), \beta_0 + b(x_i; \phi)) \\
p_\textrm{LDF}(\theta_i \given x_i; \phi) &= \textrm{Beta}(\theta_i \given \alpha_0 + a(x_i; \phi), \beta_0 + b(x_i; \phi))
\end{align}
Binomial regression is a limiting case of beta-binomial regression in which the sample size $\delta \rightarrow \infty$, resulting in the posterior collapsing into a degenerate distribution with mass at $\mu(x_i; \phi)$ only.
Further, beta-binomial regression is a special case of our approach where $a(x_i; \phi) = \delta \mu(x_i; \phi)$, and $b(x_i; \phi) = \delta (1 - \mu(x_i; \phi))$.

The optimal L2 regularization weights for Binomial and Beta-Binomial regression: of $0.00123$ and $5.214 \times 10^{-5}$, respectively.  These were identified by randomly-sampling configurations on a log-scale, using the same hyper-parameter tuning procedure as our method.



\section*{Appendix D. Homicide Rate Inference Supplement}
\renewcommand{\myAppendixPrefix}{D}
\renewcommand{\theequation}{\myAppendixPrefix\arabic{equation}} 
\setcounter{equation}{0} 
\label{sec-crime-supp}
\suppressfloats

\subsection*{Data}

The specific socio-demographic features used, as listed by the US Census Bureau's 2011 Reference Information Files, are shown in Table \ref{table:socios1} and \ref{table:socios2}'s ``feature'' column. 
Features and whether they are transformed to per-capita units is shown in Tables \ref{table:socios1} and \ref{table:socios2}. The per-capita transformation is performed by dividing the given feature of interest by the population feature, ``Resident population (April 1 - complete count) 2010.'' Determinations were made to make per-capita transformations based on whether the feature was being presented on a county-wide count basis or as a percent by the 2010 Census. Counts-denominated features were transformed while percent-denominated features were not. Additionally, the feature ``Resident population (April 1 - complete count) 2010'' was provided without any processing. 

\begin{table}[!htb]

\begin{center}
		\begin{tabular}{|p{1.5cm} | p{9cm} | p{2.5cm}|} 
		\hline
		\textbf{ID} & \textbf{Feature Description} & \textbf{Provided Per-Capita} \\ 
		\hline		
		1 & Resident population (April 1 - complete count) 2010 & False \\ 
		\hline
		2 & Civilian labor force unemployment 2010 & True \\ 
		\hline
		3 & People of all ages in poverty - percent 2009 & False \\ 
		\hline
		4 & Land area in square miles 2010 & False \\ 
		\hline
		5 & Male population 2010 & True \\ 
		\hline
		6 & Educational attainment - persons 25 years and over - percent bachelor's degree or higher 2005-2009 & False \\ 
		\hline
		7 & Households with Food Stamp/SNAP benefits in the past 12 months, total 2005-2009 & True \\ 
		\hline
		8 & Components of change - net international migration for July 1, 2008 to July 1, 2009 & True \\ 
		\hline
		9 & Components of change - cumulative estimates - net domestic migration, April 1, 2000 to July 1, 2009 & True \\ 
		\hline
		10 & Vote cast for president - Democratic 2008 & True \\ 
		\hline
		11 & Vote cast for president - Republican 2008 & True \\  
		\hline
		12 & Nonfamily households 2010 & True \\ 
		\hline
		14 & New private housing units authorized by building permits - total 2010 (20,000-place universe) & True \\ 
		\hline
		15 & People under age 18 in poverty - percent 2009 & False \\ 
		\hline
		16 & Families with income in the past 12 months (in 2009 inflation-adjusted dollars) of less than \$10,000 & True \\ 
		\hline
		17 & Families with income in the past 12 months (in 2009 inflation-adjusted dollars) of \$10,000 to \$14,999 & True \\ 
		\hline
		18 & Families with income in the past 12 months (in 2009 inflation-adjusted dollars) of \$15,000 to \$19,999 & True \\ 
		\hline
		19 & Families with income in the past 12 months (in 2009 inflation-adjusted dollars) of \$20,000 to \$24,999 & True \\  
		\hline
		20 & Families with income in the past 12 months (in 2009 inflation-adjusted dollars) of \$25,000 to \$29,999 & True \\  
		\hline
		21 & Families with income in the past 12 months (in 2009 inflation-adjusted dollars) of \$30,000 to \$34,999 & True \\  
		\hline
		22 & Families with income in the past 12 months (in 2009 inflation-adjusted dollars) of \$35,000 to \$39,999 & True \\  
		\hline
		23 & Families with income in the past 12 months (in 2009 inflation-adjusted dollars) of \$40,000 to \$44,999 & True \\  \hline
		24 & Families with income in the past 12 months (in 2009 inflation-adjusted dollars) of \$45,000 to \$49,999 & True \\  
		\hline
	\end{tabular}
\end{center}
\caption{\small \textbf{Sociodemographic features (1 of 2).}  The first 24 sociodemographic features from the US 2010 Census that are used as ``auxiliary data'' for the crime rate experiments presented. The last column describes whether the feature is provided to the models on a per-capita basis.}
\label{table:socios1}
\end{table}

\begin{table}[!htb]
	
	\begin{center}
		\begin{tabular}{|p{1.5cm} | p{9cm} | p{2.5cm}|} 
    		\hline
    		\textbf{ID} & \textbf{Feature Description} & \textbf{Provided Per-Capita} \\ 
    		\hline
    		25 & Families with income in the past 12 months (in 2009 inflation-adjusted dollars) of \$50,000 to \$59,999 & True \\  
    		\hline
    		26 & Families with income in the past 12 months (in 2009 inflation-adjusted dollars) of \$60,000 to \$74,999 & True \\  
    		\hline
    		27 & Families with income in the past 12 months (in 2009 inflation-adjusted dollars) of \$75,000 to \$99,999 & True \\  
    		\hline
    		28 & Families with income in the past 12 months (in 2009 inflation-adjusted dollars) of \$100,000 to \$124,999 & True \\  
    		\hline
    		29 & Families with income in the past 12 months (in 2009 inflation-adjusted dollars) of \$125,000 to \$149,999 & True \\  
    		\hline
    		30 & Families with income in the past 12 months (in 2009 inflation-adjusted dollars) of \$150,000 to \$199,999 & True \\  
    		\hline
			31 & Families w/ income in the past 12 months (in 2009 inflation-adjusted dollars) of \$200,000 or more & True \\ 
			\hline
			32 & Federal Government expenditure - total FY 2010 & True \\ 
			\hline
			33 & Accommodation and Food Services: accommodation (NAICS 721) - establishments with payroll 2007 & True \\ 
			\hline
			34 & Accommodation and Food Services: food services \& drinking places (NAICS 722) - establishments with payroll 2007 & True \\ 
			\hline
			35 & Commercial banks and savings institutions (FDIC-insured) - total deposits (June 30) 2010 & True \\ 
			\hline
			36 & Employment in all industries, net change 2000 - 2007 & True \\ 
			\hline
			37 & All persons under 18 years without health insurance, percent 2007 & False \\ 
			\hline
			38 & All persons 18 to 64 years without health insurance, percent 2007 & False \\ 
			\hline
			39 & Median household income 2009 & False \\ 
			\hline
			40 & Ethnicity-related feature \#1 & True \\ 
			\hline
			41 & Ethnicity-related feature \#2 & True \\  
			\hline
			42 & Ethnicity-related feature \#3 & True \\  
			\hline
			43 & Ethnicity-related feature \#4 & True \\  
			\hline
			44 & Ethnicity-related feature \#5 & True \\  
			\hline
			45 & Ethnicity-related feature \#6 & True \\  
			\hline
			46 & Ethnicity-related feature \#7 & True \\  
			\hline
			47 & Ethnicity-related feature \#8 & True \\  
			\hline
		\end{tabular}
	\end{center}
    \caption{\small \textbf{Sociodemographic features (2 of 2).}  The remaining 23 sociodemographic features from the US 2010 Census that are used as ``auxiliary data'' for the crime rate experiments presented. The last column describes whether the feature is provided to the models on a per-capita basis.}
	\label{table:socios2}
\end{table}

\subsection*{Baseline Approaches}

The likelihood of the primary data $y_i \given \theta_i \sim \poissonpmf(n_i \theta_i)$, where $n_i$ quantifies a measure of ``exposure'' (e.g. population when modeling per-capita rates), leads to the consideration of two possible baseline approaches: Poisson regression and Gamma-Poisson regression, sometimes referred to as Negative Binomial regression under certain parameter restrictions.  Both baselines fit their parameters to maximize the predictive likelihood,
\begin{align}
p(y \given x; \phi) = \prod_{i=1}^M p(y_i \given x_i; \phi)
 = \prod_{i=1}^M \int_0^\infty p(y_i \given \theta_i) p(\theta_i \given x_i; \phi) \dif\theta_i
\end{align}
Baseline approaches focus on learning a single transformation
\begin{align}
\mu(x_i; \phi) = \exp(\phi^\top x_i)
\end{align} 
where the inner product $\phi^\top x_i$ additionally includes a bias term that is omitted for brevity.  Gamma-Poisson  regression additionally learns an over-dispersion parameter $\delta$, which specifies the degree to which this distribution is overdispersed relative to the Poisson distribution.  As before, we consider the implied distribution over the latent variable.

For Poisson regression we have a posterior $p_\textrm{Po}(\theta_i \given x_i; \phi)$ with the entirety of its mass collapsed onto a single point, $\theta_i = \mu(x_i; \phi)$:
\begin{align}
p_\textrm{Po}(y_i \given x_i; \phi) &= \poissonpmf(y_i \given n_i \mu(x_i; \phi)) \\
p_\textrm{Po}(\theta_i \given x_i; \phi) &= \text{Dirac}(\theta_i - \mu(x_i; \phi))
\end{align}
The coupling of the mean and the variance under this model caused very poor performance, hence it has been omitted here for most visual comparisons.
Under Gamma-Poisson regression we have
\begin{align}
p_\textrm{GaPo}(y_i \given x_i; \phi) &= \gammapoissonpmf(y_i \given \alpha_0 + \delta^{-1}, n_i^{-1}(\beta_0 + \delta^{-1} \mu(x_i; \phi)^{-1})) \\
p_\textrm{GaPo}(\theta_i \given x_i; \phi) &= \gammapdf(\theta_i \given \alpha_0 + \delta^{-1} , \beta_0 + \delta^{-1} \mu(x_i; \phi)^{-1}).
\end{align}
The Gamma-Poisson baseline adopts a common uninformative prior, $\alpha_0 = \beta_0 = 0$.  From an interpretation standpoint, the shape parameter $\delta^{-1}$ corresponds to a certain number of arrivals in a certain number of intervals $\delta \mu(x_i; \phi)$.  Additionally, as $\delta \rightarrow 0$ the Gamma-Poisson distribution approaches the Poisson distribution with mean $\mu(x_i; \phi)$.

\FloatBarrier

Under our LDF approach with a single conjugate mapping we have
\begin{align}
p_\textrm{LDF}(y_i \given x_i; \phi) &= \gammapoissonpmf(y_i \given \alpha_0 + a(x_i; \phi), n_i^{-1}(\beta_0 + b(x_i; \phi))) \\
p_\textrm{LDF}(\theta_i \given x_i; \phi) &= \gammapdf(\theta_i \given \alpha_0 + a(x_i; \phi), \beta_0 + b(x_i; \phi)),
\end{align}
where $a(x_i; \phi)$ maps $x_i$ into a number of arrivals in a number of intervals $b(x_i; \phi)$.  The higher $b(x_i; \phi)$ the lower the posterior uncertainty.  We can again identify the Gamma-Poisson regression baseline as a special case of our approach by term-matching: $a(x_i; \phi) = \delta^{-1}$ and $b(x_i; \phi) = \delta^{-1} \mu(x_i; \phi)^{-1}$.  
For mixtures of conjugate mappings, referred to as LDF-MM, each component has the form specified above and the component assignment is marginalized out.



\clearpage
\small
\bibliography{refs}

\end{document}